\PassOptionsToPackage{table}{xcolor}
\documentclass[10pt,twoside,onehalfspacing,copyright,hyper, print]{VUthesis}

\usepackage {tikz}
\usetikzlibrary{calc,positioning}
\usetikzlibrary{shapes.geometric}
\definecolor {processblue}{cmyk}{0.96,0,0,0}
\usetikzlibrary{shapes.geometric, arrows, positioning, automata, positioning, fit, backgrounds, arrows.meta, calc}
\makeindex

\usepackage{bm}
\usepackage{amsfonts}

\usepackage{MnSymbol}
\usepackage[ruled, vlined,linesnumbered]{algorithm2e}
\usepackage[noend]{algpseudocode}
\usepackage{lipsum}
\usepackage{bookmark}
\usepackage{booktabs}
\usepackage{xltabular} 

\usepackage[dutch, english]{babel}
\usepackage{calc}  
\usepackage[font=small,format=plain,labelfont=bf,up,textfont=it,up,margin=10px,indention=5px, skip=5pt]{caption}
\usepackage{capt-of}
\usepackage{enumerate}
\usepackage{enumitem}
\usepackage{esvect}
\usepackage{float}
\usepackage{graphicx}
\usepackage{hyphenat}
\usepackage{hyperref}
\usepackage{ltablex}
\usepackage{listings}
\usepackage{multicol}
\usepackage{microtype}
\usepackage{multirow}
\usepackage{makeidx}
\usepackage{microtype}
\usepackage[square,semicolon,numbers]{natbib}
\usepackage[autolanguage,np]{numprint}
\usepackage{ragged2e}
\usepackage{rotating}
\usepackage{units}
\usepackage{stmaryrd}
\usepackage{supertabular}
\usepackage{subcaption}
\usepackage[countmax]{subfloat}
\usepackage{url}
\usepackage{verbatim}
\usepackage{wrapfig}
\usepackage{xspace}
\usepackage[para,online,flushleft]{threeparttable}
\usepackage{makecell}

\usepackage{graphicx}
\usepackage{wrapfig}
\usepackage{float}
\usepackage{ltablex}
\usepackage{capt-of}
\usepackage{supertabular}
\usepackage{changepage}
\usepackage{glossaries}
\usepackage{tabu}
\usepackage{docmute}

\makeglossaries

\newglossaryentry{ER}
{
    name=ER,
    description={The field of Evolutionary Robotics}
}

\definecolor{lime}{HTML}{A6CE39}
\usepackage{tabularx}
\usepackage{fp}
\usepackage{colortbl}
\usepackage{collcell}
\usepackage{pdfpages}

\usepackage{xr}
\graphicspath{ {images/} }

\ExplSyntaxOn
\tl_new:N \l__bourne_tmpa_tl
\cs_new_protected:Npn \bourne_path_push:
  {
    \seq_put_left:NV \l_file_search_path_seq \CurrentFilePath
    \AddToHookNext {file/after/\CurrentFile} { \bourne_path_pop: }
  }
\cs_new_protected:Npn \bourne_path_pop:
  { \seq_pop_left:NN \l_file_search_path_seq \l__bourne_tmpa_tl }
\NewDocumentCommand \topinput { }
  { \AddToHookNext {file/before} { \bourne_path_push: } \input }
\ExplSyntaxOff

\AtBeginEnvironment{longtabu}{\footnotesize}{}{}

\usepackage[explicit]{titlesec}

\usepackage{bookmark}

\renewcommand*{\thesection}{\Roman{section}}
\renewcommand*{\thesubsection}{\thesection.\Alph{subsection}}

\begingroup
  \usepackage{MnSymbol}
    \global\DeclareUnicodeCharacter{27F3}{\ensuremath{\textbf{$\bm{\lcirclearrowright}$}}}
    \global\DeclareUnicodeCharacter{27F2}{\ensuremath{\textbf{$\bm{\rcirclearrowleft}$}}}
    
\endgroup

\newcounter{tabenum}
\edef\citet{\noexpand\leavevmode
            \noexpand\protect
            \expandafter\noexpand\csname citet \endcsname}

\DeclareGraphicsExtensions{.pdf,.png,.jpg}

\makeindex
\floatstyle{ruled}
\newfloat{previous}{b}{lop}
\floatname{previous}{}
\title{\huge Robots Need Some Education\\ \large On the complexity of learning in evolutionary robotics}

\singlelinetitle{Robots Need Some Education}

\dutchsinglelinetitle{Robots Need Some Education}

\submissionDate{15}{September}{2025}

\author{Fuda van Diggelen}
\degreeHeld{MSc.}
\authorformal{\normalsize Fuda van Diggelen}
\birthplace{Kunming, China}

\rector{prof.dr. J.J.G. Geurts}
\phdfaculty{Faculteit der Bètawetenschappen}
\defensedate{vrijdag 21 maart 2025 om 9.45 uur}
\defenselocation{Aula}

\promotor{prof.dr.~A.E.~Eiben\\}
\copromotor{dr.ir.~E.~Ferrante\\}
\committee{\small
 \section*{Members of the committee}
 \begin{tabular}{@{\extracolsep{\fill}}ll}
Prof. Dr. ~K.~Glette & University of Oslo \\
Prof. Dr. ~D.~Floreano & École Polytechnique Fédérale de Lausanne \\
Dr. ~A.V.~Kononova  & Leiden University \\
Prof. Dr.~E.~Hart &	Edinburgh Napier University \\
Prof. Dr.~F.~Harmelen  &	Vrije Universiteit Amsterdam 
 \end{tabular}
}

\begin{document}
\includepdf[pages=1,fitpaper=true]{general/img/frontcover.pdf}

\frontmatter

\mainmatter
\thumbtrue
\begin{acknowledgements}
{
\setlength{\parindent}{0pt}
\setlength{\parskip}{1em}

\vspace{-2.5em}
On the first day of my PhD, I nervously knocked on the door of my new supervisor, professor Guszti Eiben, with a fair bit of anxiety. Why had I chosen to pursue a degree in Computer Science after graduating from Mechanical Engineering? What the hell did I know about hardware, computing, programming, or anything in-depth related to computers? What if I wasn’t good enough? I shared these thoughts with Guszti, to which he replied with a smile...  ``No need to worry, I don’t know anything about computers either."

The cheerful mentorship I received during my PhD has been truly delightful. I am deeply grateful to my supervisors, Guszti Eiben, Eliseo Ferrante, and Nicolas Cambier, for their support throughout the years. Without their help, completing this work would not have been possible.

Work is important, but having fun is more importanter. I will always look back fondly on the small `coffee breaks' with colleagues: flying drones in the office, practicing handstands, fun little game nights, cooking club, bicycle lessons, Russian dance sessions, late-night techno parties in the lab, professional table tennis competitions, high-stakes food challenges, and discussing micro-dosing with my favorite Polish professor. Thank you all for these amazing memories. 

Finally, I would like to express appreciation to my family—Hylda van Diggelen, Paul van Diggelen, and Lianne van Diggelen. Ontzettend bedankt voor alle liefde en steun door al de jaren.

Love to you all,\\
Fuda van Diggelen

}
\end{acknowledgements}

\begin{summary}
    \begin{center}
    \large \textbf{Summary}
    \end{center}

    Evolutionary Robotics and Robot Learning are two fields in robotics that aim to automatically optimize robot designs. The key difference between them lies in what is being optimized and the time scale involved. 
Evolutionary Robotics is a field that applies evolutionary computation techniques to evolve the morphologies or controllers, or both \cite{vargas2014horizons}. Robot Learning, on the other hand, involves any learning technique aimed at optimizing a robot's controller in a given morphology. In terms of time scales, evolution occurs across multiple generations, whereas learning takes place within the `lifespan' of an individual robot.

The long-term goal of Evolutionary Robotics is to create adaptive systems where a population of robots evolve autonomously, optimizing both their physical structure and control system through an evolutionary process. In the context of evolution, adaptability is a trait of the population. Conversely, when it comes to learning, it is the robots' controller that exhibits adaptability. In the end, both forms of adaptation aim to enhance the robots' task performance.



Integrating Robot Learning with Evolutionary Robotics seems like a natural fit for improving robot design. Unfortunately, integration requires the careful design of suitable learning algorithms in the context of evolutionary robotics. The effects of introducing learning into the evolutionary process are not well-understood and can thus be tricky. This thesis investigates these intricacies and presents several learning algorithms developed for an Evolutionary Robotics context.

My dissertation is structured into three parts:

\textbf{Part I} - investigates the complex interaction between learning algorithms and evolutionary processes, provides statistical tools for evaluating different learning algorithms, and explores the dynamics of optimization and the reality gap. The interactions present a counterintuitive finding: learning can negatively affect the evolutionary process. Learning can bias evolution by converging to simple designs that learn quickly, overfitting simulators and exacerbating the reality gap.

\textbf{Part II} - investigates model-agnostic learning methods for autonomous robots. Evolutionary algorithms can produce arbitrary robot designs for environments with minimal prior knowledge. This presents a challenging requirement for learning. Nevertheless, robots are expected to be able to autonomously perform tasks `in the wild'. Here, I cover continuous self-modeling for adaptive feedback control and rapid skill acquisition for quickly learning locomotion.

\textbf{Part III} - explores how learning can be extended beyond individual robot. In group settings, swarms of robots can obtain abilities beyond that of any individual robot inside. Here, I demonstrate how such emergent capabilities can be learned and used to solve complex tasks, both in homogeneous and heterogeneous populations of robots.

Overall, this thesis offers a comprehensive analysis of Robot Learning within the context of Evolutionary Robotics presented as a collection of peer-reviewed works. Each part of the dissertation combines rigorous theoretical analysis with practical hardware implementations, demonstrating the validity of the ideas presented. As a result, my thesis provides unique insights into how an evolving population of robots can effectively integrate learning. From the ``birth" of a robot, learning about its environment, to its ``adulthood" as a functioning member of a robotic society.
\end{summary}

\chapter{Introduction}
\label{ch1:introduction} 

 
 \newpage

      

\section{Motivation and Contributions}

The field of \textit{Evolutionary Robotics} (ER) envisions self-adapting systems where populations of robots evolve their overall design, whether it be controllers, sensors, composition, and building materials. Evolution here is employed as a computational technique \cite{eiben2003introduction}, through which robots within the system are automatically designed and optimized. Such an algorithmic approach should be capable of producing intelligent artificial life...

\begin{quote}
    \textit{Given the fact that evolution can produce intelligence,\newline \hspace{1em} it is plausible that artificial evolution can produce artificial intelligence} \cite{Eiben}
\end{quote}

Historically, the field of robotics focused on extremely precise control in hyper-controlled environments (in specific machinery, labs, and factories). 
Current research in robotics is shifting more towards less controllable environments (outside of a lab), for which designing robots is more challenging. 
ER solves this challenge by designing robot-generating algorithms \cite{bongard2013evolutionary}, that could be more suited to complex environments and less biased than human design. This would be attractive for deployment in desolated places with unknown and dynamic environments or when facing unforeseen circumstances where humans cannot intervene. For example, an ER system could be sent to Mars to begin mining for resources, build infrastructure, and develop cities, making it a habitable environment before humans arrive. For this, an ER system designs robots in an autonomously adapting population. 

How do we design the robot designer? This is the essence of ER research. Unfortunately, a straightforward approach to answering this question remains unclear, making the field of ER very broad and interdisciplinary \cite{floreano2008evolutionary}. Research topics can range from computer science to mechanical engineering, biology, neuroscience, and even philosophy, with a key ingredient at its core, the Evolutionary Algorithm (EA) to optimize robot design. The EA drives the design process to promote the proliferation of certain advantageous traits in the robot population. A structural overview of the main phases of a robot evolution process is captured by the Triangle of Life (ToL) model \cite{eiben2013triangle}.

\begin{figure}[ht]
    \centering
    \includegraphics[width=0.425\textwidth]{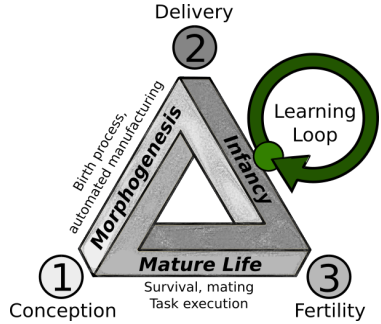}
    \caption{Triangle of Life}
    \label{fig:tol_intro}
\end{figure}

The ToL formulates the different `stages' in the life of a single robot within an evolutionary process. During its \textit{mature life}, the robot performs tasks, strives for survival, and reproduction \cite{de2023interacting} (bottom side, \autoref{fig:tol_intro}). Throughout this `mature life', the selection probabilities of the robots are influenced by the given tasks \cite{de2020Tasks}, environmental factors \cite{miras2019effects} and possibly other criteria in the EA \cite{miras2018effects}. When robots are considered good candidates for `mating', a new robot is \textit{conceived} (Node 1 \autoref{fig:tol_intro}). By combining robot designs \cite{jelisavcic2019lamarckian, gupta2021embodied, luo2023enhancing}, we can (hopefully) obtain a new robot design that inherits good features from its parents. Here, there are several choices regarding the representation of robots and the corresponding reproduction operators (mutation and crossover) that influence the evolutionary dynamics \cite{nygaard2017overcoming, veenstra2017evolution}. 

Following `conception', the birthing takes place, which is described as \textit{morphogenesis} (left side, \autoref{fig:tol_intro}). In the future, fully autonomous ER systems will assemble new robots without human intervention \cite{angus2023practical}, making robot evolution an autonomous process. In the end, a new robot is \textit{delivered} with (hopefully) better traits than its parent(s). Unfortunately, a new robot with a unique body requires a unique controller, and evolutionary reproduction often results in a mismatch between body and brain. Without further optimization (fine tuning) such an individual would not be attractive enough to be selected as a potential parent. 

This problem is addressed in the \textit{infancy} stage (right side, \autoref{fig:tol_intro}). In this stage, at the start of a `robot life', robots must undergo a learning process \cite{eiben2020if}.

\begin{quote}
    \textit{If it evolves it needs to learn} \cite{eiben2020if}
\end{quote}

The central challenge I face in this thesis is how to effectively implement robot learning in the context of evolutionary robotics. 
Adaptive systems are abundant in nature, as they help in tackling unforeseen problems within the randomness of an uncontrolled environment. A continuously adapting robot design, be it the body or brain, provides additional robustness and consistency in robot performance \cite{Wright2015}. 

Robot learning exhibits adaptability by changing the \textit{robot controller} \cite{GUO2023composite}. Similarly, ER exhibits adaptability by changing the \textit{robot population} \cite{chatzilygeroudis2019survey}. 
In this light, the integration of robot learning is a logical improvement to the overall adaptive capabilities of the evolving population. Unfortunately, simply nesting robot learning as an additional loop inside the EA brings some complications, as the amount of compute grows rapidly with each additional learning iteration. Furthermore, the evolutionary process operates by testing the `lifetime' performance of many robots over multiple generations, often with minimal prior knowledge of the task(s) or robot body designs. Robot learning, on the other hand, is focused on improving the controller within the `lifespan' of an individual robot, often on a well-defined task with a given robot body.

In the context of ER, robot learning becomes significantly more complex due to the interactions between the two adaptive systems.
Correct integration of robot learning into ER should benefit not only the specific robot itself but also the evolutionary process that optimizes the robot population. Unlike most works on robot learning, \textit{learning itself is not the goal}. The effectiveness of the ER system depends on the effectiveness of the \textit{robot population} not on the effectiveness of a specific learning algorithm. In the end, learning occurs on the individual level, but the effectiveness of the robot population considers interaction beyond this individual level \cite{de2023interacting}. 

A learning algorithm can improve the performance of certain morphologies, which then influences the evolutionary dynamics. Ideally, robot learning helps unlock the true potential of a given morphology. Therefore, it can help guide evolution and thus help in the proliferation of better robot designs \cite{nolfi1999learning}. Unfortunately, learning algorithms are often flawed, introducing biases, and creating more noise when evaluating robot performance. This requires the delicate design of suitable learning algorithms within the ER with a holistic perspective.

\subsection{Research goal}
This thesis investigates the application of learning within ER. The objective is to create learning algorithms that enhance both the learning performance and the evolutionary process.
Various challenges are identified and tackled at multiple layers of control abstraction. These include individual-level control for direct motor commands, self-modeling for adaptive feedback control, the acquisition of behavioral skill repertoires, and coordinated population-level behaviors in robotic swarms. The subsequent sections will present this in detail.

\textbf{Learning is a tool not an objective}. It is crucial to emphasize that learning is not the ultimate goal in ER. Learning serves as an initial phase before utilizing the acquired skill(s) to complete tasks in the real-world. These tasks can be complex and often require continuous adaptation or quick (re)learning of skills. It could be argued that the robot's behavior in the real world --specifically during the final phase of its operational life as shown in \autoref{fig:tol_intro}-- carries greater weight in the evolutionary process than the performance of the learning algorithm. 

\textbf{Learning with minimal prior knowledge}. The ER system assumes little to no knowledge on the environment it is employed in. Furthermore, adaptivity is achieved through continuously changing the designs of robots in the population. Therefore, minimal prior knowledge is present for learning: robot structure, sensors, material, and its environment are some of the unknowns. This necessitates the employment of model-agnostic algorithms during the learning stage. Learning here should be focused on autonomy (as in autonomous robots) through gaining `understanding' of their environment and acquiring locomotion skills.

\textbf{Emergent population based efficacy}. An evolving system that can solve complex tasks requires high-level coordination between individuals. ER is well-suited for exploring the ways in which a diverse group of robots, i.e. a heterogeneous swarm, can learn to cooperate efficiently. Such a heterogeneous swarm can obtain so-called emergent capabilities that extend beyond the individuals' performance. The ER population should learn to leverage these collective behaviors to complete complex tasks like collecting resources and building infrastructure. 

This dissertation presents a collection of studies on robot learning within an ER system. The main goal of this thesis is to integrate robot learning and ER in a mutually beneficial design. More concretely, I show how to tackle different robot learning tasks on several levels of control within the context of evolutionary robotics. The presented works cover different learning tasks that enable evolvable robots to continuously adapt within an interacting `robot society' from scratch. Starting from understanding the world through adaptive feedback control, to model-agnostic functional skill acquisition, and complex (heterogeneous) swarm control. 

The main contributions of the thesis are:

\begin{enumerate}
    \item Overall, a detailed investigation on the interaction between learning and ER. Including in-depth analysis tools, test suite, and datasets.
    \item Development of a model agnostic adaptive feedback algorithm that can be applied on any type of robot.
    \item Development of a fast skill acquisition algorithm capable of learning multiple skills in parallel.
    \item Development of open-source evolutionary aided design software for learning emergent control in a (heterogeneous) population of robots.
    \item Real world implementations of most of the presented controllers/learning methods. 

\end{enumerate}

In addition to these contributions, this dissertation provides valuable information on how ER can position itself within the field of robotics and learning. From an engineering point of view, ER can come across as an infeasible idea, far-fetched from what is currently relevant. This stereotype is reinforced by the lack of real world robot experiments and the strong inclination of research to focus on \textit{virtual creatures} in unrealistic environments (resembling more of a videogame than potential robots with function). Looking at this type of research, it is hard to imagine how it would lead to a capable population of robots that build roads and cities on Mars. From learning autonomy to solving complex problems on population-level, this work presents how robot learning could facilitate the functioning of an ER population.

\section{Scope}
\label{pt0ch3:overview}
The presented dissertation is a selection of papers published in a span of 4 years. Each paper is presented in a separate chapter, which I bundled into three distinct parts. Before inclusion, the chapters underwent slight editorial changes to harmonize the thesis, notation, figure size, and references within chapters. It should be noted that in Chapter \ref{ch5:MP} the Appendix is presented online at \url{https://www.nature.com/articles/s41467-024-50131-4\#Sec28} and in Chapter \ref{ch7:SC2} additional pre-liminary hardware experiments were included. The three parts are structured as follows.

\begin{enumerate}[label=\textbf{\Roman*}:,leftmargin=*]
    \item Learning in the context of Evolutionary Robotics
    \item Autonomy for the unknown
    \item An evolving robot population
\end{enumerate}

\textbf{\autoref{PI} - Learning in the context of Evolutionary Robotics}.
Nesting learning in evolutionary optimization adds complexity on several levels: 1) Learning is model-agnostic (limited knowledge on the environment, robot design, and tasks); 2) Learning is resource limited, computationally and physically the EA has to evaluate many robots (and the main focus is the robot's performance during its `mature life'); 3) Learning is not the ultimate goal, but done in service of a bigger task. These constraints require new techniques to assess the quality of learning algorithms, uniquely positioned in the field of robot learning. 
In Chapter \ref{ch2:compare_learning}, I introduce the right tools to analyze learning algorithms for ER purposes. Special consideration is given to the rapidly increasing number of evaluations (due to the nested learning loop), model-agnosticity (to reduce bias during evolution), and the importance of robustness and consistency in ER. A more efficient way to analyze learning in the context of ER is presented through a representative test suite of evolved robots. Using this test suite, I can quickly compare the performance of several learning algorithms and propose two new performance metrics to gauge efficiency and efficacy. 
In Chapter \ref{ch3:RG}, I focus on the interaction between evolution and the reality gap (RG). The RG indicates the exploitation of inaccuracies in simulations, which is unavoidable in simulator-based optimization. The RG can lead to unrealistic results that `perform well' in simulation but are unattainable in the real world. Learning in ER exacerbates the RG problems, as unrealistic designs start to dominate the robot population. Building on the analysis tools developed in Chapter \ref{ch2:compare_learning}, I analyze the dynamics of the RG and create robust heuristics to limit its effects on the robot population.

In summary, \autoref{PI} provides essential insights on the difficulties in learning within ER and presents the tools to analyze the learning performance of different algorithms and their dynamic with respect to the RG. 

\textbf{\autoref{PII} - Autonomy for the unknown}. 
An attractive use case for ER is the development of robots that adapt to an environment with limited information and accessibility. In situations where environments are unfamiliar and evolution can yield unexpectedly innovative designs, autonomous robot control needs to be achieved with very limited prior knowledge. Currently, learning without a proper model takes considerable time, as distinguishing noise from self-imposed movements can be fuzzy. The design of fast model-agnostic learning algorithms is a must. 
In Chapter \ref{ch4:IMC}, I focus on `understanding' the world through self-modeling for feedback control. Proper movement execution (how to deal with noise and disturbances) is important to ensure consistency in control, especially outside of a controlled environment. In this setting, robots learn to model the interactions between its sensory input and controller output on the fly. The resulting model predicts the consequences of the robot's actions and integrates this information to move with minimum error. Note that the focus is task-independent. Instead, the goal is to learn predictive models to provide reliable feedback control in the real world, regardless of any (learning) task. 
In Chapter \ref{ch5:MP}, I present a fast learning method to obtain multiple skills in parallel. Learning a repertoire of skills forms the basis for more complex behaviors in robotics, as many tasks can require the use of several combinations of skills. ER requires this learning process to be extremely fast --evolution considers performance during the `mature life', not the actual learning performance-- for any morphology. I developed a novel model-agnostic method to learn multiple basic locomotion skills within 15 minutes from scratch. The resulting skill repertoire is used to solve a more complex task in a preliminary target following experiment. 

In summary, \autoref{PII} presents two model-agnostic learning algorithms for robot control in the real world. The resulting works show an effective way to obtain viable skills with feedback control for any type of modular robot.

\textbf{\autoref{PIII} - An evolving robot population}. 
Up until now, the phrase `robot population' has been used loosely in the context of an ER optimization process and as an instance of a group of (interacting) robots.  Although most of the ER field is not concerned with the latter (with some notable exception \cite{buresch2005effects, de2023interacting, miconi2008evosphere}), a population where individuals
closely interact and collaborate will enhance overall performance. 
For example, a single worker is not able to complete the task to `build a city' on its own, but as a collective group they obtain an emergent capability to do so. Thus, inter-individual interactions can have a positive influence on the overall performance of the whole population (in the EA). For ER it will be beneficial to not only consider the optimization of a single robot design, but, additionally, how they interact as a group (i.e. robot swarms). In \autoref{PIII}, I focus on the automated design of such complex swarm controllers.
In Chapter \ref{ch6:SC}, I provide an open-source pipeline (in collaboration with Jie Luo and Tugay Karagüzel) to evolve a repertoire of (robot) behaviors in the form of a reservoir neural network, for complex swarm coordination. Here, I introduce the learning pipeline for swarm control, and showcase its effectiveness in an emergent gradient sensing task (meaning, an individual is unable to solve this task on its own).
In Chapter \ref{ch7:SC2}, I extend the complexity of the controller to evolve a heterogeneous swarm with undefined specialization. Heterogeneity is expected in the population, as the EA provides variation between robots. Here, it is shown that heterogeneity increases swarm performance in terms of scalability and robustness, using an online adaptive mechanism that switches between the evolved behaviors.

In summary, \autoref{PIII} considers the optimization of group coordination to increase the capabilities of the robot population. The resulting pipeline is capable of obtaining emergent capabilities, using a reservoir of behaviors, in a heterogeneous robot swarm.

 
\subsection*{List of Papers}
\label{pt0ch3subsec:papers}

This thesis is the result of four years of research and is based on the content of 3 journal papers and 3 conference papers. These papers are listed below, along with details of my contribution to each one.

\begin{table}[ht!]\footnotesize
    \setlength{\textfloatsep}{0.7\baselineskip plus 0.2\baselineskip minus 0.5\baselineskip}    
    \footnotesize
    \caption{\small Paper overview: Roman and Arabic numerals in the left-most columns indicate thesis \textbf{Part} and \textit{Chapter}}
    \label{tab:parameters_exp}
    \centering
    \renewcommand{\arraystretch}{1}

    \begin{tabular}{lrllll}
    \multicolumn{2}{r}{ } & Topic & Paper & Year \\ 
    \toprule
    \multicolumn{2}{r}{\textbf{I\hphantom{II}} $2$} & Learning in evolutionary robotics & [P1] & $2023$ \\
     & $3$ & Evolutionary dynamics of the reality gap & [P2] & $2021$ \\ \midrule
    \multicolumn{2}{r}{\textbf{II\hphantom{I}} $4$} & Self-modelling and learning & [P3] & $2020$ \\ 
    & $6$ & Fast skill acquisition in evolvable robots & [P4] & $2024$ \\ \midrule
    \multicolumn{2}{r}{\textbf{IV}\hspace{0.5pt} $7$} & Controller for evolvable swarms & [P5] & $2022$ \\
     &$6$  & Emergent sensing in heterogeneous swarms & [P6] & $2024$ \\
    \bottomrule
    \end{tabular}
\end{table}

\subsection*{Contribution}
The following section presents my personal contribution to all chapters. The names of the main contributors are shown in \textbf{bold}. For all papers, the main body of the paper was written by me.
\begin{flushleft}
\textbf{Part I - Evolution, Robotics and Learning}\vspace{0.5mm}\hrule\vspace{1mm}
\textit{Chapter 2}: \textbf{van Diggelen, F.}, Ferrante, E., \& Eiben, A. E. (2023). Comparing Robot Controller Optimization Methods on Evolvable Morphologies. \textit{Evolutionary Computation}, \textit{32}(2), pp. 105–124. MIT Press. \href{https://doi.org/10.1162/evco_a_00334}{doi: 10.1162/evco\_a\_00334}.
\end{flushleft}

I created the 20 robots test suite and implemented all optimization code in Revolve \cite{hupkes2018revolve}. In addition, I designed the statistical framework to compare different learning algorithms for ER and built all analysis tools.

\begin{flushleft}
\textit{Chapter 3}: \textbf{van Diggelen, F.}, Ferrante, E., Harrak, N., Luo, J., Zeeuwe, D., \& Eiben, A. E. (2021). The influence of robot traits and evolutionary dynamics on the reality gap. \textit{IEEE Transactions on Cognitive and Developmental Systems}, \textit{15}(2), pp. 499-506. IEEE Press. \href{https://doi.org/10.1109/TCDS.2021.3112236}{doi: 10.1109/TCDS.2021.3112236}.
\end{flushleft}

I designed and implemented the heuristic measures to analyze the behavior in simulation for reality gap prediction. I conducted most of the real-world experiments.

\begin{flushleft}
\textbf{Part II - Autonomy for the unknown}\vspace{0.5mm}\hrule\vspace{1mm}
\textit{Chapter 4}: \textbf{van Diggelen, F.}, Babuska, R., \& Eiben, A. E. (2020). The effects of adaptive control on learning directed locomotion. \textit{In 2020 IEEE Symposium Series on Computational Intelligence (SSCI)}, pp. 2117-2124. IEEE Press. \href{https://doi.org/10.1109/SSCI47803.2020.9308557}{doi: 10.1109/SSCI47803.2020.9308557}.
\end{flushleft}

I developed the adaptive feedback controller in C++ and conducted the experiments in Revolve.

\begin{flushleft}
\textit{Chapter 5}: \textbf{van Diggelen, F.}, Cambier, N., Ferrante, E., \& Eiben, A. E. (2024). A model-free method to learn multiple skills in parallel on modular robots, \textit{Nature Communications}, \textit{15}(1), pp. 6267. Springer-Nature Publishing group. \href{https://doi.org/10.1038/s41467-024-50131-4}{doi: 10.1038/s41467-024-50131-4}.
\end{flushleft}

I designed the learning method and performed the mathematical analysis on the CPG structure (in collaboration with Aart Stuurman). I conducted all experiments in the real world and built the custom multi-camera tracking system, needed for the experimental work.

\begin{flushleft}
\textbf{Part III - An evolving robot population}\vspace{0.5mm}\hrule\vspace{1mm}
\textit{Chapter 6}: \textbf{Van Diggelen, F., Luo, J., Karagüzel, T. A.}, Cambier, N., Ferrante, E., \& Eiben, A. E. (2022). Environment induced emergence of collective behavior in evolving swarms with limited sensing. \textit{In Proceedings of the Genetic and Evolutionary Computation Conference (GECCO)}, pp. 31-39. ACM Press. \href{https://doi.org/10.1145/3512290.3528735}{doi: 10.1145/3512290.3528735}.
\end{flushleft}

I built part of the evolutionary pipeline for swarm evolution. In addition, I conducted the analysis and worked on the controller design.

\begin{flushleft}
\textit{Chapter 7}: \textbf{Van Diggelen, F.}, De Carlo, M., Cambier, N., Ferrante, E., \& Eiben, A. E. (2024). Emergence of Specialized Collective Behaviors in Evolving Heterogeneous Swarms. \textit{Parallel Problem Solving from Nature (PPSN XVIII)}, \textit{LNCS 15149}, pp. 53-69. Springer Nature Switzerland. \href{https://doi.org/10.1007/978-3-031-70068-2_4}{doi: 10.1007/978-3-031-70068-2\_4}. 
\end{flushleft}

I designed the adaptive heterogeneous swarm controller and implemented all of the code and analysis. I wrote the embedded software for real swarm robotics experiments.
\cleardoublepage
\part{Learning in the context of Evolutionary Robotics}\label{PI}

\chapter{Learning in evolutionary robotics}
\label{ch2:compare_learning}

 \begin{previous}
{\small\textsf{
\begin{flushleft}
\noindent Chapter~\ref{ch2:compare_learning} was published as:\\
\vspace{10pt}
 van Diggelen, F., Ferrante, E., \& Eiben, A. E. (2023). Comparing Robot Controller Optimization Methods on Evolvable Morphologies. \textit{Evolutionary Computation}, \textit{32}(2), pp. 105-124. 
 \href{https://doi.org/10.1162/evco_a_00334}{doi: https://doi.org/10.1162/evco\_a\_00334}.
\end{flushleft}}}
\end{previous}
\clearpage
\normalsize
\topinput{1/main}

\chapter{Reality Gap}
\label{ch3:RG}
 \begin{previous}
{\small\textsf{
\begin{flushleft}
\noindent Chapter~\ref{ch3:RG} was published as:\\
\vspace{10pt}
 van Diggelen, F., Ferrante, E., Harrak, N., Luo, J., Zeeuwe, D., \& Eiben, A. E. (2021). The influence of robot traits and evolutionary dynamics on the reality gap. \textit{IEEE Transactions on Cognitive and Developmental Systems}, \textit{15}(2), pp. 499-506. \href{https://doi.org/10.1109/TCDS.2021.3112236}{doi: 10.1109/TCDS.2021.3112236}.
\end{flushleft}}}
\end{previous}
\clearpage
\normalsize
\topinput{2/Main}

\part{Autonomy for the unknown
}\label{PII}
\chapter{Self-modelling for adaptive feedback control}
\label{ch4:IMC}
 \begin{previous}
{\small\textsf{
\begin{flushleft}
\noindent Chapter~\ref{ch4:IMC} was published as:\\
\vspace{10pt}
 van Diggelen, F., Babuska, R., \& Eiben, A. E. (2020). The effects of adaptive control on learning directed locomotion. \textit{In 2020 IEEE Symposium Series on Computational Intelligence (SSCI)}, pp. 2117-2124. \href{https://doi.org/10.1109/SSCI47803.2020.9308557}{doi: 10.1109/SSCI47803.2020.9308557}.
\end{flushleft}}}
\end{previous}
\clearpage
\normalsize
\topinput{3/AAC_main}

\chapter{Fast skill acquisition}
\label{ch5:MP}
 \begin{previous}
{\small\textsf{
\begin{flushleft}
\noindent Chapter~\ref{ch5:MP} was published as:\\
\vspace{10pt}
van Diggelen, F., Cambier, N., Ferrante, E., \& Eiben, A. E. (2024). A model-free method to learn multiple skills in parallel on modular robots, \textit{Nature Communications}, \textit{15}(1), pp. 6267. \href{https://doi.org/10.1038/s41467-024-50131-4}{doi: 10.1038/s41467-024-50131-4}.
\newline
\noindent The Appendix is provided online at: \newline\url{https://www.nature.com/articles/s41467-024-50131-4\#Sec28}
\end{flushleft}}}
\end{previous}
\clearpage
\normalsize
\topinput{4/sn-article}

\part{An evolving robot population}\label{PIII}
\chapter{Emergent sensing}
\label{ch6:SC}
 \begin{previous}
{\small\textsf{
\begin{flushleft}
\noindent Chapter~\ref{ch6:SC} was published as:\\
\vspace{10pt}
Van Diggelen, F., Luo, J., Karagüzel, T. A., Cambier, N., Ferrante, E., \& Eiben, A. E. (2022). Environment induced emergence of collective behavior in evolving swarms with limited sensing. \textit{In Proceedings of the Genetic and Evolutionary Computation Conference (GECCO)}, pp. 31-39. \href{https://doi.org/10.1145/3512290.3528735}{doi: 10.1145/3512290.3528735}.
\end{flushleft}}}
\end{previous}
\clearpage
\normalsize
\topinput{5/main}

\chapter{Adaptive Heterogeneous Swarms}
\label{ch7:SC2}
 \begin{previous}
{\small\textsf{
\begin{flushleft}
\noindent Chapter~\ref{ch7:SC2} was published as:\\
\vspace{10pt}
Van Diggelen, F., De Carlo, M., Cambier, N., Ferrante, E., \& Eiben, A. E. (2024). Emergence of Specialized Collective Behaviors in Evolving Heterogeneous Swarms. \textit{Parallel Problem Solving from Nature (PPSN XVIII)}, \textit{15149}, pp. 53-69. \href{https://doi.org/10.1007/978-3-031-70068-2_4}{doi: 10.1007/978-3-031-70068-2\_4}. 
\end{flushleft}}}
\end{previous}
\clearpage
\normalsize
\thumbtrue
\topinput{6/AAMAS-2024-Formatting-Instructions-CCBY/AAMAS_2024_sample}

\bookmarksetup{startatroot}
\chapter{Concluding Remarks}
\label{ch10:conclusion} 


This thesis sheds light on the difficulties regarding robot learning in the context of an evolutionary robotics system. At the start of my research, I developed the tools necessary to analyze the performance of learning algorithms and gauge how they affect the evolutionary process. Subsequently, I designed several learning methods that help improve the capabilities of robot (populations) for unknown designs and environments. This includes the development of 1) Robust adaptive feedback control for any type of robot; 2) A model-free method for fast multi-skill learning in parallel; 3) Automated design for online adaptive swarm controller with emergent specialized capabilities. In the end, this collection of works shows how robot learning could benefit evolution in a `robot society'.

\newpage
ER investigates the development of an autonomously adapting population of robots that can complete complex tasks for uncertain or dynamic environments. It is hard to envision how such a complex idea can be realized in the near future, given the current state of robotics and the limited applications of ER in the real world. On the other hand, many of the challenges currently faced in robotics seem to be limited to toy-problems in a controlled environment of a laboratory (e.g. balancing a pendulum, following a trajectory, picking up a cube). This work provides a fundamental bridge between robot learning and ER and provides a glimpse of what a fully autonomous and adaptive population of evolving robots could be.

In \autoref{PI}, I developed tools that improve our understanding of learning dynamics during an ER process. Integrating robot learning in ER formalizes a unique process that requires the delicate design of several learning stages to improve the functionality of the population. Unfortunately, designing suitable learning algorithms is challenging, as the nested learning loop leads to a significant increase in total workload. 

Chapter \ref{ch2:compare_learning} provides the robot test suite and methods for analyzing learning performance in the context of ER. This process was thoughtfully designed to obtain a representative cross-section of the robot varieties encountered during an evolutionary process. Throughout this thesis, this test suite was used to provide valuable insights on the effects of introducing learning in ER. 
This chapter evaluated various robot learning algorithm and introduced new statistical metrics measures: \textit{Robustness} (to indicate relative model agnosticism and potential morphological biases of the learning algorithm) and \textit{Consistency} (to gauge the unpredictability of learning performance and affects the sensitivity to early convergence in ER). 

In Chapter \ref{ch3:RG}, I paid special attention to learning performance and the reality gap. Physics simulators --due to their inherent simplifications and errors-- present potential inaccuracies that optimizers will exploit for higher performance. This issue is exacerbated in ER, as such exploits propagate unrealistic robot designs within the population during evolution; resulting in a contradicting dynamic: a high performing robot leads to a worse performing ER population. The onset of this dynamic was detectable, allowing us to design heuristic measures to reduce the exploitation of the simulator. 

Determining a 'best' learning algorithm is complex, as it is highly contextual for the specific task, environment, and design space. In general, the tools and metrics developed in \autoref{PI} provide a general framework to evaluate learning algorithms in ER. In the end, this part provides a comprehensive overview of how learning influences the ER process. 


\autoref{PII} focused on robot learning with minimal prior knowledge. ER systems are designed to be adaptable for unknown environments and therefore minimal prior knowledge on, surroundings, possible tasks, and optimal the robot designs can be assumed. As a consequence, robot learning should be model-agnostic. Furthermore, getting autonomy quickly is crucial, as learning is an intermediate process before the robots' mature life in ER (see \autoref{fig:tol_intro}).

In Chapter \ref{ch4:IMC}, I implemented adaptive models for an Internal Model Control (IMC) design. This bioinspired design obtains optimal feedback control by self-modeling the interaction with the environment. Optimal adaptive feedback control is extremely robot-specific and, due to nonlinearities, highly unstable when applied improperly. This model-agnostic method continuously updated sensorimotor mappings (i.e. self-models) `on the fly' to predict future sensor input, subsequently used to mitigate the effects of noise. The resulting feedback controller can serve on any type of robot design, due to its model-agnostic design. IMC significantly improved robustness to noise and perturbations and, most remarkably, improved the overall performance of the evolutionary algorithm on a locomotion learning task. The latter can be attributed to the fact that continuous self-adapting models play a crucial role in preventing consistent exploits of the simulator, thereby increasing the consistency of learning. 

In Chapter \ref{ch5:MP}, I designed a novel method to quickly obtain multiple skills in parallel. Through thorough mathematical analysis I showed that optimizing initial states of central pattern generator networks (or any recurrent system like Reinforcement Learning, Recurrent NN \textit{etc.}) is less complex than traditional methods that use weight optimization. The tools in \autoref{PI} are used to assess the difference between ISO and WO and other algorithms in this model-agnostic skill learning. The method is able to obtain a repertoire of skills (turning left/right, jumping, `forward' locomotion) in under 15 minutes in the real world, and utilizing the learned behaviors in a more complex target following task.

Overall, \autoref{PII} presents the early stages of learning in ER, where the robot obtains autonomy. This entails learning to `understand' the world through IMC and rapidly acquiring locomotion skills for more complex tasks. The algorithms I developed showed great proficiency in fast robot learning on a wide variety of different robots. 

In \autoref{PIII}, I built an evolutionary pipeline for the automated design of swarm controllers. In the end, a successful physical implementation of this system was presented. The method extends the idea of a repertoire of behavioral skills (formally defined in \autoref{PII}) as a starting point for learning group coordination.

In Chapter \ref{ch6:SC}, I developed the evolutionary pipeline and introduce the task of group-level coordination for emergent sensing.
Reservoir neural networks are the basis for optimizing coordinated movements between robots in a swarm. The reservoir consists of random input-output functions, whose combined signal is optimized in the last layer of the network. In the future, the `random' reservoirs could be replaced by the repertoire of kills from Chapter \ref{ch5:MP} or more sophisticated forms of functionalities. In the end, the pipeline was able to learn \textit{emergent capabilities} beyond the abilities of the individuals in the swarm. Such `swarm behavior' shows the potential of a `robot society' to enhance the performance on a group level. 

In Chapter \ref{ch7:SC2}, I extended this work for an evolving population that can consists of different functioning robots (a heterogeneous swarm). Heterogeneity introduces additional benefits (with respect to homogeneous swarms), as parts of the swarm can divide their tasks into sub-tasks for specialization. The presented learning algorithm automatically identified proper sub-tasks and optimized specialized behaviors accordingly. The resulting controller achieved higher performance in terms of robustness, scalability, changing environments, and the reality gap. In addition, I show an extension of our heterogeneous control with an online adaptive algorithm to select different instances of adaptive control, which is successfully employed on a real swarm. 

In general, \autoref{PIII} provides insight into how learning at the population level could improve the effectiveness of the ER system and complete more complex tasks. Most ER works do not consider the population as interactive robots (with some notable exception \cite{buresch2005effects, de2023interacting, miconi2008evosphere}). When ER is employed in the real world, effective coordination should be part of the learning process.

\section{Future Work}
There is a multitude of research questions worth exploring to better understand the interaction between learning and ER. Here, several broader research directions are outlined with some concrete examples. For future work I propose to investigate:
\begin{enumerate}
    \item \textbf{Investigate the evolutionary design of learning algorithms}. Meta-learning considers the design of learning algorithms as an additional optimization process, e.g. hyperparameter optimization or network architecture \cite{floreano2008neuroevolution}. In addition to this meta-learning approach, we can consider evolving learning goals for optimizing curriculum design (repertoire of skills) or diversity of capabilities in the population. 
    
    \item \textbf{Employ IMC for model-based tasks}.
    The sensorimotor model learnt during IMC provides valuable information that can be used for other (learning) tasks. For example, we can use the predictive models to assess the probability of proper movement execution; create synthetic learning experience; and warm-start the initial-state search process.

    \item \textbf{Extend rapid parallel skill acquisition to dynamic environments}. Behavioral skills can be challenged under rapid environmental changes (e.g. transitioning between water v.s. ground environments). Quickly re-calibrating skills in response to environmental changes ensures robust performance across diverse conditions. 
    
    \item \textbf{Transfer learned swarm behaviors among individuals in the population}. The more abstract a learned behavior, the more generalizable it is among different individuals. The abstract level of swarm control enables transferability of learned behaviors to significantly enhance learning performance. We can consider skill transfer between individuals.

    \item \textbf{Integrate experiences during learning in the evolutionary process}. Lamarckian evolution, i.e. inheriting experiences during your parents' life, should be carefully designed as most learned behaviors are too robot specific. Nevertheless, I expect certain `general knowledge' to be transferable between generation, thus evolvable. For example, hyper-parameter of learners (neural network design; learning rates; ets.), abstract level control (group level coordination; sensory fusion), and CPG reservoirs (evolve reservoir weights between generations). 
\end{enumerate}

\noindent From a higher-level perspective, I am convinced that ER is capable of creating the robots of tomorrow. The personalization of chat bots demonstrates the ability of artificial intelligence to enhance \textit{virtual robots} that are better adapted to their \textit{virtual environment} (the human with whom it interacts). ER is a highly capable method for \textit{physical robots} to better adapt to their \textit{physical environment}. Unfortunately, at present, the field is rather limited in terms of real-world applications. The EA, as a key ingredient at its core, provides a powerful tool that can `optimize anything with minimal prior knowledge'. This might have led to a limited interest in adapting technical expertise, and testing developed methods outside of the simulator. The assumption that `successful algorithms in simulation will work in reality' has been challenged by the reality gap. The next frontier lies in the application of ER for real robots. In the future, when tasks grow more complex, it might be more time-efficient to use ER in fully physical systems rather than bridging this reality gap. Instances where this could be the case are robots with soft materials, for hard-to-simulate environments like underwater and sandy terrains, or chaotic/complex multi-robot systems. Such applications are difficult to design for in software and are highly challenging for human engineering as well. In this context, I believe that ER can present a highly valuable solution to design robots of tomorrow.
\label{PageNum}

\def\PageNum{\getpagerefnumber{PageNum}}
\def\ChapNum{\thechapter}
\cleardoublepage
\newpage\null\cleardoublepage\newpage


\bibliographystyle{chicago}
\bibliography{merged}
\newpage
\pagestyle{fancy}
\fancyhead{}




\printglossary 

\end{document}







\vspace{-3em}
\section{Introduction}\label{sec1}
Basic locomotion primitives, such as walking and turning, are a prerequisite for more complex behavior in legged robots (e.g., foraging or environmental monitoring). However, obtaining controllers for such skills is a very challenging task that requires many iterations in design. A popular approach is the use of bio-inspired Central Pattern Generators (CPGs) that are modeled after neural circuits in the spinal cord of various animals \cite{marder2001central}. These (artificial) neural networks encode continuous-time signals that are smooth, (quasi-)periodic, and parameterized by only a few variables, which makes them attractive for learning, as locomotion requires such behavior \cite{bucher2015central}. CPGs can be interconnected to form a complex network of CPGs that produce a more complex signal, i.e. a CPG-network. 

\begin{figure}[ht!]
\captionsetup{width=.95\textwidth}
    \centering
    \includegraphics[width=0.99\textwidth]{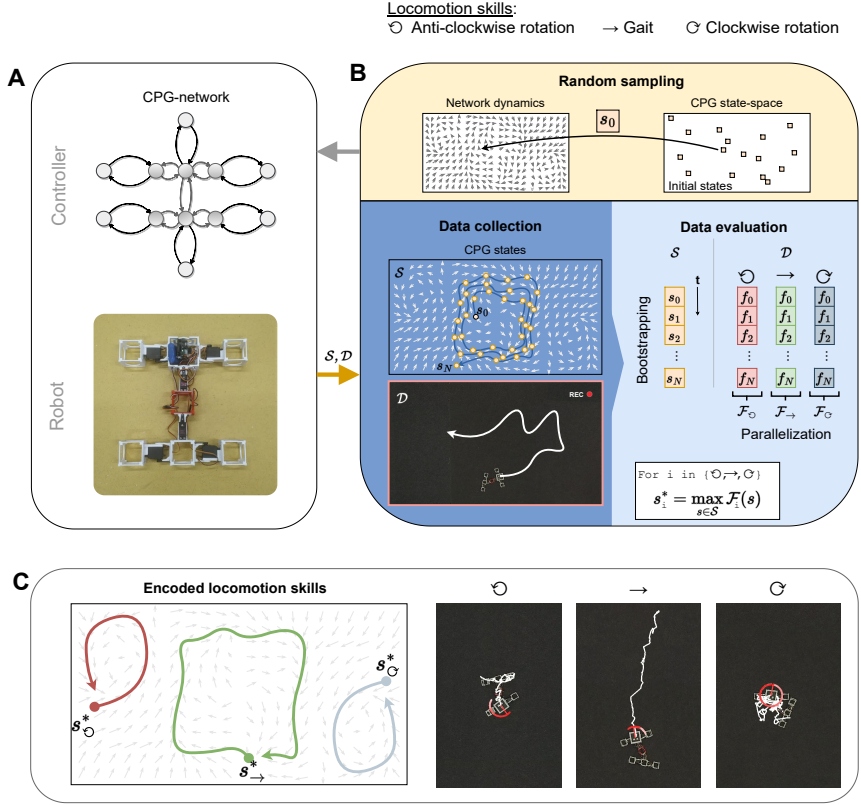}
    \caption{\footnotesize Initial State Optimization on Central Pattern Generators (CPGs) quickly finds different locomotion skills (Gait: \Gait; Clockwise rotation: \RotR; Anti-clockwise rotation: \RotL). \textbf{A.} Our robot is initialized with a network of interconnected CPGs in which we freeze all weights. \textbf{B.} \textit{Top}, we sample a random initial state of the CPG-network and produce a time-series (based on its dynamics) that provides motor commands to our robot. \textbf{B.} \textit{Bottom-left}, we collect data of the intermediate CPG states ($\mathcal{S}$) and the corresponding robot behavior ($\mathcal{D}$). \textbf{B.} \textit{Bottom-right}, we consider the intermediate CPG states as additional `initial state' samples (\textit{bootstrapping}) and assess the fitness values for multiple skills at the same time (\textit{parallelization}). \textbf{C.} In the end, we re-test the best initial state sample for each skill \{$\bm{s^{*}_{\RotL}}, \bm{s^{*}_{\Gait}}, \bm{s^{*}_{\RotR}}$\} encoded in the same CPG-network. The real-world behavior is shown in three pictures on the right with the robot position trajectory as a white line, and its orientation in red fading on a (semi-)circle.}
    \label{fig:method4}
\end{figure}

In biology, neuroplasticity is a key element in the optimization of CPG-networks \cite{pearson2000neural}. This form of adaptation is a slow indirect process of operant conditioning to improve its overall function. In robotics, a similar method of weight optimization (WO) is the standard approach to training CPG-networks, where the connection strength is changed to improve locomotion \cite{ijspeert2007swimming, sproewitz2008learning, Ijspeert2013, minati2018versatile, korkmaz2019locomotion, campanaro2021cpg, manoonpong2021insect, ruppert2022learning, thor2022versatile, rudin2022learning, 9932888, li2023development, szorkovsky2023towards}. In this paper, we present a novel way of optimizing a CPG-network, where rather than focusing on conventional WO, we instead focus on improving the initial state of the network (Initial State Optimization, ISO). \autoref{fig:method4} provides a general overview of our method.

ISO is a more direct process than WO in which we learn how to correctly initiate the network-state that produces the right behavior. The CPG-network is not adapted through WO, and only function as a behavioral encoder with several `motor primitives'. Our approach is based on the observation that the CPGs in spinal centres can produce multiple locomotion patterns that are encoded within the same network \cite{pearson2000neural}. In animals, it is well documented that the motor cortex and/or peripheral stimuli can start locomotion patterns with signals to the spinal cord that initialize the CPGs \cite{grillner1985central} (although these signals are frequency encoded by a drive signal and not a single command). Initial states can be seen as the movement command for a specific (cyclical) motor pattern memorized by the higher-order brain centres (like the motor cortex). Here, our `start command' is an input signal that sets the CPG-network in the correct initial state. 

We provide a mathematical analysis of the difference between ISO and WO in Appendix \ref{secA1}. We show that, due to the dynamics of CPGs, WO suffers from many local optima caused by redundancies and nonlinearities in `weights search-space', which is not the case for ISO. This mathematical analysis also shows that our random CPG-networks contain infinitely many quasi-periodic cycles due to their elliptic fixed point. As a result, we suggest that instead of optimizing limit-cycle behaviors using WO, ISO should be considered to acquire semi-cyclical patterns for robot locomotion. 

The conclusions of our mathematical analysis are supported by additional exploratory experiments (Appendix \ref{secA3}) showing that ISO can obtain a performance similar to that of WO in a much shorter time frame (\autoref{fig:results_init}). Furthermore, we identify two additional practical tricks that can be utilized during ISO to speed up optimization (see \autoref{fig:method4}): 1) \textit{Bootstrapping}, which comprises evaluating intermediate network states as additional potential initial state samples (an in-depth parameter analysis can be found in Appendix \ref{secA8}); 2) \textit{Parallelization}, which involves assessing different locomotion skills in parallel (we show the efficacy of this method up to 100 `skills' in parallel Appendix \ref{secA9}). An video explanation of the method: supplementary videos 1; and can also be found here \url{https://fudavd.github.io/multi-skill-learning/}. Using ISO, our robots only need a few trials to learn multiple locomotion skills in parallel. 

We demonstrate ISO on six different robots that learn the following three locomotion skills: gait, clockwise rotation, and anti-clockwise rotation (for a total of 18 learning tasks). Our experimental validation is two-fold. First, in simulation, we compare ISO learning performance with WO and a SOTA baseline reinforcement learning method, Proximal Policy Optimization (PPO, \cite{schulman2017proximal, shahid2022continuous, handa2023dextreme}). The results show that our method outperforms PPO and achieves 50\% of the WO performance in less than 5\% of the time, which corresponds to our findings in preliminary experiments (Appendix \ref{secA6}). Second, we perform real-world ISO experiments on our set of robots and compare its real-world performance with WO transferred from simulation to real. To this end, we transfer the best controllers out of 30 repetitions to present the strongest and most robust solution found by WO. The results show that WO suffers significantly from the sim-to-real transfer and is outperformed by ISO on 11/18 tasks ($p<0.001$) in significantly shorter optimization time.
A more detailed layout of our CPG-network, the description of the system, the ISO and WO method, and experimental setup can be found in \autoref{sec4}. The Appendix provides both the mathematical derivation of our propositions and the results of our preliminary experiments and extended results. 

Our conclusions are supported by exploratory experiments on a range of random time-series (Appendix \ref{secA3}) showing that ISO can obtain a performance similar to that of WO in a much shorter time frame for any target behavior (\autoref{fig:results_init}). Finally, we show the \textit{model-agnosticity} of our method by learning 6 locomotion skills with ISO on a test suite of 20 different robots (Appendix \ref{secA10}).

\section{Results}\label{sec2}
\begin{figure}[htpb]
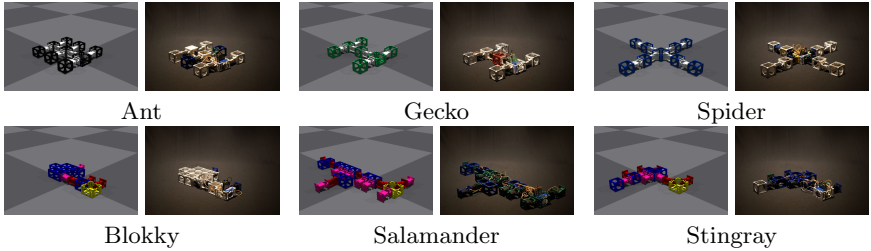

\centering
\begin{minipage}[c]{\textwidth}
\small
        \begin{minipage}[t]{\textwidth}
            \begin{minipage}[t]{0.32\textwidth}
                \centering
                \includegraphics[width=0.48\textwidth ]{figures/robots/sim/ant.png} \includegraphics[width=0.48\textwidth ]{figures/robots/real/ant.png}\\
                Ant
            \end{minipage}
            \hfill
            \begin{minipage}[t]{0.32\textwidth}
                \centering
                \includegraphics[width=0.48\textwidth]{figures/robots/sim/gecko.png} \includegraphics[clip=true, width=0.48\textwidth]{figures/robots/real/gecko.png}
                Gecko
            \end{minipage}
            \hfill
            \begin{minipage}[t]{0.32\textwidth}
                \centering
                \includegraphics[width=0.48\textwidth ]{figures/robots/sim/spider.png} \includegraphics[clip=true, width=0.48\textwidth]{figures/robots/real/spider.png}
                Spider
            \end{minipage}
        \end{minipage}
        \vspace{1em}
        \begin{minipage}[c]{\linewidth}
            \begin{minipage}[t]{0.32\textwidth}
                \centering
                \includegraphics[width=0.48\textwidth]{figures/robots/sim/blokky.png} \includegraphics[clip=true, width=0.48\textwidth]{figures/robots/real/blokky.png}
                Blokky
            \end{minipage}
            \hfill
            \begin{minipage}[t]{0.32\textwidth}
                \centering
                \includegraphics[width=0.48\textwidth]{figures/robots/sim/salamander.png} \includegraphics[clip=true, width=0.48\textwidth]{figures/robots/real/salamander.png}
                Salamander
            \end{minipage}
            \hfill
            \begin{minipage}[t]{0.32\textwidth}
                \centering
                \includegraphics[width=0.48\textwidth]{figures/robots/sim/stingray.png} \includegraphics[clip=true, width=0.48\textwidth]{figures/robots/real/stingray.png}
                Stingray
            \end{minipage}
        \end{minipage}
           \caption{Test suite of six robots as rendered in the simulator (left) and their physical twin (right).}\label{fig:robots}
\end{minipage}
\end{figure}

We compared WO with ISO by learning three different movement skills on six different modular legged robots (see \autoref{fig:robots}; more technical details in \autoref{tab:robots_spec}), thus resulting in 18 different learning tasks. The three locomotion skills are defined as: (1) Gait (\Gait), absolute planar displacement speed in centimetres per seconds [\unit[]{\cm\per\s}]; \textit{2 \& 3) clockwise} and \textit{anti-clockwise rotation} (\RotR/\RotL), planar rotation in radians per seconds [\unit[]{\radian\per\s}]. First, the performance of ISO and WO is compared with PPO in simulation (parameters are detailed in \autoref{tab:parameters_opt}); then ISO and WO are compared in the real world (without PPO since it performed worst in the simulation experiments).

\subsection{Simulation experiment}
We optimize our robots in simulation for \unit[300]{\min} for each skill using ISO, WO, and compare their performance with a PPO baseline \cite{schulman2017proximal}. We use cleanRL's continuous action PPO algorithm for Isaac Gym with default parameters \cite{huang2022cleanrl}. For WO, we use a SOTA differential evolution algorithm (RevDE, \cite{tomczak2020differential}) in which every trial lasts for one minute ($t_{trial}=\unit[60]{\s}$, $N_{trials}=300$). Trial-time is chosen to differentiate between transient behavior (starting from standing still to moving) and steady state behavior of the locomotion skill. Too short of an evaluation time focuses more on transient behavior which will be exploited by the optimizer (e.g. the robot may only fall sideways for high rotation speed). Random search is used for ISO, in which we halve the number of trials and double the trial time to increase the effectiveness of bootstrapping based on our hyper parameter sweep (see Appendix \ref{secA8}) ($t_{trial}=\unit[120]{\s}$, $N_{trials}=150$). This results in a 600 fold increase in `initial state' samples for each trial (using a rolling window evaluation length $t_{eval}=\unit[60]{\s}$)). It is important to note here that per robot WO requires three times as much time to learn all three skills with respect to ISO, due to the parallelization trick (\unit[900]{\min} in WO vs. \unit[300]{\min} in ISO). A summary of these parameters is presented in \autoref{tab:parameters_exp4}. 

\begin{figure}[htpb]
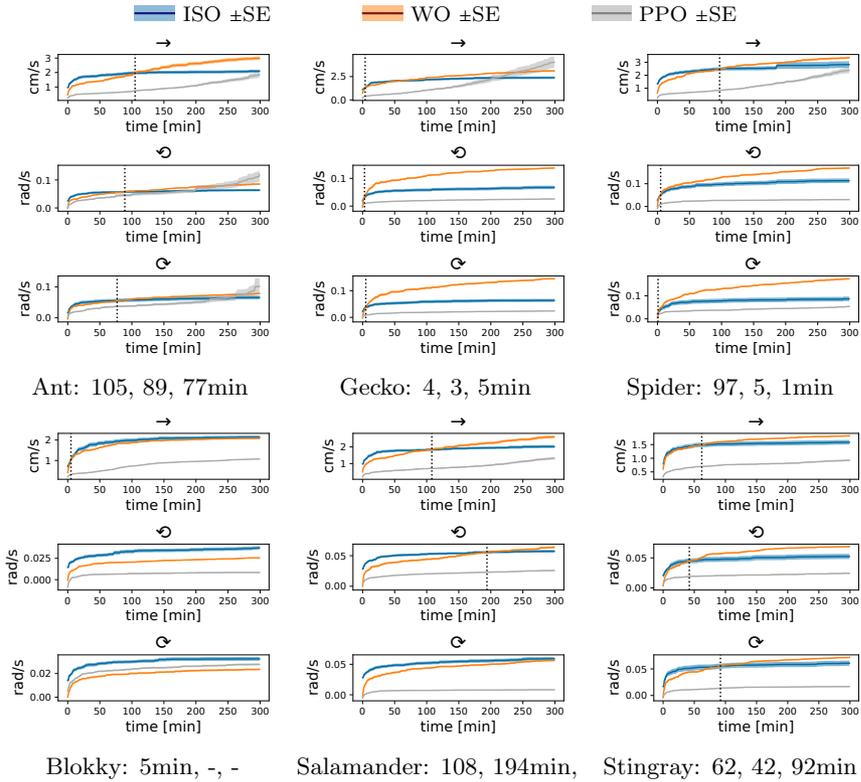

\centering
\begin{minipage}[c]{0.7\textwidth}
\centering
\footnotesize
    \includegraphics[height=0.2cm]{figures/Results/tteq_curve/legend_state0.pdf}
    ISO $\pm\textrm{\footnotesize SE}$
    \hspace{1cm}
    \includegraphics[height=0.2cm]{figures/Results/tteq_curve/legend_weight.pdf}
    WO $\pm\textrm{\footnotesize SE}$
    \hspace{1cm}
    \includegraphics[height=0.2cm]{figures/Results/tteq_curve/legend_ppo.pdf}
    PPO $\pm\textrm{\footnotesize SE}$
\end{minipage}
\begin{minipage}[c]{\textwidth}
    \small
        \begin{minipage}[t]{0.32\textwidth}
            \centering
            \includegraphics[width=\textwidth]{figures/Results/tteq_curve/ant.pdf}
            Ant: 105, 89, 77\unit[]{\min}
        \end{minipage}
        \hfill
        \begin{minipage}[t]{0.32\textwidth}
            \centering
            \includegraphics[width=\textwidth]{figures/Results/tteq_curve/gecko.pdf}
            Gecko: 4, 3, 5\unit[]{\min}
        \end{minipage}
        \hfill
        \begin{minipage}[t]{0.32\textwidth}
            \centering
            \includegraphics[width=\textwidth]{figures/Results/tteq_curve/spider.pdf}
            Spider: 97, 5, 1\unit[]{\min}
        \end{minipage}
    \end{minipage}
\begin{minipage}[c]{\textwidth}
    \small
        \begin{minipage}[t]{0.32\textwidth}
            \centering
            \includegraphics[width=\textwidth]{figures/Results/tteq_curve/blokky.pdf}
            Blokky: 5\unit[]{\min}, -, -
        \end{minipage}
        \hfill
        \begin{minipage}[t]{0.32\textwidth}
            \centering
            \includegraphics[width=\textwidth]{figures/Results/tteq_curve/salamander.pdf}
            Salamander: 108, 194\unit[]{\min}, -
        \end{minipage}
        \hfill
        \begin{minipage}[t]{0.32\textwidth}
            \centering
            \includegraphics[width=\textwidth]{figures/Results/tteq_curve/stingray.pdf}
            Stingray: 62, 42, 92\unit[]{\min}
        \end{minipage}
\end{minipage}
\captionsetup{width=.95\textwidth}\footnotesize
\caption{\label{fig:res_tteq_curve} Initial State optimization (ISO, blue curves) outperforms  both PPO (gray) and Weight optimization (WO, orange). WO is outperformed in the first \unit[64]{\min} on average. Learning curves show Mean Best Fitness ($N_{rep}=30\pm\mathrm{SE}$) for gait (\Gait) and anti-/clockwise rotation (\RotL, \RotR) skills for the six different robots in simulation. The vertical dotted lines indicate the first cross-over point after which WO begins to outperform ISO, i.e., Time Till Equal Quality (TTEQ, higher is better). The specific TTEQ for each skill is presented in each sub-caption (ordered in: \Gait, \RotL, \RotR), while no cross-over point is denoted with a dash (-).}
\end{figure}

The learning curves (\autoref{fig:res_tteq_curve}) show the Mean Best Fitness (MBF, i.e. the highest fitness so far in time, see \autoref{alg:mbf}) averaged over 30 repetitions for each experiment $\pm$ Standard Error ($N_{rep}=30$, $\pm$SE), which is a metric that allows for the comparison of different learners \cite{van2023comparing}. For each morphology, we also report the Time-to-Equal Quality (TTEQ) for each skill, which indicates the first time that the MBF ISO is equal to the MBF WO (see the subcaption of \autoref{alg:tteq}, higher TTEQ is better). The results show an overall better performance for ISO at the start of the learning process, which shows a more efficient skill acquisition than WO or PPO. This outperformance is consistent throughout the entire optimization process for Blokky \Gait, \RotL and \RotR and Salamander \RotR, which shows in the absence of TTEQ values. In congruence with the preliminary experiments, the ISO sharply outperforms WO at the beginning before plateauing for the remainder of the run (\autoref{fig:results_init}). PPO overall tends to be the most inefficient. This is unsurprising, as the number of allowed time-steps is several orders of magnitude smaller with respect to most PPO applications (180,000 vs 10,000,000 in the original PPO paper \cite{schulman2017proximal}).

These findings are unsurprising since Appendix \ref{secA2} reveals the linearity of ISO vs. the complex nonlinear coupling of WO. The parallelization trick in ISO makes learning efficiency independent to the number of skills learned, shown up to a 100 `skills' in parallel \autoref{fig:APP_n_targets}. This means that ISO has constant complexity $\mathcal{O}(k)$, even though \RotL, and \RotR are counter opposite behaviors. Furthermore, even with random CPG-networks, the chances of finding a good initial state are relatively high (we find good performance quickly with random sampling and bootstrapping). Already, in the first trials, ISO can immediately identify better sub-samples than WO, thereby outperforming WO in efficiency. This outperformance is maintained for \unit[64]{\min} on average (mean TTEQ). Here, 64 minutes underestimate the true mean TTEQ, as we omitted instances where WO never crossed ISO (Blokky \RotL and \RotR; Salamander \RotR).

\begin{figure}[htpb]
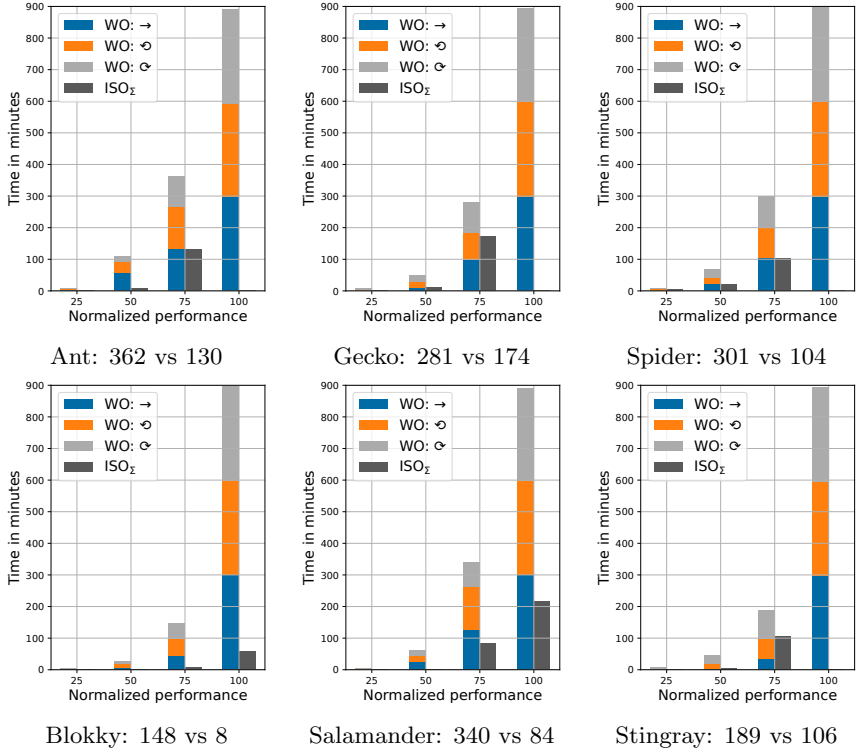

\centering
\begin{minipage}[c]{\textwidth}
\small
        \begin{minipage}[t]{0.32\textwidth}
            \centering
            \includegraphics[width=\textwidth]{figures/Results/perf_bar/ant_perf.pdf}
            Ant: 362 vs 130
        \end{minipage}
        \hfill
        \begin{minipage}[t]{0.32\textwidth}
            \centering
            \includegraphics[width=\textwidth]{figures/Results/perf_bar/gecko_perf.pdf}
            Gecko: 281 vs 174
        \end{minipage}
        \hfill
        \begin{minipage}[t]{0.32\textwidth}
            \centering
            \includegraphics[width=\textwidth]{figures/Results/perf_bar/spider_perf.pdf}
            Spider: 301 vs 104
        \end{minipage}
    \end{minipage}
\begin{minipage}[c]{\textwidth}
\small
    \begin{minipage}[t]{0.32\textwidth}
        \centering
        \includegraphics[width=\textwidth]{figures/Results/perf_bar/blokky_perf.pdf}
        Blokky: 148 vs 8
    \end{minipage}
    \hfill
    \begin{minipage}[t]{0.32\textwidth}
        \centering
        \includegraphics[width=\textwidth]{figures/Results/perf_bar/salamander_perf.pdf}
        Salamander: 340 vs 84
    \end{minipage}
    \hfill
    \begin{minipage}[t]{0.32\textwidth}
        \centering
        \includegraphics[width=\textwidth]{figures/Results/perf_bar/stingray_perf.pdf}
        Stingray: 189 vs 106
    \end{minipage}
\end{minipage}
\captionsetup{width=.95\textwidth}\footnotesize
\caption{\label{fig:res_norm_perf} Initial State optimization (ISO$_\Sigma$) obtains 50\% performance of Weight optimization (WO$_{\Gait}$, WO$_{\RotL}$, WO$_{\RotR}$) in less than 5\% of the time. The bars indicate the time needed to obtain normalized performance quality with respect to WO ($N_{rep}=30$). In the sub-caption we specify the total time needed to attain 75\% relative performance in minutes (WO time vs. ISO time).}
\end{figure}

Comparing the MBF learning curves for each skill provides a distorted overview of the efficiency of ISO with respect to WO, because we pit one parallel run of the former against three sequential runs of the latter. In \autoref{fig:res_norm_perf}, we show the simulated time for different normalised performance thresholds at [$25, 50, 75, 100$] percent performance with respect to WO. Normalization is based on the final WO MBF values (i.e., performance at \unit[300]{\min}), for each skill for all six robots. For WO, we calculate the relative performance for each skill and show the total simulated time as a stacked bar ($WO_\Gait+WO_\RotL+WO_\RotR$, with a maximum time of \unit[900]{\min}). For ISO, we calculate the combined relative performance (that is, the total sum of relative performance over all three skills) as a dark grey bar ($ISO_\Sigma$, see \autoref{alg:norm_perf}). $ISO_\Sigma$ relative performance provides a fair comparison, as the learning in ISO is parallelized. In the subcaptions of \autoref{fig:res_norm_perf} we also report the specific time needed to obtain 75\% relative performance for each robot.

With respect to normalized performance, it becomes clear that ISO is extremely fast with respect to WO. On average, it takes WO \unit[270]{\min} to reach 75\% performance vs. \unit[101]{\min} for ISO, with all morphologies taking less than \unit[200]{\min} in the latter. At the end of the optimization, four of the six morphologies (Ant, Gecko, Spider, and Stingray) never reached the same level of performance as WO, although the final performances of the Blokky and Salamander morphologies exceeded WO up to 140\%. Furthermore, we found that ISO can already achieve 50\% relative performance (with respect to WO MBF) within the first 10 minutes (i.e., 5 trials) for all but the Spider morphology: Ant \unit[9.33]{\min}, Gecko \unit[10]{\min}, Spider \unit[21]{\min}, Blokky \unit[2]{\min}, Salamander \unit[2.66]{\min}, Stingray \unit[6]{\min}. Resultantly, we achieve 50\% of WO performance in less than 5\% of the time, using ISO.

\subsection{Real world experiment}
\begin{figure}[htpb]
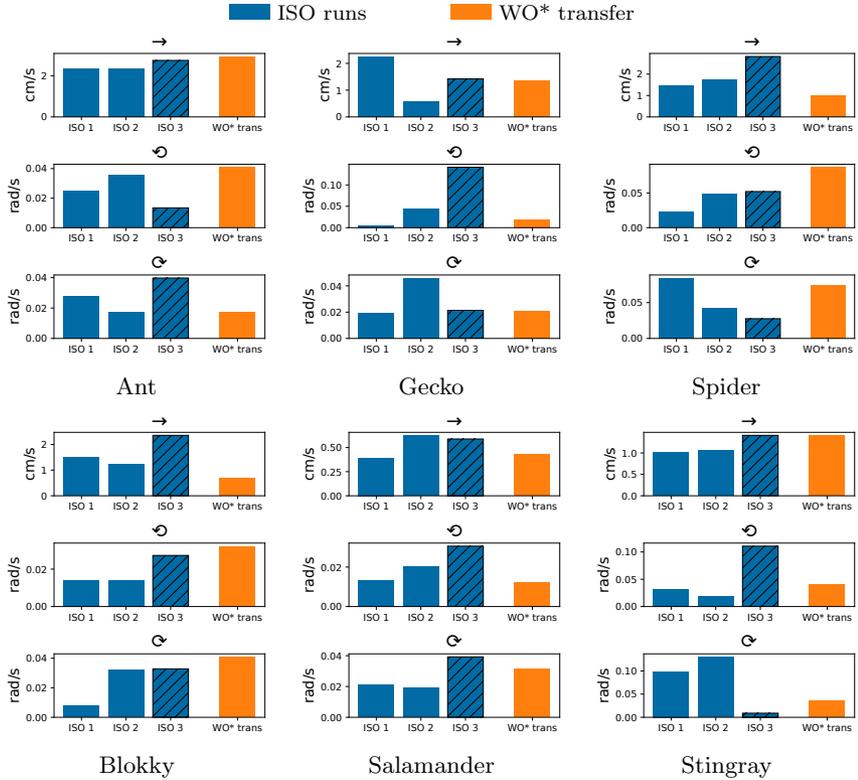

\centering
\begin{minipage}[c]{0.75\textwidth}
\centering
\footnotesize

    \includegraphics[height=0.2cm]{figures/Results/real/legend_state0_real.pdf}
    ISO runs
    \hspace{3em}
    \includegraphics[height=0.2cm]{figures/Results/real/legend_weight_real.pdf}
    WO* transfer
    
\end{minipage}
\begin{minipage}[c]{\textwidth}
    \small
        \begin{minipage}[t]{0.32\textwidth}
            \centering
            \includegraphics[width=\textwidth]{figures/Results/real/ant_real.pdf}
            Ant
        \end{minipage}
        \hfill
        \begin{minipage}[t]{0.32\textwidth}
            \centering
            \includegraphics[width=\textwidth]{figures/Results/real/gecko_real.pdf}
            Gecko
        \end{minipage}
        \hfill
        \begin{minipage}[t]{0.32\textwidth}
            \centering
            \includegraphics[width=\textwidth]{figures/Results/real/spider_real.pdf}
            Spider
        \end{minipage}
    \end{minipage}
\begin{minipage}[c]{\textwidth}
    \small
    \begin{minipage}[t]{0.32\textwidth}
        \centering
        \includegraphics[width=\textwidth]{figures/Results/real/blokky_real.pdf}
        Blokky
    \end{minipage}
    \hfill
    \begin{minipage}[t]{0.32\textwidth}
        \centering
        \includegraphics[width=\textwidth]{figures/Results/real/salamander_real.pdf}
        Salamander
    \end{minipage}
    \hfill
    \begin{minipage}[t]{0.32\textwidth}
        \centering
        \includegraphics[width=\textwidth]{figures/Results/real/stingray_real.pdf}
        Stingray
    \end{minipage}
\end{minipage}
\captionsetup{width=.95\textwidth}\footnotesize
\caption{\label{fig:res_real} Initial State optimization (ISO, blue) on real robots produces similar performance to the sim-to-real transfer of the best out of the 30 runs from Weight Optimization (WO*, orange). The bars show the performance of each physical robot concerning three different skills: gait (\Gait) and anti-/clockwise (\RotL, \RotR). The striped bars indicate the overall best ISO run (based on the total rank over all the skills).}
\end{figure}

In the real-world experiment, we test our ISO method on the physical twin of our simulated robots. Here, we restrict ourselves to only \unit[15]{\min} of total experiment time for each robot, including re-evaluation of the acquired skill. By including the overhead, we can therefore assess five trials of \unit[2]{\min}, and re-test each acquired skill for \unit[1]{\min}. Based on the results of the simulation experiments, this extremely tight budget is sufficient to obtain good performance. Moreover, the \unit[15]{\min} limit is an extremely efficient time for real-world experiments and compares well (as a benchmark time) with respect to other state-of-the-art (SOTA) learning algorithms (\unit[12]{}-\unit[20]{\min}\cite{thor2022versatile, rudin2022learning, smith2022walk}). We compare the performance of the real ISO experiments with the overall best WO* controller for each skill (in this instance the star denotes the best out of 30 runs). This method is also known as offline learning with sim-to-real transfer \cite{hofer2021sim2real}. The results of the real experiments are presented in \autoref{fig:res_real}. In the real world, WO* performs similar to ISO, which equates to a 57\% reduction as a result of the sim-to-real gap. In the end, ISO outperforms WO* in 11 out of 18 skills, based on the overall best run (ranked over all skills within a morphology, shown as striped bars). We want to emphasise that we present the strongest and most robust version of WO, where each skill was optimized for 30 different runs separately for a total of (vs. 1 single run of all skills in parallel). Therefore, we argue that real-world ISO is preferable on the grounds that it circumvents the sim-to-real problem, requires less computational resources (local cluster vs. Raspberry Pi), and is more scalable with respect to additional skills (parallelization means constant time complexity $\mathcal{O}(k)$).

ISO can learn all three skills on all six robots in under \unit[15]{\min}. We found a small average performance drop with ISO in real robots compared to the ISO simulation experiment. This means that the real-world ISO performance corresponds to the predicted ISO performance, which, as such, indicates that ISO is robust in performance regardless of simulation or real-world. The overall best real-world ISO runs (based on total rank for each skill within a morphology) show a 54\% performance relative to final WO MBF values (compared to 50\% in simulation for the same amount of time). If we compare these runs with WO* we see that we outperform eleven out of eighteen learning tasks. The behavior of these runs are shown in the Appendix \autoref{fig:App_real_best}. 
If we zoom in closer, we see that ISO has a random bias for rotation in a certain direction for a single run (an effect that smoothened out when we averaged the 30 runs in simulation). The random CPG-weights might induce a bias for a given rotational skill, as there are only a limited set of frequencies available in the CPG-network defined by the purely imaginary eigenvalues in its canonical form (shown in Appendix \ref{secA1}). Slow weight adaptation might be able to resolve this problem by changing the frequency content of the CPG-network.

Optimizing in simulation is inevitably vulnerable to the sim-to-real gap \cite{jakobi1995noise, mouret201720, ligot2020simulation}, due to model approximations and simulations errors being exploited and overfitted by the optimizer. When transferring WO* on the real robots, we observe on average a 57\% drop in performance compared to simulation, while ISO did not drop significantly in performance (-2\%). The specific sim-to-real gap for each morphology and skill is presented in the Appendix \autoref{tab:APP_real}. These results show that the WO* controller is badly impacted by the sim-to-real gap, while ISO performance is robust in both conditions. In the end \autoref{fig:res_real} shows that we get comparative performance with ISO in the same wall-clock time as sim-to-real WO* ($\sim$\unit[11]{\min} ISO vs. $\sim$\unit[4]{\min} for each skill with sim-to-real WO*). It should be noted that this comparison is extremely favorable toward WO*, as we took the best controller over 30 repetitions (which took about six hours in total per robot). Furthermore, in our experiments, we only provided three skills. The wall-clock time of WO increases linearly with respect to the number of skills, while ISO remains constant (also demonstrated in Appendix \ref{secA9}). Real-world ISO is thus competitive with sim-to-real even in the most favorable circumstances for WO. 

\subsection{Preliminary feedback control experiments}
Our CPG-controller allows the perturbation of hidden ($y$-neuron) states that are not directly linked to the signals sent to the servomotors. This will provide smooth behavioral changes as our CPG defines behavior in velocities. Using this technique, we extend our controller with heading feedback for a targeted locomotion task on our spider robot in simulation (specific details on the design are presented in Appendix \ref{secA12}). The proposed method robustly reached the target in a random location 4m from the origin for all trials ($N=60$) under 5 minutes.

\section{Discussion}\label{sec3}
An advantage of CPGs in robotics is that their behavior is relatively stable and is encoded cheaply using few parameters. Instead of learning a whole time-series of motor input or a policy of actions at every point in state-space, we now only need to describe the CPG-network to realize useful behavior. 
Our mathematical analysis (in Appendix \ref{secA2}) sheds light on the effects of optimizing the different design variables of a CPG-network: network structure, connection strength, and initial state. Based on our findings, we developed a simple and efficient method for learning multiple skills using the initial state.

In doing so, we provide strong evidence for the potential advantage of learning skills through ISO. The results of the simulation experiments showed that ISO can achieve 50\% of WO performance in less than 5\% of the simulated time, highlighting its outperformance in terms of efficiency. The real robot experiments show the advantage of ISO for online multi-skill learning vs. offline WO and transferring the best controllers. We consider task complexity not only on the final behavior (moving from A to B, or rotating), but also on the difficulty to learn the behavior (DofF, amount of prior knowledge, a-symmetry in morphology and control etc.). Compared to other SOTA learning algorithms (\unit[12]{}-\unit[20]{\min} \cite{thor2022versatile, rudin2022learning, smith2022walk}), we perform similarly in complex locomotion tasks. In addition, we achieve this learning performance without the use of a simulator, limited a priori knowledge, and a predefined model. Our method is the first of its kind, both in terms of speed (the number of skills that it learns in a short amount of time, also seen in Appendix \ref{secA9}) and generality (model-free control tested on several different morphologies). 

The automated design of our network structure outlined in \autoref{sec5} implies only local synchronization of the oscillator frequencies. This is unlike most work on CPGs where a central frequency strictly defines a limit-cycle phase to coordinate periodic movements in time \cite{ijspeert2007swimming, sproewitz2008learning, Ijspeert2013, minati2018versatile, korkmaz2019locomotion, campanaro2021cpg, manoonpong2021insect, ruppert2022learning, thor2022versatile, rudin2022learning, 9932888, li2023development, szorkovsky2023towards}. Our method does not require such an `internal clock' and is capable of learning quasi-periodic movements in much shorter time-scales, while it can also produce non-periodic control. In addition, the formal basis for the strength of our method is detailed in Appendix \ref{secA1}. We show that, irrespective of random weights, the SO2-oscillator network exhibits marginally stable behavior due to their elliptic fixed point, which proves the existence of (infinitely) many quasi-periodic patterns. The expressiveness of ISO is demonstrated by estimating 100 randomly generated time-series in parallel (Appendix \ref{secA9}) and learning 6 skills in 20 robots (Appendix \ref{secA10}). 

The importance of initial states is not limited to the specific application of SO2-oscillators analyzed in this work. For any optimization task (irrespective of the controller/optimization strategy), the initial state of the robot must be in a good region of the state space. If this is not the case, then the optimizer must solve two tasks: 1) find the optimal (cyclical) behavior; 2) move from the bad initial state towards the region of optimal behavior (a transient pattern). \cite{hwangbo2019learning} reported that in their RL algorithm, the initial state distribution was designed and tuned separately for each task, which took a specialist around two days to complete. The sensitivity of certain dynamical systems to the initial state has already been described as the \textit{butterfly effect} in chaos theory. Thus, in any systems with a recurrent structure (ODEs, LSTMs, reinforment learning, recurrent neural networks, other forms of CPGs, reservoir computing, etc.) the behavior of a controller is extremely sensitive to its initial condition. Nevertheless, we found that most work does not consider tuning/optimizing its initial state (if reported at all, initial states are often set to 0 or 1). 

Learning robot control is a topic of wide interest, with current methods relying on specific robot models and/or massive computing to learn models and behaviors in simulation \cite{rudin2022learning}. Such learning methods have trouble adapting or learning new behaviors as integrating/adapting learned skills often require sequential learning. This involves problems due to catastrophic forgetting (a problem that arises in lifelong/(deep) reinforcement learning \cite{parisi2019continual, kirkpatrick2017overcoming}), or the curse of dimensionality (e.g., MAP-elites, \cite{cully2015robots, mouret2015illuminating}), and overall time constraints \cite{hwangbo2019learning, ruppert2022learning}. Parallelization circumvents catastrophic forgetting and results in constant time complexity $\mathcal{O}(k)$ to learn multiple skills. In addition, massive parallel computing continues to be plagued by the sim-to-real gap, which we can circumvent by not relying on simulation at all \cite{jakobi1995noise}. Other model-free controllers are several magnitudes slower than our ISO method \cite{mahmood2018benchmarking}. Similar other CPG-based controllers rely on human expertise to design networks with precise hyper-parameters\cite{goldschmidt2017neurocomputational, rudin2022learning}, and focus on a specific robot morphology \cite{smith2022walk, 9932888}. In contrast, our method is truly model-free and straightforward in that it defines a network structure based on morphology with random weights. Finally, the process of storing and selecting patterns shares some resemblance with reservoir computing \cite{lukovsevivcius2009reservoir} or the use of motor primitives \cite{9655714}. However, ISO has the added advantage that it learns multiple skills in parallel and does not suffer from catastrophic forgetting. 

The generalizability of ISO shows in the number of robots tested and its efficacy in learning multiple skills/behaviors in parallel (Appendix \ref{secA10}). Real-world testing was possible due to the time-efficiency of ISO. The design space of modular robots (the set of all possible morphologies) is large and incredibly diverse, insofar as it contains robots comprising various shapes, a-symmetrical bodies, several or many limbs with potentially varying lengths. Successfully learning multiple skills in parallel on our wide variety of morphologies can be attributed to the model-free nature of the CPGs. Alongside this, we created our CPG-network structure based on simple morphological rules that are not specific to our design and, as such, can be altered without any consequences for the algorithm.This shows the robustness of our method for different types of designs and tasks.

The open-loop skills obtained during ISO can be used for more complex tasks. We demonstrated this through higher-level control in an additional targeted locomotion task (Appendix \ref{secA12}).
In the future, we can extend our work by modulating gait speed and heading direction during forward locomotion. For this, we propose only modulation of our ODE (Equation \ref{eq:As}) by scaling the weight matrix for frequency adaptations (on $\bm{A}$), and output state for amplitude control. Additionally, in previous work we have successfully implemented other forms of complex feedback control as an extension of CPG-networks: low-level Internal Model Control \cite{van2020effects}, smooth speed and heading \cite{lan2020learning}; which could be implemented for ISO which would require feasible adaptations. 

Furthermore, our learning framework has been originally motivated in the context of a robot evolution system, where morphologies and controllers evolve in tandem and "newborn" robots undergo individual learning to optimize their inherited brain to their body \cite{eiben2013triangle}. This requires learning methods that are robust and fast, i.e, work for "all" possible morphologies and need relatively few search steps/evaluations before being employed in the real world. By leveraging the additional bootstrapped samples in our optimizer, we can employ other data-driven techniques for this end, for example, surrogate models and MAP-elites/novelty-search \cite{allard2022hierarchical, nordmoen2020quality, le2020sample} help to increase learning speed and encourage diversity in solutions. Parallelized fast learning of modular behaviors can help in the design of these hierarchical control structures such as finite-state machines or behavioral trees \cite{francesca2014automode, allard2022hierarchical, 10.3389/frobt.2021.684304}.

Finally, we want to emphasize the perspective of the CPG-network as a behavioral encoder. CPGs as a behavioral reservoir provide a simple explanation for the multi-functionality of the spinal cord \cite{linden2022movement}. The behavioral encoder effectively stores various movement patterns that can be initiated with the appropriate `start' signal, which corresponds to the initial state. From such a perspective, we could speculate that ISO quickly learns network input signals that activate movement skills, whereas WO slowly fine-tunes network dynamics during life, and evolution optimizes network structure every generation. This idea is not too far-fetched given the fact that newborn babies exhibit predefined locomotion movement patterns when specific states trigger the walking reflex \cite{dominici2011locomotor}. Our work might be the first to shed light on these bio-inspired modes of optimization in CPGs over different time-scales.

\textbf{Conclusion:}
Learning multiple movement skills is an important challenge for legged robotics insofar as it is a prerequisite for viable complex behavior. Considering the specificity of each robot, complex/dynamic environments, recovery after damage, etc., we argue that there is an important need for model-free controllers that quickly learn. In this paper, we show that this is achievable using CPG-networks and ISO. ISO is preferable over WO in terms of time-efficiency and with respect to the sim-to-real problem. Future work should aim at improving the search method, further utilizing bootstrapping and parallelization, and implementing a more complex and higher-level of control overall.

\section{Methods}\label{sec4}
This section describes the CPG-network controller, optimization methods, and simulated/physical experiment and setup. This does not include any formal analysis of the optimization of the network itself. For this, we refer the reader to Appendix \ref{secA2}. 

\subsection{CPG-network controller}\label{sec5}
CPGs are pairs of artificial neurons that reciprocally inhibit and excite each other to produce oscillatory behavior \cite{bucher2015central}, and can be mathematically described as a set of Ordinary Differential Equations (ODEs) \cite{ijspeert2008central}. In extant literature, several different models for CPGs have been proposed \cite{ijspeert2007swimming, sproewitz2008learning, Ijspeert2013, minati2018versatile, campanaro2021cpg, pasandi2020programmable, manoonpong2021insect, ruppert2022learning, thor2022versatile, rudin2022learning, 9932888, li2023development}. 
Although some details may vary, in essence the oscillatory behavior within the CPG is invariably defined in terms of the connectivity between neurons and their initial state. We focus on an inter-connected network of SO2-oscillators \cite{pasemann2003so} whose dynamics are described by an anti-symmetric weight-matrix that results in (quasi-)periodic behavior. 

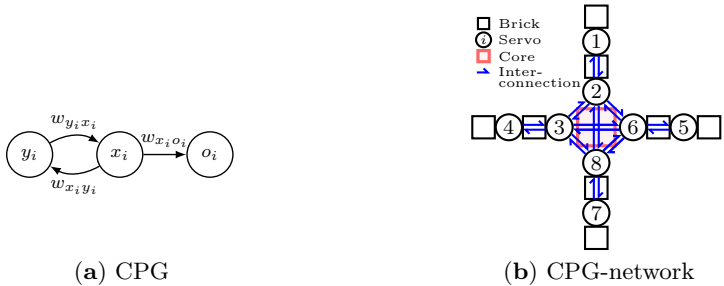
\begin{figure}[htpb]
\centering
\begin{minipage}[c]{\linewidth}\small
        \begin{minipage}[t]{0.45\linewidth}
            \centering
            \centering
\resizebox{0.6\linewidth}{!}{
    \begin {tikzpicture}
	    [-latex,
		auto,
		node distance =1.375cm and 1.375cm,
		on grid,
		semithick,
		state/.style = {circle,draw,inner sep=0pt,minimum width = 0.7 cm}]
	\node[state] (X) {$x_i$};
	\node[state] (Y) [left = of X] {$y_i$};
	\node[state] (Output) [right = of X] {$out_i$};
	\path (X) edge [bend left = 30] node [] {$w_{x_iy_i}$} (Y);
	\path (Y) edge [bend left = 30] node [] {$w_{y_ix_i}$} (X);
	\path (X) edge  node [] {$w_{x_io_i}$} (Output);
	\node (ghost) [below of = X]       {};
	\end {tikzpicture}
}\\
            (\textbf{a}) CPG
        \end{minipage}
        \hfill
        \begin{minipage}[t]{0.45\linewidth}
            \centering
            \centering
\resizebox{0.6\linewidth}{!}{%
\begin {tikzpicture}
	    [
		auto,
		node distance =0.3375cm and 0.3375cm,
		on grid,
		thick,
		state/.style = {circle,draw,inner sep=0pt, minimum width = 0.35 cm},
		box/.style={rectangle,draw,minimum width = 0.3 cm, minimum size=0.3 cm},
		core/.style={rectangle,draw=red!60,fill=red!5,very thick,minimum width = 0.5 cm, minimum size=0.5 cm}]

	\node[box] (B_l1) {};
	\node[state] (O4) [right = of B_l1] {$4$};
	\node[box] (B_l2) [right = of O4] {};
	\node[state] (O3) [right = of B_l2] {$3$};
	\node[core] (core) [right = 0.49cm of O3] {};
	\node[state] (O2) [above=0.48cm of core] {$2$};
	\node[box] (B_t2) [above=of O2] {};
	\node[state] (O1) [above=of B_t2] {$1$};
	\node[box] (B_t1) [above=of O1] {};
	\node[state] (O6) [right= 0.49cm of core] {$6$};
	\node[box] (B_r2) [right= of O6] {};
	\node[state] (O5) [right=of B_r2] {$5$};
	\node[box] (B_r1) [right=of O5] {};
	\node[state] (O8) [below=0.48cm of core] {$8$};
	\node[box] (B_b2) [below=of O8] {};
	\node[state] (O7) [below=of B_b2] {$7$};
	\node[box] (B_b1) [below=of O7] {};
	\path[draw=blue] (O4.10) to node [] {} (O3.170);
	\path[draw=blue] (O3.-170) to node [] {} (O4.-10);
	\path[draw=blue] (O6.10) to node [] {} (O5.170);
	\path[draw=blue] (O5.-170) to node [] {} (O6.-10);
	\path[draw=blue] (O2.100) to node [] {} (O1.-100);
	\path[draw=blue] (O1.-80) to node [] {} (O2.80);
	\path[draw=blue] (O7.100) to node [] {} (O8.-100);
	\path[draw=blue] (O8.-80) to node [] {} (O7.80);
	\path[draw=blue] (O3.50) to node [] {} (O2.-140);
	\path[draw=blue] (O2.-120) to node [] {} (O3.30);
	\path[draw=blue] (O3.10) to node [] {} (O6.170);
	\path[draw=blue] (O6.-170) to node [] {} (O3.-10);
	\path[draw=blue] (O3.10) to node [] {} (O6.170);
	\path[draw=blue] (O6.-170) to node [] {} (O3.-10);
	\path[draw=blue] (O6.150) to node [] {} (O2.-60);
	\path[draw=blue] (O2.-40) to node [] {} (O6.130);
	\path[draw=blue] (O2.-80) to node [] {} (O8.80);
	\path[draw=blue] (O8.100) to node [] {} (O2.-100);
	\path[draw=blue] (O6.-130) to node [] {} (O8.40);
	\path[draw=blue] (O8.60) to node [] {} (O6.-150);
	\path[draw=blue] (O6.-130) to node [] {} (O8.40);
	\path[draw=blue] (O8.60) to node [] {} (O6.-150);
	\path[draw=blue] (O3.-30) to node [] {} (O8.120);
	\path[draw=blue] (O8.140) to node [] {} (O3.-50);
	\end {tikzpicture}
}\\
            (\textbf{b}) CPG-network
        \end{minipage}
           \caption{(\textbf{a}) A single CPG. (\textbf{b}) The CPG-network schematic of our Spider robot, containing eight CPGs (numbers) with ten pairs of neighboring connections (blue arrows).}\label{fig:CPG_model}
\end{minipage}
\end{figure}

\autoref{fig:CPG_model}a shows a single coupled SO2-oscillator (i.e., CPG) formulated as two paired neurons (an $x_i$- and $y_i$-neuron). Here, $i$ denotes the specific joint that we associate with this CPG. The change of a neuron's state is calculated by multiplying the current state of the opposite neuron with a weight ($w$). The weights $w_{x_iy_i}$ and $w_{y_ix_i}$ represent the connection strength between each pair of neurons. An output neuron $o_i$ is coupled to the $x$-neuron with connection strength $w_{x_io_i}$. The output of $o_i$ serves as the input command for the servomotors in a specific joint (\textit{i.e.} servo). In our case, we selected a tangent hyperbolic activation function as our output neuron to normalize the motor input between [-1,1], which corresponds to the min/max angular positions of the joints. It is important to note here that the output neuron only reads out the current state of the $x$-neuron, and, hence, neither influences the CPG dynamics nor does it have an initial state.

To enable more complex output patterns, symmetrical/bidirectional connections between neighboring joints (denoted within neighbor set $\mathcal{N}$) are added between oscillators' $x$-neurons (see \autoref{fig:CPG_model}b). We define the set of neighboring joints in joint $i$ ($\mathcal{N}_i \subset \mathcal{N}$) as the set of all joints $j\neq i$ located within a 2-module distance from $i$. That is to say, if we represent our robot morphology as a connected a-cyclic graph (a tree with the top node being the Core module), then all joints $j$ that are connected within two or less edges away from joint $i$ are considered neighboring joints in $\mathcal{N}_i$. The set $\mathcal{N}_i$ contains all the neighboring inter-connections with $x_i$ in the CPG-controller, with its resulting ODE shown in \autoref{eq:ODE_neurons}. 

        \begin{align}\label{eq:ODE_neurons}
                \Dot{x}_i = w_{y_ix_i}y_i + \sum_{j \in \mathcal{N}_i} x_{j} w_{x_jx_i} &&\Dot{y}_i = w_{x_iy_i}x_i
        \end{align}
        
The connections with the neighboring joints $j$ at $i$ are denoted as $w_{x_jx_i}$. To simplify the search space, the following relations are imposed between the weights: $w_{x_iy_i} = -w_{y_ix_i}$; $ w_{x_jx_i} = -w_{x_ix_j}$; $w_{x_io_i} = 1$.
The size of the search space for WO and ISO can now be formulated as follows: For a given robot, the number of CPGs is equal to the total joints ($k$), while the number of neighboring pair connections ($|\mathcal{N}|$) is based on the modular tree-based distance. This means that for WO we will have $N_{weights}=k+|\mathcal{N}|$ weights to optimize; and for ISO $N_{state}=2\cdot k$ initial states (two neurons for each CPG oscillator).

\subsection{System description}
\subsubsection{Robots}
Our robots consist of modules which are based on the RoboGen framework \cite{auerbach2014robogen}. A modular design space has the distinct advantage of rapid real-world prototyping through pre-fabrication of 3D printed modules and off-the-shelf components, as also shown in other frameworks \cite{pollack2000golem, veenstra2017evolution, faina2013edhmor}. We altered the original Robogen module designs to prevent breakage, improve servomotors, and allow feedback control. This new design consists of three components: a core component (\unit[9$\times$9$\times$6]{\cm}, \unit[250]{gr}), brick components (\unit[6$\times$6$\times$6]{\cm}, \unit[30]{gr}), and active hinges (\unit[12$\times$6$\times$6]{\cm}, \unit[69]{gr}). The core component is a container with a micro-controller (Raspberry Pi 3B+ \footnote{\url{https://www.raspberrypi.org/products/raspberry-pi-3-model-b-plus/}}), a custom designed bridge to our servomotors (as detailed in \cite{van2021influence}) and a battery (LiPO, \unit[3.7]{V}, \unit[7800]{mAh}). It has four attachment slots on its lateral faces, to which other modules can be attached via screws. The brick component is a smaller cube with similar attachment points also on its lateral sides. The active hinge component is a joint actuated by a servomotor (Tower Pro MG996R, \footnote{\url{https://www.towerpro.com.tw/product/mg996r/}}) with an attachment point on both ends. 

For our learning experiment, we use six different legged robot designs based on our modular framework. This set can be divided into two groups: (1) Bio-inspired \textit{hand}-designed robots that resemble animals with distinct locomotion patterns (top row in \autoref{fig:robots}). (2) Evolutionary {automated}-designed robots via a novelty-search algorithm (bottom row in \autoref{fig:robots}). The latter group evolved to be fast in gait and morphologically different \cite{van2023comparing, miras2019effects}. A description of each robot, along with its corresponding CPG-network parameters, are presented in \autoref{tab:robots_spec}.

\begin{table}[ht]
    \centering
    \footnotesize
    \begin{tabular}{lcccccc}
                        & Ant & Gecko & Spider & Blokky & Salamander & Stingray \\ \toprule
         Evolved/Designed& Des. & Des. & Des. & Evo. & Evo. & Evo. \\ 
         Blocks         & 8   & 6     & 8      & 9      & 10         & 5 \\
         Servos: $k$    & 8   & 6     & 8      & 5      & 14         & 9 \\
         $N_{weights}$ $\left(|\mathcal{N}|\right)$  & 20(12)  & 13(7) & 18(10) & 7(2) & 24(10) & 18(9) \\
         $N_{state}$    & 16  & 12    & 16     & 10     & 28         & 18 \\
    \end{tabular}
    \caption{Technical description of our six robots. }
    \label{tab:robots_spec}
\end{table}

\subsubsection{Simulator}
For our simulation experiments, we use a high-performance GPU-based 3D physics simulator for robot learning named Isaac gym \cite{makoviychuk2021isaac} (with an NVIDIA GeForce RTX 3070). Isaac gym enables parallelization during optimization. During WO this allows us to simulate a whole population of robots every generation of our evolutionary algorithm, and in ISO to simulate all our trials at once. We set the simulator update time to \unit[60]{\hertz} with simulator sub-step 2 and use the PhysX physics engine with temporal Gauss-Seidel solver. The robots are spawned in a neutral position (all joint angles \unit[0]{\radian}) centred on a flat horizontal plane, as illustrated in \autoref{fig:robots}. Our world-frame coordinate system has its origin on a flat surface with $x$/$y$-direction spanning across the ground and a positive z-direction perpendicular to the surface pointing `up', i.e., opposite to the direction of gravity (\unit[9.81]{\metre\per\s$^{-2}$}). During simulation, we log both the position \{$x$, $y$, $z$\} and orientation \{$\phi_x$, $\phi_y$, $\phi_z$\} of the robot core component, each time we update the controller servomotor input (update frequency is \unit[10]{\hertz}). At the end of a trial, the logged robot state data ($\mathcal{D}$) are used for evaluation. 

\subsubsection{Real world}
For our real world experiments we used our own tracking-system, which merges the footage from six cameras (Dahua, IPC-HFW1320SP) into a single video-stream on a central laptop. Merging the cameras creates an image of 3964x1776 pixels with a resolution of \unit[0.26]{m\metre\per\px$^{-1}$}. The video-stream provides a top-down view of our lab at 30 frames per second. We made a custom tracking script, using ArUco markers \cite{Garrido-Jurado20142280}, to log both the position and orientation of the Core module every 0.1s (\unit[10]{\hertz}) in the same world-frame coordinates as the simulation experiments (positive $z$-direction up). From a central laptop a start command automatically uploads a CPG controller via WiFi. The states of the robot and controller are buffered until the end of a trial, after which fitness values are calculated and all the data are saved. The CPG-controller runs locally using Python on a Raspberry Pi that sends the corresponding PWM servomotor commands with PIGPIO. For stable servomotor behavior, the CPG-controller commands are provided at \unit[64]{\hertz}. 

A trial is initiated by first starting/recording the merged video-stream, proceeded by sending a start signal from the laptop to the Raspberry Pi. During a trial, all intermediate controller states are logged locally on the Raspberry Pi, while all frames from the video-stream are buffered on the laptop. If the Raspberry Pi is finished, then a stop signal is sent to the laptop and the video-stream stops. In this way, the video-stream recording will always cover both the beginning and end of the trial. We match the intermediate controller states to the buffered video frames in order to obtain the CPG initial state samples corresponding to the captured robot states (position and orientation). 

\subsection{Controller optimization}
We assess three basic locomotion skills in this paper \Gait, \RotL and \RotR for 60 seconds ($t_{eval}=\unit[]{60\s}$). Gait fitness is defined as the absolute planar distance traveled divided by time ($f_{\Gait} = [\unit[]{\cm\per\s}]$, see \autoref{eq:fit_gait}). Rotational fitness is defined as the amount of horizontal rotation ($\phi_z$) in the preferred direction divided by time ($f_{\RotR} = [\unit[]{\radian\per\s}]$, see \autoref{eq:fit_rot}). For $f_{\RotL}$ we use a stable unwrapping algorithm (see \autoref{alg:unwrap}) that accounts for precession projection artefacts to prevent exploitation by our learning algorithm \cite{9308272}. We run $N_{rep}$ repetitions of the same experiment, with a single experiment consisting of several $trials$ that provides $samples$ for the optimizer. In the case of ISO, a single trial contains multiple initial state samples due to the bootstrapping trick. WO and ISO each have their own procedure which will be explained in turn shortly. A summary of the respective optimizer parameters is shown in \autoref{tab:parameters_opt}.


\begin{table}[ht!]
    \setlength{\textfloatsep}{0.7\baselineskip plus 0.2\baselineskip minus 0.5\baselineskip}    
    \footnotesize
    \caption{\small Hyperparameters of the RevDE learning algorithm}
    \label{tab:parameters_opt}
    \centering
    \renewcommand{\arraystretch}{1}
    \begin{tabular}{l ll}
    \toprule
    \textbf{RevDE}         & Value     & Description \\ \midrule
    $\lambda$           & 30        & Population size \\
    $\mu$               & 10        & Top-samples size  \\
    $F$                 & 0.5       & Scaling factor\\ 
    $CR$                & 0.9       & Crossover probability \\ 
    $N_{weights}$        & $k+|\mathcal{N}|$      & Search space size\\ 
    $t_{trial}$        & \unit[60]{\s}      & Trial time \\ 
    \bottomrule
    \toprule
    \textbf{Random Search} & Value     & Description \\ \midrule
    $\textit{U}(m,n)$           & [-1, 1]        & Uniform distribution \\ 
    $N_{state}$        & $2\cdot k$      & Search space size \\
    $t_{trial}$        &  \unit[120]{\s}      & Trial time \\
    \bottomrule
    \end{tabular}
\end{table}

\subsubsection{Weight optimization}\label{sec6}
For Weight optimization (WO) we represent our CPG-network as a vector of real numbers ($\mathbb{R}^{N_{weights}}$) that define the connection strength within and between oscillators. Here, the initial state of the CPG is fixed during optimization to $\frac{1}{2}\sqrt{2}$ for all neurons. This value induces a $\frac{1}{2}\pi$ phase-offset within a single oscillator pair between the $x$- and $y$ neuron. 
The WO search-space size for each robot is shown in \autoref{tab:robots_spec} ($N_{weights}$). Every time we sample a new vector of weights from the search space, we test for 60 seconds for each trial ($t_{trial}=\unit[60]{\s}$) after which a fitness value is assigned for the corresponding skill.

To optimize our vector of weights we use an evolutionary algorithm called Reversible Differential Evolution (RevDE, \cite{tomczak2020differential}). This algorithm iteratively proposes a solution by applying a special set of linearly reversible transformations to produce a wide coverage over the search space when selecting new samples from search space \cite{tomczak2020differential}. RevDE works as follows (see \autoref{alg:RevDE}): (1) initialize a population of $\lambda$ uniformly sampled weight vectors, $pop \in \mathbb{R}^{\lambda\times N_{weights}}$ in range $\textit{U}[-1,1]$; (2) Obtain fitness values ($\mathcal{F}$) over all $\lambda$-samples; (3) select the top $\mu$-samples to obtain genotype vectors $\mu_1\in\mathbb{R}^{\mu\times N_{weights}}$; (4) randomly row shuffle $\mu_1$ twice to obtain two additional vectors ($\mu_2,\mu_3$), and create new samples with the following linear relations: $\lambda_1=\mu_1+F(\mu_2-\mu_3)$, $\lambda_2=\mu_2+F(\mu_3-\lambda_1)$,  $\lambda_3=\mu_3+F(\lambda_1-\lambda_2)$. Apply uniform crossover with probability $CR$ and repeat from (2). The number of evaluated samples $(\lambda$-samples) is three times that of the top-size $\mu_1$, which is set to 10 (\autoref{tab:parameters_opt}). The parameters are set to scaling factor $F=0.5$ and cross-over rate $CR=0.9$, which have been shown to provide robust performance over different morphologies \cite{van2023comparing}.

\begin{algorithm}[!ht]\footnotesize
  \caption{\small RevDE weight optimization of CPG networks}
  \label{alg:RevDE}
    Initialize pop: $pop \longleftarrow \textit{U}(-1,1)$  \Comment{$\in \mathbb{R}^{\lambda\times N_{weights}}$}
    
    \For{$N_{gen}$}{
        $\mathcal{D} \longleftarrow \textrm{TEST}(pop, t_{trial}=60) $ \Comment{Obtain data for evaluation}
        
        $\mathcal{F} \longleftarrow \textrm{EVAL}(\mathcal{D}) $ \Comment{Obtain list of fitness values $\mathcal{F}\in \mathbb{R}^{\lambda}$}
        
        $index \longleftarrow\mathrm{arg sort}(\mathcal{F})$ 
        
        $\mu_1=pop\left[index\left[:\mu\right]\right]$ \Comment{$\in \mathbb{R}^{\mu\times N_{weights}}$} 
        
        $\mu_2=\mathrm{shuffle}(\mu_1)$
        
        $\mu_3=\mathrm{shuffle}(\mu_2)$
        
        $\lambda_1=\mu_1+F(\mu_2-\mu_3)$ 
        
        $\lambda_2=\mu_2+F(\mu_3-\lambda_1)$
        
        $\lambda_3=\mu_3+F(\lambda_1-\lambda_2)$
        
        $pop \longleftarrow \mathrm{CR}\left(\left[\left(\lambda_1, \mu_1\right); \left(\lambda_2, \mu_2\right) ; \left(\lambda_3, \mu_3\right)\right]\right) $  \Comment{$\mathbb{R}^{\lambda\times N_{weights}}$}
    }
\end{algorithm}

\subsubsection{Initial state optimization}\label{sec7}
For ISO, we represent our CPG-network as a vector of real numbers ($\mathbb{R}^{N_{state}}$) that define the initial states of all neurons within our network. Here, all weights (both within and between oscillators) are sampled randomly ($\textit{U}[-1,1]$) and remain the same during optimization. 
The ISO search-space size for each robot is shown in \autoref{tab:robots_spec} ($N_{state}$). In every trial we test a random initial state for 120 seconds ($t_{trial}=\unit[120]{\s}$); in order to bootstrap the number of samples for each trial, we evaluate intermediate CPG-states with a rolling window $t_{eval}=\unit[60]{\s}$, which results in $N_{samples/trial}=600$. 

At the beginning of an ISO experiment, we fixate the weights of the CPG-network with random values between $\textit{U}[-1,1]$. For every trial, we `start' this random CPG-network with random uniformly sampled initial states ($\textit{U}[-1,1] \in \mathbb{R}^{N_{state}}$) and log the intermediate states every \unit[0.1]{\s}. This \textit{bootstraps} the number of samples by ten more initial states per second. The fitness values are then calculated over the logged sub-samples using a rolling window. We only consider samples with a full \unit[60]{\s} window length, thereby ignoring only considering the first $600$ initial states, $N_{samples/trial} = (t_{trial}-t_{eval})/0.1$. Fitness is calculated on each of these 600 sub-sample for all skills in \textit{parallel)}. Per trial for each skill ($\bm{s_{\RotL/\Gait/\RotR}}$), we only store the (sub)-sample with the highest fitness per skill \{$\bm{s^{*}_{\RotL}}, \bm{s^{*}_{\Gait}}, \bm{s^{*}_{\RotR}}$\} (see \autoref{fig:method4} B. data-evaluation). Therefore, a single trial provides $N_{skills}$ fitness values (one for each skill) corresponding to the (intermediate) CPG-state during the trial with the best behavior for that skill. Subsequently, we repeat this process by loading the same CPG-network (in terms of weights) with a random initial state, until our budget (in time) is spent. The pseudo-code for ISO is shown below (see \autoref{alg:State0}). To clarify, the pseudo-code presents a parallelized application (i.e., we TEST all initial states at the same time). In the real-world we TEST each trial sequentially.



\begin{algorithm}[!ht]\footnotesize
  \caption{\small Random search initial state optimization of CPG networks}
  \label{alg:State0}
  Initialize initial state trials: $trials \longleftarrow \textit{U}(-1,1)$ \Comment{$\in \mathbb{R}^{N_{trials}\times N_{state}}$}
  
  $\mathcal{D}_{trials} \longleftarrow \textrm{TEST}(trials, t_{trial}=120) $ \Comment{Obtain data for evaluation}
  
  $N_{samples/trial} \longleftarrow (t_{trial}-t_{eval})/0.1$
  
  $N_{eval} \longleftarrow t_{eval}/0.1$

    \For{$\mathcal{D}_{trial}$ \textbf{in} $\mathcal{D}_{trials}$ }{ 
    
        $samples\longleftarrow\left\{\mathcal{D}_{trial}[n:n+N_{eval}]\right\}_{n=0}^{N_{samples/trial}}$ \Comment{Obtain list of samples}
        
        $\mathcal{F}_{samples} \longleftarrow \textrm{EVAL}(samples) $\Comment{ $\in \mathbb{R}^{N_{samples/trial} \times N_{skills}}$}
        
        $\mathcal{F}_{trial} \longleftarrow  \max\left(\mathcal{F}_{samples}, \textrm{ axis}=0\right)$ \Comment{Best fitness per skill $\mathcal{F}_{trial}\in\mathbb{R}^{N_{skills}}$}  
        }
\end{algorithm}

\subsubsection{Proximal Policy Gradient}
We use the continuous action-space PPO algorithm of CleanRl \cite{huang2022cleanrl}, which is based on the original paper \cite{schulman2017proximal}. Here, we adapt our fitness functions (\autoref{eq:fit_gait}, \autoref{eq:fit_rot}) with $t_{eval}=0.1$ for our intermediate rewards. This corresponds to the update frequency of our controller, i.e. our rewards are the (rotational) distance traveled per controller update. Per experiment we run 300 trials of 60 seconds (similar to WO and ISO) with a total of 180,000 time-steps per skill (540,000 total). We update our learner in batches of 10 trials that are split into 100 mini-batches over 10 epochs. The remaining parameter values are taken from the original paper.

\subsection{Experimental setup}
We perform two different experiments: the first is a performance comparison between WO and ISO in simulation, in the second we compare ISO with sim-to-real WO in real-world. The experimental conditions are shown below in \autoref{tab:parameters_exp4}.

\begin{table}[ht!]
    \setlength{\textfloatsep}{0.7\baselineskip plus 0.2\baselineskip minus 0.5\baselineskip}    
    \footnotesize
    \caption{\small Experimental setup. Number of repetitions $N_{rep}$ indicated in brackets, and $N_\#$ indicating number of samples.}
    \label{tab:parameters_exp4}
    \centering
    \renewcommand{\arraystretch}{1}

    \begin{tabular}{r llllll}
        Sim (30)     & $N_{params}$ & $t_{total}$\unit[]{[\min]} & $t_{trial}$\unit[]{[\s]} & $N_{trials}$ & $N_{ \#/trial}$  & $N_{\#}$  \\ 
    \midrule
    \textbf{ISO} & $2\cdot k$ & $300$ & $120$ & $150$ & $600$  & $90,000$ \\
    \textbf{WO} & $k+|\mathcal{N}|$ & $900$ & $\phantom{1}60$ & $300\times3$ & $1$  & $900$\\
    \textbf{PPO} & $2(512 k+256^2)$ & $900$ & $\phantom{1}60$ & $300\times3$ & $600$  & $540,000$ \\
    \bottomrule
    \toprule
            Real (3)     & $N_{params}$ & $t_{real}$\unit[]{[\min]} & $t_{trial}$\unit[]{[\s]} & $N_{trials}$ & $N_{ \#/trial}$  & $N_{\#}$  \\ 
    \midrule
    \textbf{ISO} & $2\cdot k$ & $15$ & $120$ & $5$ & $\leq600$  & $\leq3,000$ \\
    \end{tabular}
\end{table}
\subsubsection{Simulation experiments}
For the simulation experiments we carry out a `per skill' comparison of the fitness per simulated time in minutes between ISO and WO. It should be noted that the simulated time for WO is reported relative to each skill (as opposed to the cumulative time needed to learn the three skills sequentially). For the purpose of statistical analysis, we repeat each optimization 30 times ($N_{rep}=30$). We compare the MBF learning curves (\autoref{alg:mbf}), ISO out-performance time (\autoref{alg:norm_perf}) and TTEQ (\autoref{alg:tteq}).

For WO, a separate experiment is run for each skill. An optimization run for a single skill takes place over \unit[300] trials, each of which are \unit[60]{\s} in length. At the end of each trial, a single fitness value is calculated as a sample for the RevDe optimizer. We log the `best-fitness-so-far' for every trial. In the end, the total simulated time to learn three skills is $t_{total} = 300\times3=\unit[]{900\min}$. 

For ISO, all three skills are learnt simultaneously in a single run. A single run takes place over \unit[150] trials, each of which are \unit[120]{\s} in length. At the end of each trial, rolling window fitness values are calculated for all skills with the same evaluation time as WO, i.e., \unit[60]{\s}. Within a trial, the best sub-sample fitness represents the trial fitness for each skill. We log the 'best-fitness-so-far' for every trial. In the end, the total time in simulation needed to learn three skills is $t_{total} = \unit[300]{\min}$. 

\subsubsection{Real experiments}
Our real-world experiments are the physical instantiation of \autoref{alg:State0}. We start by placing a real robot in the centre of a \unit[]{7x4\metre} arena. We run a CPG-network with random weights and an initial state from a central laptop for \unit[120]{\s}. A custom built tracking system using ArUco markers tracks the robot's position and orientation every \unit[0.1]{\s}. At the end of the trial, the tracking data is then matched to the controller state at that time, and a rolling fitness is calculated for all the skills. This process is repeated until all the trials are complete.

A perfectly tracked run produces 600 samples, but due to noise and image merging artefacts, this number can sometimes be less. Although it rarely occurs in practice, missing robot data is filled in with the last previously captured state from the buffer. This operation only affects the fitness values when data is missing at the end of a trial because \Gait fitness only considers the state at the end and \RotL/\RotR fitness performs a sum of the angular differences between subsequent samples, which is zero when samples are copied. Furthermore, filling in missing data with the last-seen values underestimates fitness values, insofar as it effectively shortens the evaluation period while simultaneously dividing by the same fixed evaluation time $t_{eval}=60$.

For ISO we test five random initial states per run ($N_{trials}=5$) with three repetitions ($N_{rep}=3$). After all the trials are finished, we calculate the best overall initial state sample $s^{0*}$ for each skill and re-test in order to obtain the final fitness for each skill in a single run. Including overhead and re-testing the acquired skill, the total time for the real robot experiments is $t_{total} \leq \unit[15]{\min}$.

For the sim-to-real experiments we re-run the overall best WO* controller from the 30 runs completed in simulation for each skill on the physical robots. We re-test each WO* controller one time. We measure the performance in a similar fashion as we do for the ISO real setup. Ultimately, we compare the real-world performance of all three ISO runs and WO*.

\subsection{Metrics}
For the simulated environment, we apply 30 repetitions of each learning experiment (30x, 2 optimizers, six robots, 3 skills). We report on MBF over time $\pm$ Standard Error ($N_{rep}=30$, $\pm SE$). The best performance over time indicates the highest fitness so far. For each skill, the fitness is calculated as follows:

\subsubsection{Gait fitness}
We define gait fitness as the absolute planar displacement speed in centimetres per seconds [\unit[]{\cm\per\s}]. We define our world frame's origin at the ground, where the x- and y-directions span the horizontal ground and the z-direction points up. Gait performance is then calculated as follows: 

\begin{equation}\label{eq:fit_gait}
    f_{\Gait}(t) = \frac{\sqrt{(x_{t+t_{eval}}-x_{t})^2 + (y_{t+t_{eval}}-y_{t})^2}\phantom{test}}{t_{eval}}
\end{equation}

Where $t_{eval}$ represents the evaluation time in seconds (\unit[60]{\s}), and ($x_t$, $y_t$) the planar coordinates at time $t$.

\subsubsection{Rotational fitness}
We define rotation fitness as the rotational speed around the $z$ axis of our world frame in radians per seconds [\unit[]{\radian\per\s}]. Our skills \RotL and \RotR only define which rotational direction is preferred which, in turn, changes the sign of $f_{\RotL}$. We calculate the rotational fitness as follows:

\begin{equation}\label{eq:fit_rot}
    \phantom{_g}f_{\RotL}(t) = \frac{1}{t_{eval}}\sum_{t+dt}^{t+t_{eval}}\mathrm{UNWRAP}_{\phi_z}\left(\bm{\phi}_t, \bm{\phi}_{t-dt}\right)
\end{equation}

Here, $\bm{\phi_t}$ denotes a vector of angles [$\theta_x, \theta_y, \theta_z$] at time $t$. We add the differences in planar angle $\phi_z$ between the subsequent samples to obtain the total amount of planar rotation achieved. The differences are unwrapped from [$0,2\pi$) with a custom unwrapping algorithm (from \cite{9308272}), which accounts for precession angle artefacts that can be exploited by optimizers. The pseudo-code is shown below in \autoref{alg:unwrap}.

\begin{algorithm}[ht]\footnotesize
    \caption{\small\label{alg:unwrap}UNWRAP: Calculate relative $z$-angle between subsequent time-steps}
    \SetKwFunction{unwrap}{UNWRAP}
    \SetKwProg{Fn}{Function}{:}{}
    \Fn{\unwrap{$\bm{\phi}_t, \bm{\phi}_{t-dt}$}}{
        
            $r_{x_{t-dt}}=R_{zyx}(\bm{\phi}_{t-dt})[:2, 0]$ \Comment{Planar $x$-axis orientation vector at $t=t-dt$}
            
            $r_{x_t}=R_{zyx}(\bm{\phi}_t)[:2, 0]$     \Comment{Planar $x$-axis orientation vector at $t=t$}
            
            $dot = r_{x_{t-dt}}[0] * r_{x_t}[0] + r_{x_{t-dt}}[1] * r_{x_t}[1]$   \Comment{Dot product between $r_{x_{t-dt}}$ and $r_{x_t}$}
            
            $det = r_{x_{t-dt}}[0] * r_{x_t}[1] - r_{x_{t-dt}}[1] * r_{x_t}[0]$   \Comment{Determinant between $r_{x_{t-dt}}$ and $r_{x_t}$}
            
            delta$ = \arctan2D(det, dot)$       \Comment{Relative $z$-angle}
            
            \Return delta
        }
\end{algorithm}

\subsubsection{Mean Best Fitness}
For each skill, we log the fitness values of a single run for each trial $\mathcal{F}_{trials}=\left[f_{skill_1}, f_{skill_2}, \dots,f_{skill_{N_{trials}}}\right]$ $\in \mathbb{R}^{N_{trials}}$. The repeated experiments ($N_{rep}=30$) are concatenated $\mathcal{F}_{runs} \in \mathbb{R}^{N_{rep}\times N_{trials}}$. This allows us to calculate the MBF as follows:

\begin{algorithm}[ht]\footnotesize
  \caption{\small\label{alg:mbf}Mean Best Fitness}
  
  \For{$ind$ \textbf{in} $N_{rep}$}{
    $\mathcal{F}_{trials}=\mathcal{F}_{runs}[ind, :]$ \Comment{Obtain fitness values for run $ind$}
    $\mathcal{F}_{best}=[]$
    
    \For{$\mathcal{F}_{trial}$ \textbf{in} $\mathcal{F}_{trials}$}{ 
    
        \eIf{$\mathcal{F}_{trial}>\mathcal{F}_{best}$ \textbf{or} $\mathcal{F}_{best}=[]$}{
            $\mathcal{F}_{best}\mathrm{.append}(f_{trial})$}{
            $\mathcal{F}_{best}\mathrm{.append}(\mathcal{F}_{best}[-1])$}
    }
    $\mathcal{F}_{runs}[ind, :] = \mathcal{F}_{best}$  \Comment{Replace run with best-so-far fitness values}
  }
  $MBF = \mathrm{mean}\left(\mathcal{F}_{runs}, \textrm{ axis}=0\right)$ \Comment{Average over all repetitions}
\end{algorithm}

\subsubsection{Time Till Equal Equality}
TTEQ measures the first cross-over point for the MBF curves. It indicates how long a certain optimizer method outperforms another one.

\begin{algorithm}[ht]\footnotesize
  \caption{\small\label{alg:tteq}Time Till Equal Quality}
    $ind = MBF_{ISO} < MBF_{WO}$ \Comment{Obtain indexes where ISO is lower than WO}
    
    $TTEQ = \min{t[ind]}$ \Comment{Obtain the first time where ISO is lower than WO}
\end{algorithm}

\subsubsection{Normalized Performance}
We normalize the performance based on the MBF end-values of WO (as it is the best performing method overall in simulation). While we apply normalization for each skill for WO, for ISO we apply normalization based on the summed performance of all the skills (denoted with $\Sigma$). Summed normalization indicates an average relative performance over time to learn multiple skills in parallel. This allows us to generalize across multiple skills and compare both the parallel- and sequential-learning time-efficiency of ISO and WO, respectively.

\begin{algorithm}[ht!]\small
  \caption{\label{alg:norm_perf}Normalized Performance}
    $norm_{\Sigma}=0$
    
    $ISO_{\Sigma}=[0]^{N_{trials}}$
    
    \For{$skill$ \textbf{in} ${\text{\Gait, \RotL, \RotR}}$}{
    
    $norm = MBF_{WO_{skill}}[-1]$
    
    $norm_{\Sigma} \mathrel{+}= norm$
    
    $WO_{{skill}_{norm}} = MBF_{WO_{skill}}/norm$
    
    $ISO_{\Sigma} \mathrel{+}= MBF_{ISO_{skill}}$
    }
    
    $ISO_{\Sigma} = ISO_{\Sigma}/norm_{\Sigma}$
\end{algorithm}

\subsection{Statistical Analysis}
All optimization experiments in simulation were conducted 30 times ($N_{rep}=30$). 
Within robot/tasks comparison of ISO vs. WO is subject to a standard student \textit{t}-test after passing checks for normality (Shapiro-Wilk test) and equal variances (\textit{F}-test). Significant differences are considered at $p<0.05$, with Bonferronni correction for aggregated results when comparing WO vs ISO on all skills and morphologies ($p<0.05/m$ with $m=18$ learning tasks).


\textbf{Code and Data Availability}
The raw and processed data generated in this study have been deposited in the Dataverse database and are available at \url{https://doi.org/10.34894/XZQZAZ}. 
Learning algorithm, CPG analysis, and experimental code (including code from Appendix) are peer-reviewed on CodeOcean \cite{codeocean}: \newline\url{https://codeocean.com/capsule/3493679/tree/v1}.\newline
Real world experiments tools (including: analysis, fitness functions, merged camera stream, tracking software): \newline\url{https://github.com/fudavd/multi-stitch-stream}.

\section*{Acknowledgments}
\addcontentsline{toc}{section}{Acknowledgments}
This work is supported by the Technology Innovation Institute (TII).




\section{Introduction}
Animal groups in nature have evolved collective motion behaviors for certain benefits, like increasing environmental awareness \cite{couzin2005-animalgroups, kearns2010-bacterialswarming} and safety from predation \cite{olson2013-predatorconfusion}. Implementing such behaviors in robot swarms has several applications, but designing adequate robot controllers is a great challenge. The principal problem is that the controller can only govern the individual robots directly, but the desired behavior is defined at the group level \cite{hasselmann2021empirical}. Human developers have typically no good understanding of the link between the controller details and the induced swarm behavior that is an indirect result of the interactions between individual swarm members and the environment. In the current practice, controllers in swarm robotics require an extensive manual design and fine-tuning process before obtaining desired behavior. 

We developed an artificial evolutionary pipeline to automatically design and evaluate a robot controller for the swarm to complete a collective level task. On a conceptual level, we may note that we hereby address a higher level of complexity than traditional Evolutionary Computing (EC), even one step higher than evolutionary robotics (ER). As explained in \cite{eiben2003introduction}, in traditional EC we have a 3-step chain from genotypes to fitness values, genotype $\rightarrow$ phenotype $\rightarrow$ fitness, while in ER the chain is 4-fold, genotype $\rightarrow$ phenotype $\rightarrow$ behavior $\rightarrow$ fitness. In the case of robot swarms, we have one more level, that of the group behavior: genotype $\rightarrow$ phenotype $\rightarrow$ individual behavior $\rightarrow$ group behavior $\rightarrow$ fitness. Obviously, the arrow from individual behavior to group behavior is complex in itself, thus optimizing along the whole chain from genotype to fitness is very intricate. This makes the endeavour of evolving controllers for swarms far from trivial. 

For this paper we consider a very challenging task, where a swarm consisting of robots with very limited sensing and communication abilities must follow the gradient of an environmental feature, e.g., light intensity, temperature, or radiation. To solve this problem, we define genotypes that represent neural networks (phenotypes) in a robot and postulate that all robots of the swarm have the same controller. The fitness of a genotype / phenotype will be defined by the behavior of the swarm. Our goal is to investigate whether an evolutionary algorithm is capable of finding controllers that enable the swarm to follow the gradient. 

By constraining a single agent's sensor capabilities to only sense the local scalar value of the gradient, we know that this task is not solvable for an agent in isolation without any memory to compute the gradient on its own. This problem is especially interesting as we thus investigate if an environment can induce a collective behavior which would normally not evolve, to complete a task that otherwise could not be solved. In contrast to other works, the optimization does not directly solve for emergent behavior but might exhibit it as a successful strategy.
To this end, we are seeking answers to the following specific research questions.

{\textbf{Research Question 1:}}
Will there be any emergent collective behavior among the members of the swarm? 

{\textbf{Research Question 2:}}
How flexible and scalable are the evolved solutions with respect to changes in the size of the robot arena and the number of robots in the swarm?

To answer our research questions, we set up a system where the controllers of (simulated) swarms of differential drive robots are evolved by Differential Evolution. We measure the flexibility and scalability by running experiments in three arenas and with three different group sizes in simulation (Isaac Gym \cite{makoviychuk2021isaac}). 

\section{Related Work}

Automated design for (swarm) robotics has a long history with varying approaches \cite{ prabhu2018survey, birattari2021automode}, which often require an optimization method that can handle high non-linearity and non-smooth objective functions. EAs proved themselves to be capable of handling such a hard task, which resulted in the development of fields like evolutionary robotics \cite{floreano2000evolutionary}. In this paper, we draw inspiration from this field and create an automated design pipeline suitable for swarm robotics. A strong difference in our approach is the level of control that we specify. Evolutionary robotics is often focused on learning low-level controls on a single robot for basic tasks like locomotion \cite{nelson2009fitness, Wright2015}. In contrast, a swarm controller indirectly specifies a higher level of behavior on the group level \cite{trianni2014evolutionary}. This additional level of abstraction creates a situation where we cannot expect a group behavior to directly correlate with the objective. Nevertheless as mentioned before, the ability to optimize control when there is no clear mapping from variables to objective is what makes evolution attractive to apply on swarms as well. 


Indeed, non-automated designs usually struggle to resolve this mapping between agent-based (i.e. microscopic) and group-level (i.e. macroscopic) models, and must therefore resort to trial-and-error. In some cases, it is however possible to establish a quantitative micro-macro link so that mathematical equations are able to indicate the settings required to achieve an objective or, vice versa, to predict the outcome of an experiment from variables' values \cite{reina2015quantitative}. Such a micro-macro link is possible for binary (leftward or rightward) alignment of multiple agents \cite{hamann2014derivation}, but it is inapplicable to more advanced alignment-dependent behaviors such as flocking \cite{reynolds1987flocks}. As a matter of fact, existing approaches are only relevant to a fairly limited set of tasks; specifically collective decision-making between a finite quantity of options, i.e., best-of-\textit{n} problems \cite{valentini2017best}.

Typically, collective perception can be represented as a best-of-\textit{n} problem \cite{Valentini2016-Collectivefeature}, but continuous representations are more susceptible to be found in real environments \cite{puckett2018collective,Berdahl2018-animalnavigation,Wahby2019-swarmdensity}. 
In formal definition collective perception is; \textit{where social interactions among individuals lead to collective computation of an environmental property by only allowing scalar measurements made by these individuals} \cite{Berdahl2018-animalnavigation}. This approach has been proven viable in biology \cite{puckett2018collective, berdahl2013emergent}, where schools of fish --incapable of sensing the light gradient-- can nevertheless follow the gradient to hide in the dark, by only considering their interactions with other fish. 

Collective perception is also studied on artificial systems. In \cite{Wahby2019-swarmdensity}, a honey-bee inspired algorithm was used for aggregation of a robot swarm as a result of local interactions between each other and an environmental feature. Unfortunately, the authors provided a discrete environmental feature to the robots plus some additional (limited) communication capabilities, which made the resulting swarm behavior not truly obtained through a collective estimation alone. Another collective perception application can be found in \cite{Valentini2016-Collectivefeature} where robots collectively estimate certain environmental features and decide about the most frequent one (i.e. white tiles or black tiles in a tile grid). Yet again collective perception was not strictly necessary as the agents in \cite{Valentini2016-Collectivefeature} could communicate through voting and direct information exchange. There are many more papers demonstrating collective behavior of robot swarms where some form of collectively perceived feature(s) is used. Unfortunately, most of them either use communication capabilities or directional information related to the environment and/or other agents \cite{Schmickl2016-FSTaxis, Varughese2019-swarmfstaxis}. An exception to the rule is presented by \cite{karaguzel2020collective}. Here, a true collective perception method is proposed to control a flocking robot swarm that can sense a gradient, in an emergent way, and follow it in the environment, using only local and scalar measurements. The only downside to this solution is the manual optimization and design of the governing rules, which is a slow process that requires a lot of knowledge. Additionally, such an approach is not easily generalizable among other tasks and is negatively effected by individual failures or dynamic environments.

To cope with setups that are not fully specified at design time, automated design has been applied for collective perception tasks as well. \cite{Campo2011-selforg-discrim} employs an evolutionary optimization framework to optimize controller parameters of robots in a swarm, to learn to discriminate different shelters in the environment and aggregate at the best to exploit. Although successful, \cite{Campo2011-selforg-discrim} presented a limited usage of automated design with collective perception, due to a highly specific task and setup, with very little freedom for artificial evolution (only parameters of pre-designed controller rules were optimized). Another example for automated design on a controller with collective perception is in \cite{Shaukat2016-sourcelocal}, where an underwater swarm could locate an radio signal source and then move towards it as a cohesive group. Unfortunately, these real-life applications --although showing successful collective perception-- are limited in their automated design as they only optimize pre-designed control parameters (a biased controller with model parameters). A much more expressive optimization was shown by \cite{floreano2007evolutionary}, where a swarm of robots, controlled by evolved neural networks (which are model agnostic approximators), learned to locate the food in their environment by emerging a communication scheme. This study showed that swarm behavior can develop with model-agnostic  controllers undergoing evolution, even if such a behavior is not directly specified in the fitness function. 

More similar to the work presented here, \cite{ramos2019evolving} and \cite{witkowski2016emergence} presented automated design approaches to optimize not only controller parameters but controllers themselves (neural networks in these particular examples). \cite{ramos2019evolving} showed that a simulated swarm could learn similar flocking behavior, when directly optimizing for fitness functions that closely relate to well-known flocking rules (e.g. \cite{reynolds1987flocks}). In our opinion, these results are unsurprising as the flocking is directly optimized for, rather than emerged. Additionally, such an approach can be unsuitable for robots, as it may limit the controller's robustness when conditions change and with possible failures \cite{floreano2000evolutionary}. In our opinion, a more robust approach would be to favor a resultant swarm behavior of swarm (behavioral fitness) rather than directly rewarding the desired actions of individuals (functional fitness) \cite{floreano2000evolutionary}. \cite{witkowski2016emergence} showed that flocking-like behavior emerged when evolving a swarm of simulated agents only for exploiting a vital source for survival and reproduction. These results show that with an environment-driven task, swarm-level behavior can emerge if it provides an advantage with respect to solving the task collectively. In this paper, we hope to see the similar collective behavior emerge, by using a very generic controller (i.e. model agnostic) and optimizer in a more strict setting. 

To summarize, this study positions itself uniquely in that it hopes to show that, with limited sensing and true collective perception (no explicit communication or direct sensing), emergent collective behavior can be induced by the environment using a model-agnostic controller that undergoes evolution. 


\section{Methodology}
Full code base can be found in the following git repository: \\ \href{https://github.com/fudavd/EC_swarm/tree/GECCO_2022}{https://github.com/fudavd/EC\_swarm/tree/GECCO\_2022}

\subsection{Controller}
As a single behavior is considered during the evolutionary process, we use the same controller on every robot in the swarm. Thus, an individual in our EA refers to a single homogeneous swarm with the controllers of its constituents being an exact copy of the brain encoded by the genome. To be clear, the controller for each member of the swarm will be the same, but their respective input will likely differ due to their different positions and orientation within the swarm. Each controller provides a target velocity to the two wheels of the corresponding differential drive robot. For this we use a fully connected reservoir Neural Network (NN) controller where we only optimize the output layer using DE. The choice of this controller is based on two advantages: 1) NN are known to be model agnostic function approximators which means that there is no prior bias on the possible behaviors. 2) reservoir NN reduce the search space significantly (only the output layer is optimized) thereby increasing the learning rate.

In more detail, as an input to the reservoir NN we provide relative positional and heading information from 4 directional sensors, plus a local value sensor reading of a scalar field map (thus in total 9 different inputs) every 0.1s. The 4 directional sensors cover a combined 360 degrees view of the robot's surrounding, with a single sensor only sensing the nearest neighbor within a specific locally defined quadrant: Front-, Back-, Left-, and Right- quadrant (90 degrees each). The information obtained are the distance and relative heading of this neighbor with respect to the controlled robot (2 dimensions of information per sensor, in 4 quadrants). If no neighbor is present within the quadrant's maximum range (2m) the sensor set its distance measured to 2.0 and relative heading to 0. Lastly, local values of the scalar field are provided based on the robot's position in the environment. All sensor inputs are pre-processed to a [-1, 1] range, before being fed into the NN. The two NN outputs specify the heading- ($w \in [-1,1]$) and forward velocity ($v \in [0, -1]$), which are rescaled and translated to correct wheel speeds (up to $\pm \unit[14]{cm/s}$ per wheel). 

The reservoir NN architecture consist of an input layer ($\mathbf{s}_{in}$) with 9 neurons, 2 hidden layers (with Softplus functions) of the same size as the input layer, and an output layer (tanh$^{-1}$ of size 2). All weights are randomly initialized between [-2, 2] with a uniform distribution. We set all biases to 0 and will only optimize the weights of the output layer during evolution (18 weights). The final NN controller can be formalized as follows:

\begin{equation}
    NN = \mathrm{tanh}^{-1}\left(\mathbf{W}_{out}\mathrm{ReLU}\left(\mathbf{W}_{h2}\mathrm{ReLU}\left(\mathbf{W}_{h1}\mathbf{s}_{in}\right)\right)\right) 
    \tag{$\textrm{with,} \quad  \mathbf{W}_{h1, h2} \in \mathbb{R}^{9\times9} \quad \textrm{and} \quad \mathbf{W}_{out} \in \mathbb{R}^{2\times9}$\newline}
\end{equation}

Each individual in the EA population receives the same reservoir ($\mathbf{W}_{h1, h2}$) in their NN controller but with varying output layers $\mathbf{W}_{out}$. We encode the reservoir NN controller's phenotype by a vector of weights with $2\times9=18$ floating point numbers (min/max value $\pm10.0$) as its genotype ($x$), specifying the rows in $\mathbf{W}_{out}$.

\begin{equation}
    x = \left[\mathbf{W}_{out_{1:}}, \mathbf{W}_{out_{2:}}\right] 
    \tag{$\textrm{with,} \quad x \in \mathbb{R}^{18}$}
\end{equation}

\begin{figure}[hpt!]
  \includegraphics[width=0.75\linewidth]{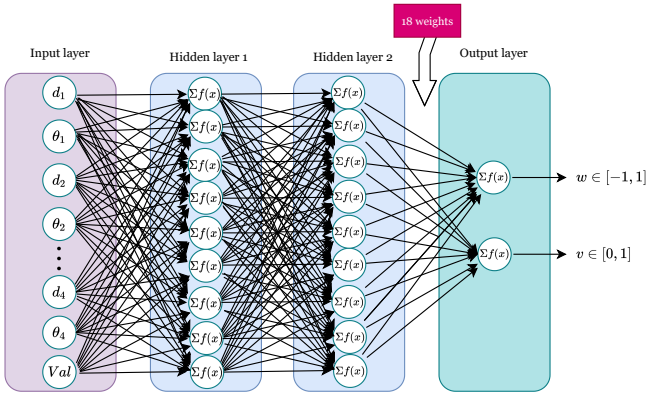}
  \centering
  \captionsetup{width=.95\textwidth}
  \caption{Controller phenotype.}
  \label{fig:NN_EC_s}
\end{figure}

The phenotype of the controller is illustrated in \autoref{fig:NN_EC_s}.

\subsection{Evolutionary Algorithm}
Evolutionary methods for solving NP-hard optimization problems have become a very popular research topic in recent years. Some of the more established methods are genetic algorithm (GA), particle swarm optimization (PSO), and differential evolution (DE), which all have shown to be able to solve hard optimization problems for various domains \citep{Deng2006,Ai2009,Wisittipanich2011}. \citep{Kachitvichyanukul2012} states that GA is more well-established because of its much earlier introduction however it has less ability to reach good solution without local search, while the more recent PSO and DE algorithms have started to attract more attention especially for continuous optimization problems. However in PSO, the best particle in the swarm exerts its one-way influence over all the remaining solutions in the population. This often leads to premature clustering around the best particle, especially if the fitness gaps are large. This is definitely an unwanted property, therefore in this paper, we choose DE to search our best solution. DE is a population-based Evolutionary Algorithm (EA) that samples new candidates by perturbing the current population \cite{storn1997differential}. The three main components in this method are as follows:

Differential mutation operator: a new candidate is generated by randomly picking a triplet from the population, $(x_i,x_j,x_k)\in X$, then $x_i$ is perturbed by adding a scaled difference between $x_j$ and $x_k$, that is:
        \begin{align}\label{eq:de1}
              y = x_i + F(x_j-x_k)
        \end{align}
where $F\in R_+$ is the scaling factor. 

Uniform crossover operator: sample a binary mask $m \in \{0, 1\}^D$ according to the Bernoulli distribution with probability p = P(md = 1) shared across all D dimensions, and calculate the final candidate according to the following formula:
        \begin{equation}\label{eq:de2}
              v = m \odot y+(1-m) \odot x_i
        \end{equation}

The last component is a selection mechanism: the authors of \cite{storn1997differential} proposed to use the “survival of the fittest” approach, i.e., combine the previous population with the new one and select N candidates with the highest fitness values, i.e., the deterministic ($\mu$ + $\lambda$) selection.

\begin{figure}[hpt!] 
  \centering \includegraphics[width=0.95\textwidth]{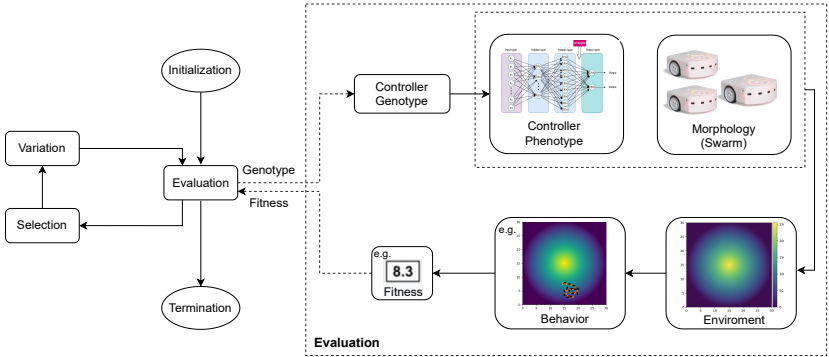}
  \vspace{0em}
  \captionsetup{width=.95\textwidth}
  \caption{Differential Evolution Framework: This is a DE framework to change the weights of the NNs of the robots to improve their controllers for the tasks. In the Evaluation box, we show examples of a robot morphology, controller, environment, behavior and fitness value.}
  \label{fig:swhole_process_EC_s}\vspace{-1em}
\end{figure}

Studies \cite{Panda2008,tomczak2020differential,Luo2021} have shown that the schemes for the trial vector (candidate) generation can have a significant influence on the algorithm's performance. Here we follow general recommendations which can be found in literature \cite{Pedersen2010} for stable exploration behavior, namely the crossover probability (CR) being fixed to a value of 0.9 and the scaling mutation factor (F) being fixed to a value of 0.5 (see \autoref{tab:parameters_EC_s}). We apply DE to change the weights of the NNs of the robots to improve their controllers for the tasks. The whole process is illustrated in \autoref{fig:swhole_process_EC_s}. 

    

\subsection{Performance measures}
\subsubsection{Fitness function}
\begin{figure}[ht!]
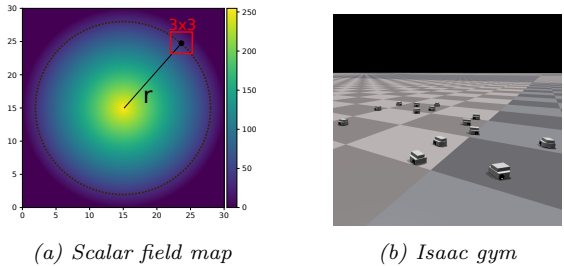

\vspace{-0.5em}
  \centering
\begin{minipage}[c]{0.75\linewidth}
\centering
        \begin{minipage}[t]{.4\linewidth}
            \centering
            \subfloat[(a) Scalar field map]{\includegraphics[width=\textwidth]{images/scalar_field2.pdf}}
        \end{minipage}
\hspace{2em}
        \begin{minipage}[t]{.35\linewidth}
            \centering
            \subfloat[(b) Isaac gym]{\includegraphics[width=\textwidth]{images/isaac_gym.png}}
    \end{minipage}
\end{minipage}
\captionsetup{width=.95\textwidth}
\caption{The swarm environment. (a) the scalar field map with its maximum value (255) in the center. Initial swarm position (shown as a black dot) is randomly placed on a circle with distance r away from the center. Individuals are randomly placed within a 3x3m bounding box (shown in red). (b) a random initial state is shown in simulation.}
\label{fig:swarm_env_EC_s}
\vspace{-1em}
\end{figure}

In a homogeneous swarm, we evaluate fitness on group level. In this research, we are mainly interested in a pure evolution-driven effects on swarm behavior without directly encouraging any form of communication, collaboration or collectiveness in the evolutionary framework. Therefore, the fitness was solely dependent on the swarm's ability to follow the increasing gradient of the scalar field which is defined in the environment (shown in \autoref{fig:swarm_env_EC_s}-a). Additionally, we would like to emphasize that our fitness function does not distinguish between robots that are sensing and following the increasing gradient collectively or as solitaries. The fitness function $F$ is defined as:

\begin{equation}
{F} = \frac{\sum_{t=0}^{T}{f_t}}{G_{max}\cdot{T}}\quad\text{and}\quad{f}_t = \frac {\sum_{n=1}^{N}{G_n}}{N}
\label{eq:r_i_sum_EC_s}
\end{equation}

Where $G_n$ is the scalar value of the grid cell in which agent $n$ (of all agents, $N$) is located at a time $t$. Thus the fitness at a specific time ($f_t$) is calculated as the mean scalar value of all swarm members. Final fitness ($F$) is calculated by averaging all $f_t$ over total simulation time $T$. At last, we normalize using the maximum scalar value $G_{max}$, always equal to $255$ for all experiments. 

\subsubsection{Behavioral measures}
Besides our fitness function $F$, we want to investigate the possible emergence of flocking behavior. For this, we analyze two additional behavioral measures. The first measure: \textit{order} ($\Phi$), is defined as follows: 

\begin{equation}\label{eq:align_EC_s}
{\Phi} = \frac{\sum_{n=1}^{N}{\varphi_n}}{N}\quad\text{and}\quad{\varphi_n}=\frac{\Big|\Big|\left(\sum_{p=1}^{P}{\angle{e^{j \theta_p}}}\right)+\angle{e^{j \theta_n}} \Big|\Big|}{P+1}
\end{equation}

Here, $\varphi_n$ defines the order value calculated for agent $n$. Which is the average current heading direction of agent $n$ (noted as $\angle{e^{j \theta_n}}$) and all its perceived neighbors $P$ (noted as $\angle{e^{j \theta_p}}$). The total swarm order $\Phi$ is then defined as the average $\varphi_n$ over all agents in the swarm. Order measure gives a powerful insight about the alignment of agent's direction of motion. If all agents move towards the same direction, then the order measure approaches to 1; and if they move in different directions, order approaches to 0.

In addition to order, the number of collisions between robots are also checked periodically. This check is not trivial since robots are modelled in a physics simulator, all collisions are realistically modeled and certainly have affect on the movement capabilities of colliding robots. In other words, when agents are collided, they can tip over, stop or drag each other because of the physical interaction between them. We will report on collision by plotting the total number of collision present in the swarm as a function of time.

The final measure to analyze is the trajectories of robots during an experiment. By including and highlighting the start and end points as well, presenting the trajectories of robots on the scalar field provides a clear intuition about the swarm's motion characteristics. Trajectories can show us if there is any correlated forward motion or rotation among the swarm and the relative difference of the local values of the scalar field between start and end points. In addition, trajectories can reveal any kind of behavioral differences between controllers that are evolved in different conditions and obtaining similar fitness. Finally visual inspection of the trajectories will be made. 

\subsection{Experimental setup}
\subsubsection{Simulator}
At the start of an evaluation, a single swarm is randomly placed at a fixed distance away from the center of the arena (sampled uniformly on a circle with radius r). Each swarm member is then randomly placed inside a 3x3m box (the red square in \autoref{fig:swarm_env_EC_s}-a) around this position with a random heading. We simulate the behavior swarm for 10 minutes ($\mathrm{dt=0.05s}$) and sample the controller every 0.1s for new motor inputs. Simulation of the swarm is done using Isaac Gym (see \autoref{fig:swarm_env_EC_s}-b), which allows us to simulate the whole population of 25 different swarm individuals in parallel. To reduce the effect of lucky runs, we evaluate every individual 2 times, in which the lowest value is taken as the final fitness. Thus, in total we evolve 25 individuals with 2 repeated tests for 100 generation in 30 different runs with 3 different evolutionary experiments (45.000 evaluations of 10 minutes in simulation).

\subsubsection{Evolutionary Experiment}
In total we perform DE optimization 3 times in different environments (3 arenas), all with the same task of collective gradient sensing. The individuals in the population are not a single robot (which is more conventional in evolutionary robotics) but rather describe a homogeneous swarm. An initial population of 25 individuals (i.e. 25 different homogeneous swarms) is randomly generated as the first generation. We test each individual twice and use the result of the lowest fitness. In each generation, 3 individuals are randomly picked to generate a new individual by crossover and mutation. The next generation is formed by the top 25 individuals from the previous generation plus the new candidates according to fitness value. The evolutionary process is terminated after 100 generations. All the evolutionary experiments are repeated 30 times independently to get a robust assessment of the performance per data set.

   
   



\subsubsection{Collective behavior, Flexibility \& Scalability}
After the three evolutionary experiments we re-test the single best controller over all 30 runs per arena (thus in total three different best controller). The best controller is tested in the same environment to evaluate if collective motion has emerged. We look at three different aspects of collective behavior over time: \textit{alignment} in terms of order (\autoref{eq:align_EC_s}, $\Phi$), \textit{performance} as mean scalar value of the swarm (\autoref{eq:r_i_sum_EC_s}, $f_t$), and \textit{collision} counted at every controller update. These metrics are plotted with respect to time.

Additionally, we investigate two types of swarm properties, namely flexibility and scalability \cite{csahin2004swarm}. Firstly, we (cross-)validate the overall best controllers per environment, meaning that the same swarm was re-tested in its own environment and the other two arenas for comparison. A similar performance in the other environments indicates flexibility of the found solution since changing environment sizes indicate a slower or faster change in the local values of the gradient for the same amount of displacement. Since the best controllers for each environment size is evolved to be the best with the corresponding rate of change of local values, different environment size and different rate of change of local values can be regarded as environmental variations. Secondly, to see the scalability of our control solution, we re-tested the overall best controller per arena, in the same arena with different swarm sizes (here we resized the initialization box accordingly). For both flexibility and scalability experiments we test the best controller 30 times per condition. 
The experimental parameters we used in the experiments are described in \autoref{tab:parameters_EC_s}.
\begingroup

\renewcommand{\arraystretch}{0.6} 

\begin{table}[htp!]
\footnotesize
\caption{Main experiment parameters}
\centering
\vspace{-1em}
\begin{tabular}[b]{{p{0.20\linewidth}| p{0.15\linewidth}| p{0.5\linewidth}}}
\toprule
\textbf{DE}         & Value & Description \\ \midrule
~Swarm size 			 & ~14     & ~Number of robots in a swarm 	\\
~Arena size 			 & ~10/30/45     & ~Swarm environment 	\\

~Population size  & ~25    & ~Number of individuals (swarms) per generation     \\
~Generations      & ~100   & ~Termination condition for each run             \\ 
~Mutation (F)         & ~0.5   & ~Mutation factor for individuals        \\ 
~Crossover (CR)         & ~0.9   & ~Probability of crossover for individuals        \\ 
~Evaluation time  & ~10    & ~Duration of an evaluation in minutes \\ 
~Repetitions      &  ~30    & ~Number of repetitions per experiment \\ 
\bottomrule 
\toprule
\textbf{Flexibility}         & Value & Description \\ \midrule
~Swarm size 			 & ~14     & ~Number of robots in a swarm 	\\
~Arena size 			 & ~10/30/45     & ~Swarm environment 	\\
~Evaluation time  & ~10    & ~Duration of an evaluation in minutes \\ 
~Repetitions      &  ~30    & ~Number of repetitions per experiment \\ 
\bottomrule 
\toprule
\textbf{Scalability}         & Value & Description \\ \midrule
~Swarm size 			 & ~5/14/50     & ~Number of robots in a swarm 	\\
~Arena size 			 & ~10/30/45    & ~Swarm environment 	\\
~Evaluation time  & ~10    & ~Duration of an evaluation in minutes \\ 
~Repetitions      &  ~30    & ~Number of repetitions per experiment \\ 
\bottomrule 
\end{tabular}
\vspace{-1.5em}
\label{tab:parameters_EC_s}
\end{table}
\endgroup

    

\section{Experiment Results}
All reported results and analysis data/scripts can be found in the following database: \href{https://doi.org/10.34894/IREUW1}{https://doi.org/10.34894/IREUW1}

\subsection{Efficacy}
We measure efficacy by the mean and maximum fitness averaged over the 30 independent runs at every generation. 

\begin{figure}[t!]
\centering
  \includegraphics[width=0.6\linewidth]{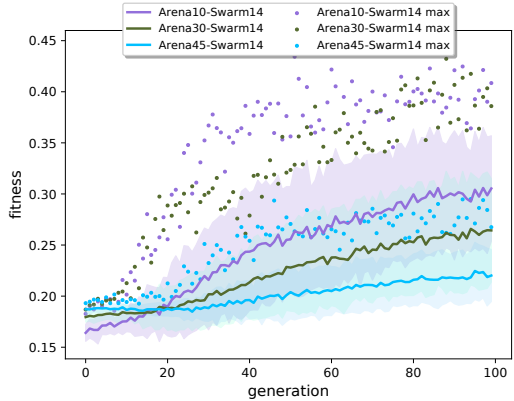}
  \captionsetup{width=.95\textwidth}
  \caption{The mean$\pm STD$ (line) and average max (dot) fitness over 100 generations (averaged over 30 runs) for all experimental condition: 10$\times$10: purple, 30$\times$30: green, 45$\times$45: blue.}
  \label{fig:fitness_EC_s}
\end{figure}

For each evolutionary experiment with swarm size 14 we run the learning task 30 times at three different arenas. \autoref{fig:fitness_EC_s} shows that the experiment in arena 45x45 has the highest average fitness at the start, followed by the experiment in arena 30x30. Then the fitness of all the evolutionary experiments increase steadily generation by generation. From around generation 28, the average fitness from arena 10x10 surpasses the other two, so does the max fitness. Similarly, around the same cross-over point, the average fitness from arena 45x45 becomes the lowest. In the end, the mean best solution over 30 runs were found in the 10x10 condition ($F_{10}=0.46 \pm 0.038$), followed by 30x30 ($F_{30}=0.39 \pm 0.055$), and 45x45 ($F_{45}=0.33 \pm 0.040$). Statistical analysis showed significant difference between all conditions ($p\ll 0.05$).

\begin{figure}[ht]
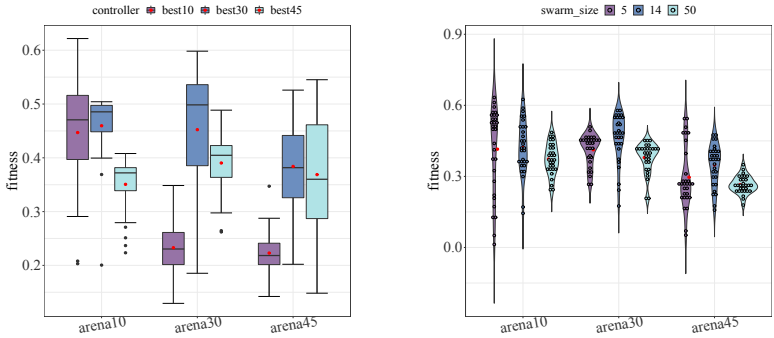

\centering
\begin{minipage}[c]{0.95\linewidth}
\centering
    \begin{minipage}[t]{0.425\linewidth}
    \subfloat[(a) Flexibility boxplot. Cross-validation of best controller in other arenas. Red dots show mean values.\label{fig:cross_validation}]{\includegraphics[width=\textwidth]{images/best_controller_cross_validation_new.pdf}}
    \end{minipage}
\hspace{2em}
    \begin{minipage}[t]{0.425\linewidth}
    \subfloat[(b) Scalability violin plot. Different swarm sizes (5: purple, 14: blue, 50: cyan). Each circle represents a single run with red dots as mean value.\label{fig:scalability}]{\includegraphics[width=\textwidth]{images/violin_scalability_new.pdf}}
      
    \end{minipage}
\end{minipage}
\captionsetup{width=.95\textwidth}
\caption{Retest experiments. Flexibility (left) and Scalability (right) were tested for 30 repetitions.}
\end{figure}

\subsection{Flexibility}
The final swarm controller should be able to operate despite variations in its environment. We re-evaluate the overall best controllers' fitness from each evolutionary optimization (3 arenas with swarm size 14) in 9 additional tests across three different arenas, namely 10x10m, 30x30m and 45x45m, with 30 repetitions.

\autoref{fig:cross_validation} shows that the best controller from the 10x10 evolutionary run (in color purple) performs best in arena10, however it significantly drops in performance when re-tested in the other two larger arenas ($p < 0.05$). For the 30x30 run (blue boxplots), the best controller performs similarly well in both the smaller arena10 and its own (arena30), while the fitness is lower in the bigger arena45. The last best controller from the 45x45 run (cyan boxplots) has a relatively steady performance across all three arenas. From this plot, we may conclude that the best controller from the biggest arena 45x45 is the most flexible one. During evolution, the most flexible controller was evolved in the hardest environment. Apparently controllers evolved in larger arenas, perform relatively (equally) well in smaller arenas, while the reverse is not the case.

\subsection{Scalability}
The swarm should be able to operate with a wide range of group sizes and support large numbers of individuals without impacting performance considerably. That is, the coordination mechanisms and strategies to be developed for swarm robotic systems should ensure cooperation of the swarm with varying swarm sizes. To measure the scalability property, we choose the best controllers from evolutionary experiments (Arena10,30,45 with swarm size 14) and apply them on three different swarm sizes --being 5, 14, 50-- within the same environment.

\autoref{fig:scalability} shows that all the best controllers are quite scalable across three arena with close mean fitness. Experiment with swarm size 14 (blue violins) has the highest mean fitness across three arenas. Experiment with swarm size 50 has a relatively lower standard deviation which indicates a more robust performance. In other words, a bigger swarm size creates a bigger sample of the map (because it has a bigger coverage) which makes the estimation of the gradient better and the performance more stable. 

\subsection{Collective behavior}
\begin{figure}
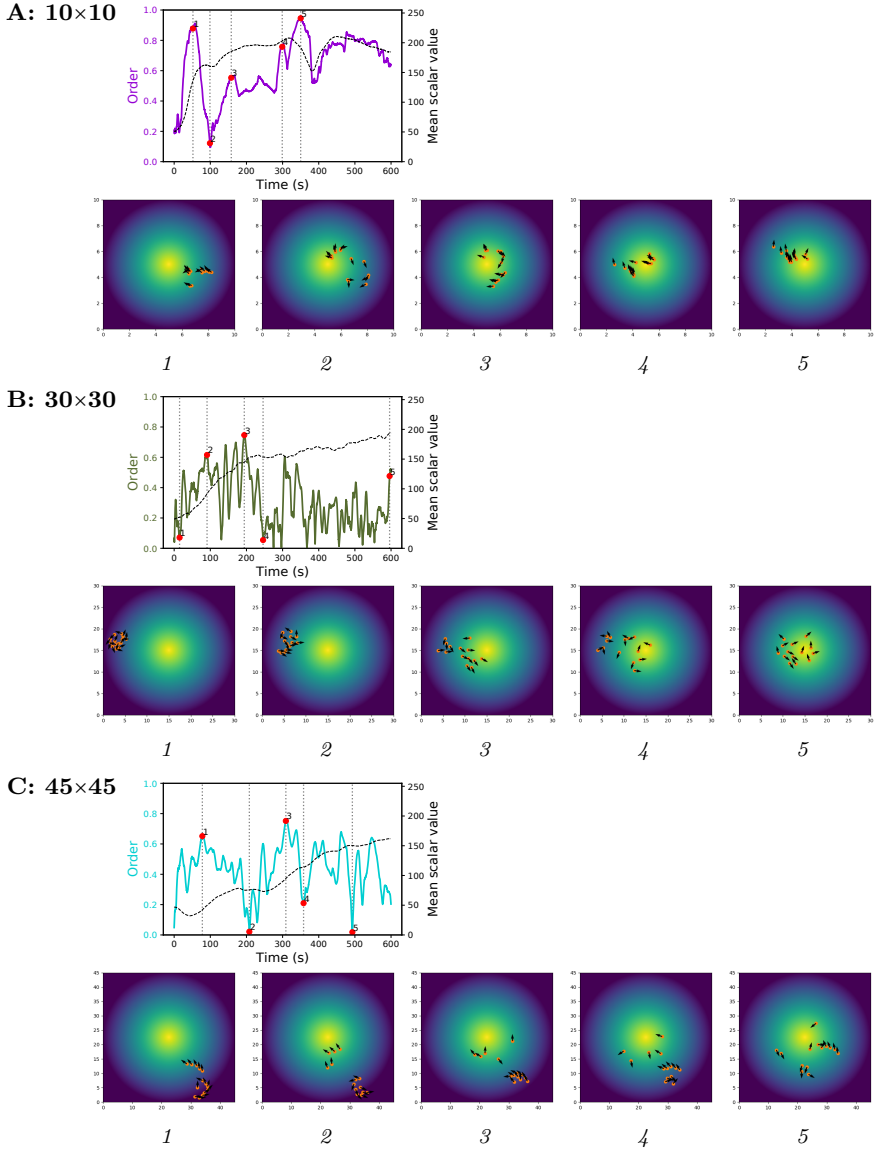
\small
    \centering
    \begin{flushleft}{\textbf{A: 10$\times$10}} \end{flushleft} \vspace{-2em}
        \begin{minipage}[t]{.375\textwidth}
        \hspace{-6.5em}
        \includegraphics[width=\textwidth]{images/10_snap/order_10x10.pdf}
        \end{minipage}
    \vspace{-1em}
    \begin{flushright}{
    \begin{minipage}[c]{0.9\textwidth}
        \begin{minipage}[t]{.185\textwidth}
            \centering
            \subfloat[1]{\includegraphics[width=\textwidth]{images/10_snap/1.png}}
        \end{minipage}
        \hfill
        \begin{minipage}[t]{.185\textwidth}
            \centering
            \subfloat[2]{\includegraphics[width=\textwidth]{images/10_snap/2.png}}
        \end{minipage}
        \hfill
        \begin{minipage}[t]{.185\textwidth}
            \centering
            \subfloat[3]{\includegraphics[width=\textwidth]{images/10_snap/3.png}}
        \end{minipage}
        \hfill
        \begin{minipage}[t]{.185\textwidth}
            \centering
            \subfloat[4]{\includegraphics[width=\textwidth]{images/10_snap/4.png}}
        \end{minipage}
        \hfill
        \begin{minipage}[t]{.185\textwidth}
            \centering
            \subfloat[5]{\includegraphics[width=\textwidth]{images/10_snap/5.png}}
        \end{minipage}
\end{minipage}} \end{flushright}\vspace{-1.5em}
\centering
    \begin{flushleft}{\textbf{B: 30$\times$30}} \end{flushleft} \vspace{-2em}
        \begin{minipage}[t]{.375\textwidth}
        \hspace{-6.5em}
            \includegraphics[width=\textwidth]{images/30_snap/order_30x30.pdf}
        \end{minipage}
        \vspace{-1em}
    \begin{flushright}{
    \begin{minipage}[c]{0.9\textwidth} 
        \begin{minipage}[t]{.185\textwidth}
            \centering
            \subfloat[1]{\includegraphics[width=\textwidth]{images/30_snap/1.png}}
        \end{minipage}
        \hfill
        \begin{minipage}[t]{.185\textwidth}
            \centering
            \subfloat[2]{\includegraphics[width=\textwidth]{images/30_snap/2.png}}
        \end{minipage}
        \hfill
        \begin{minipage}[t]{.185\textwidth}
            \centering
            \subfloat[3]{\includegraphics[width=\textwidth]{images/30_snap/3.png}}
        \end{minipage}
        \hfill
        \begin{minipage}[t]{.185\textwidth}
            \centering
            \subfloat[4]{\includegraphics[width=\textwidth]{images/30_snap/4.png}}
        \end{minipage}
        \hfill
        \begin{minipage}[t]{.185\textwidth}
            \centering
            \subfloat[5]{\includegraphics[width=\textwidth]{images/30_snap/5.png}}
        \end{minipage}
\end{minipage}} \end{flushright}\vspace{-1.5em}
\centering
    \begin{flushleft}{\textbf{C: 45$\times$45}}\end{flushleft} \vspace{-2em}
        \begin{minipage}[t]{.375\textwidth}
        \hspace{-6.5em}
        \includegraphics[width=\textwidth]{images/45_snap/order_45x45.pdf}
        \end{minipage}
    \vspace{-1em}
    \begin{flushright}{
    \begin{minipage}[c]{0.9\textwidth} 
        \begin{minipage}[t]{.185\textwidth}
            \centering
            \subfloat[1]{\includegraphics[width=\textwidth]{images/45_snap/1.png}}
        \end{minipage}
        \hfill
        \begin{minipage}[t]{.185\textwidth}
            \centering
            \subfloat[2]{\includegraphics[width=\textwidth]{images/45_snap/2.png}}
        \end{minipage}
        \hfill
        \begin{minipage}[t]{.185\textwidth}
            \centering
            \subfloat[3]{\includegraphics[width=\textwidth]{images/45_snap/3.png}}
        \end{minipage}
        \hfill
        \begin{minipage}[t]{.185\textwidth}
            \centering
            \subfloat[4]{\includegraphics[width=\textwidth]{images/45_snap/4.png}}
        \end{minipage}
        \hfill
        \begin{minipage}[t]{.185\textwidth}
            \centering
            \subfloat[5]{\includegraphics[width=\textwidth]{images/45_snap/5.png}}
        \end{minipage}
\end{minipage}\vspace{-1em}} \end{flushright} 
\captionsetup{width=.95\textwidth}
    \caption{\footnotesize Flocking behavior of the final controller re-tested in the same environment. The three line plots on the left show the correlation between order (alignment) and mean scalar value (performance). The snapshots on the right are corresponding to each line plot. The dashed black lines show the mean scalar value and the colored lines show the order value. Distinct peaks in alignment are highlighted with numbers 1-5. Please note the different scales on each map.}
    \label{fig:order_EC_s}\vspace{-1em}
\end{figure}

\autoref{fig:order_EC_s} shows the correlation of alignment (Order, $\Phi$) within the swarm members and their performance (Mean scalar value, $f_t$) over time. We observe that when order peaks, i.e. namely red point 1, 3, 4 and 5 in arena 10x10m, red points 2, 3 and 5 in arena 30x30m and red points 1,3 in arena 45x45m, the mean scalar value increase (i.e. the tangent of the dashed black line is highly positive). Vice versa, when the colored lines reach to lower order value, the mean scalar value decrease or stay the same (i.e. tangent is near zero). 

The snapshots on the right are correspond with the same numbered points in the order/mean scalar value plot. We can see that in most cases when a swarm is becoming more aligned (i.e. order is high), the direction of movement is similar to the direction of increasing gradient (i.e. towards the center). In the low alignment case (i.e. low order) most of the members disperse in different directions, causing no net difference in the mean scalar value.

Impressively, here we can see a collective `search-like' behavior emerge that is reminiscent of an optimization algorithm trying to maximize a function. Scalar values are periodically and collectively sensed by the swarm. When the order is low, the swarm samples in different direction for higher values i.e. "exploration". In contrast, during the high peaks the swarm collectively moves in the direction of the gradient which causes the mean scalar value to increase i.e. "exploiting". These exploration exploitation states are periodically changed to escape local optima and find a global maximum. Even more exciting, when mean scalar value plateaus (i.e. swarm arriving near maximum) there is a clear switch in behavior and a new pattern emerges. In the 10x10 the swarm approaches the center and overshoots it, directly followed by a fast recovery toward the center. For 30x30 controller, the agents randomly spread around the maximum, slowly creeping inward. Unfortunately, the 45x45 experiment run was too short to see a final pattern emerge.

\begin{figure}[ht]
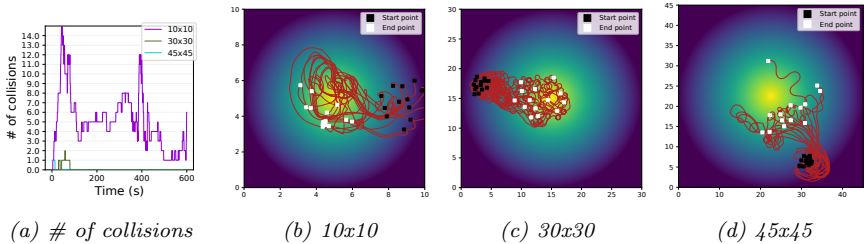

     \centering
     \begin{subfigure}{0.23\textwidth}
         \centering
         \includegraphics[width=\textwidth]{images/no_of_collisions.pdf}
         \caption{(a) \# of collisions}
         \label{fig:collis}
     \end{subfigure}
     \hfill
     \begin{subfigure}{0.24\textwidth}
         \centering
         \includegraphics[width=\textwidth]{images/traj_10x10.pdf}
         \caption{(b) 10x10}
         \label{fig:10x10}
     \end{subfigure}
     \begin{subfigure}{0.24\textwidth}
         \centering
         \includegraphics[width=\textwidth]{images/traj_30x30.pdf}
         \caption{(c) 30x30}
         \label{fig:30x30}
     \end{subfigure}
     \begin{subfigure}{0.24\textwidth}
         \centering
         \includegraphics[width=\textwidth]{images/traj_45x45.pdf}
         \caption{(d) 45x45}
         \label{fig:45x45}
     \end{subfigure}
     \caption{No. of collisions and trajectories of the best controller per arena re-tested in the same environments. Please note the different scales in subplots b, c, d.}
     \label{fig:trajectories}     \vspace{-1.5em}
\end{figure}

\autoref{fig:trajectories}-a shows the the number of collisions over 600s timestamps in three arenas. We observe that the smaller arena size, the higher no. of collisions. Moreover, at the start of each run the swarm is very chaotic and collisions occur in all environments. Fortunately, it can be observed that the swarm resolves these collision itself. In addition, occasional group splits can also be observed. Since neither the ordered swarm behavior is the main objective nor controllers are manually designed for it, we expect to see this splits time to time. What is interesting is that despite this splits, clusters continue to behave in the same way and eventually merge with the rest under the influence of the commonly sensed environmental feature.  \autoref{fig:trajectories}-(b,c,d) shows the trajectories of the best controller per arena re-tested in the same environments. The black blocks are the starting points of a swarm and the white blocks are the ending points. All of them are following the gradient well. They start with circling around, then moving towards the brighter area once they sense the different gradients. When they reach the brightest zone (yellow), they start circling again. A video of these behaviors can be found at \url{https://youtu.be/yhKFvpLa9iI}.

\section{Discussion}
The goal of this paper was to optimize the controllers of robots in a swarm for a task solely related to an environmental property and test whether a collective behavior emerges. Optimizing swarm behavior is often hard and requires vast knowledge on the specific task and the limitations of its members. Handwritten controllers therefore require a long process of prototyping which is sensitive to human biases, and limitation in engineered metrics for `good' collective behavior. In our setup we showed that even with a very simple task description, good performance can be reached with learned coordinated swarm behavior as an emergent property. 

From the results, the final best controller experiments showed that evolving in a bigger environment improves the controller in terms of flexibility. We suspect this to be caused by a less pronounced gradient in the bigger arena making the task more difficult. In other words, when evolved in a bigger arena a swarm needs to sample a bigger area to be able to sense the increasing gradient. This is because ratio between the swarm area and the total scalar field area is the smallest in the 45x45m environment. To cope with less salience of local values differences, the 45x45m flocks adopts a line formation strategy covering a big circle with more area (see \autoref{fig:order_EC_s}-C1). This is slightly different than the spiral formation seen in \autoref{fig:order_EC_s}-B1 employed by swarms evolved in the other arenas, and seem to maintain high fitness values when cross-validated. These behaviors can also be seen in the swarm trajectories (\autoref{fig:trajectories}). Lastly, we also see different behavior back in a less aligned rotation (low order) and more aligned forward motion (high order) with the best controller evolved in the largest environment. 

The alignment within the swarm members (\autoref{fig:order_EC_s}) provides some interesting information on the swarm behavior. Here, we see a correlation between increasing order value and increasing mean scalar value seen by the swarm. Vice versa, lower order value correspond to a plateau/decrease in mean scalar value. This indicates that, when a swarm is becoming more aligned, the direction of movement is similar to the direction of increasing gradient. 

As a final remark, for our specific optimization we wanted the swarm to sense the increasing gradient (while a single one is not capable of doing it) and move towards the increasing local values. It should be noted that such a fitness could be any function that we want. This opens up the possibility to ignore standard engineering practices that enforce certain swarm behavior, optimize flocking rule parameters or design complex controllers --all of which may only be optimal for a specific task-- and continuously update our swarm as an adaptive system. The model-agnostic nature of a NN with an EA enables the emergence of different swarm strategies that we don't have to pre-define (e.g. different formations, sub-group sizes, search behavior, etc.). 

\section{Conclusions and Future Work}
We have successfully developed an evolutionary system for the automated design of controllers for robot swarms. As explained in the Introduction, this is a highly nontrivial feat because of the long and complex chain between genotypes and fitness values. To test our pipeline we evolved several swarms for the task of gradient following. Here, we found that the environmental property induced the emergence of collective behavior, a result that was not explicitly formalised in the objective function. To our best knowledge, there is only one other study that has achieved this \cite{witkowski2016emergence}, but we should note that it considered simulated point agents with signalling abilities which significantly simplified the task. Ongoing work is devoted to additional validation of our results using real robots. 

Interesting directions for future research include investigating more and different gradient landscapes in addition to the circular and unimodal one used in this study. Different gradient landscapes might be circular as well but with a multimodal structure or they can be in a linear form instead of circular. Moreover, instead of having a homogeneous swarm in terms of sensing capabilities, more investigation can be conducted where the sensing abilities of the individual robots differ (some may have more or better sensors than others, while using the same generic controller). This investigation can even be stretched by extending the evolutionary approach to swarms where the controllers are not necessarily identical as they are in this study.   





\section{Introduction}
Collective motion is widely documented in groups of animals in nature, and has shown to enhance the group with capabilities that are not apparent in an individual group member. Those emergent capabilities include increased environmental awareness \cite{couzin2005-animalgroups,kearns2010-bacterialswarming}, safety from predators \cite{olson2013-predatorconfusion}, and gradient sensing \cite{puckett2018collective}. Swarm robotics aims at implementing such collective behaviors in robots, in order to leverage those advantageous emergent properties for engineering purposes. In particular, collective motion can enable robotic swarms to achieve sensing beyond the capabilities of the sensors fitted on individual agents \cite{karaguzel2020collective}, i.e. emergent perception.

The main challenge in designing collective motion behaviors, and swarm robotics systems in general, comes from the fact that designers can only implement robot controllers on individual robots, but that the desired behavior is defined at the group level \cite{hasselmann2021empirical}.
Successful group behaviors therefore depend on emergent properties, which are difficult to predict. 
An approach to this problem is to define a metric of success on the group level, and to use it as reward to optimize robot controllers automatically \cite{trianni2014evolutionary,francesca2014automode,hasselmann2021empirical}. 
However, the solutions that emerge through this process tend to overfit their training environment and, consequently, lack flexibility. 
Automated design therefore requires a framework that maintains good performance for a variety of environmental conditions. For example, modularity of the individual-level control has been proposed as a solution to increase flexibility \cite{francesca2014automode}.

We believe that, by leveraging on heterogeneity, we can achieve a modular framework at the swarm level, as opposed to the prevailing `homogeneous designs' typically found in the swarm robotics literature \cite{trianni2014evolutionary,francesca2014automode,hasselmann2021empirical}. 
Heterogeneous collective behaviors are widely observed in nature. 
For example, social insects divide their tasks in sub-tasks (the so-called division of labor), assigned to specific group members, in order to improve their efficiency \cite{ratnieks1999task}. 
Indeed, distinct sub-group behaviors can emerge when task specialization is beneficial for the group as a whole \cite{ioannou2017swarm}. 
Maintaining a `group identity' whilst splitting in sub-groups can be achieved through genetic homogeneity with phenotypic heterogeneity.
For example, within an insect colony, members share similar-to-identical genotypes \cite{ravary2007individual}; 
The various behaviors used by the insects are therefore all encoded in the same genotype, and are activated by external cues, e.g. queen pheromones \cite{holman2018queen}. 
This phenomenon, where the same genotype expresses a different phenotype by gene regulatory mechanisms, is referred to as \textit{phenotypic plasticity}. 
Although it is more commonly useful for task partitioning \cite{ratnieks1999task, ferrante2015evolution}, it has also been found in species exhibiting collective motion \cite{ariel2015locust}.

In this paper, we propose to improve the flexibility of automated designs by promoting modularity at the swarm level through heterogeneity in self-organized collective behaviors (as apposed to individual-level).
Our framework is based on phenotypic plasticity, where we evolve separate controllers for sub-groups inside the swarm, and, through a regulatory mechanism, adjust the phenotypic ratio of each group. In the end, we regulate the modular controller phenotype during deployment using a distributed online switching mechanism. 

\section{Related work}\label{sec:rel_work}
Behavioral heterogeneity induces several challenges, the first of which is task allocation, i.e. dynamically adjusting the quantity of agents assigned to each available task \cite{bayindir2016review}. The problem of achieving this goal in a decentralized manner has been widely addressed by swarm robotics. Approaches to this problem usually focus on the mechanism of task switching, and therefore use relatively simplistic, manually implemented behaviors. 
With threshold-based responses, robots initially show some preference for a given task, while simultaneously recognizing deficiencies in the accomplishment of other tasks (e.g. objects accumulating). Past some threshold, the robot switches to the corresponding task \cite{krieger2000call}. 
In some cases, this switch is probabilistic in order to avoid large-scale population switches, that might leave another task unaddressed \cite{brutschy2014self}. Mathematical modelings of task allocation have also been proposed \cite{valentini2022global}, which allows custom task allocation parameters according to their need. 

In addition to task allocation, task specialization focuses on the emergence of several complementary functions within the swarm. It often relates to physically heterogeneous robots, i.e. multi-robot systems composed of multiple robot platforms \cite{rizk2019cooperative}. It should be noted that this impairs a vital advantage of swarms, namely robustness as, in such cases, the individual members are not interchangeable with one another. A more robust design is when agents are physically identical but differ in behavioral function \cite{bettini2023heterogeneous}.
Functional heterogeneity through behavioral specialization does not suffer from this issue \cite{hussein2022autonomous}. Here, it is possible for robots with the same body to switch behavior while employed, also referred to as task partition. In such a context, maintaining group level collective behavior can be tricky as online regulatory mechanisms tend to switch individual behaviors only \cite{feola2023aggregation}.

Tuci et al. investigated the evolution of task partition (i.e. both task allocation and specialization), with a \textit{clonal} and \textit{aclonal} evolutionary process, in a physically homogeneous swarm of 5 e-pucks. Here \textit{clonal} refers to a single genotype being shared by all the agent, whereas, with \textit{aclonal}, their genotypes are all different from each other. Unsurprisingly, they found that robots performed better in their \textit{aclonal} approach \cite{tuci2013evolution, tuci2014evolutionary}, especially if their controller was optimized with a multi-objective fitness \cite{tuci2015design}, as they could address the required sub-tasks in parallel. Notably, this approach results in an efficient optimization process, as each agent in the swarm samples a different genome. Moreover, plasticity was observed, in the sense that individual robots were able to switch tasks according to environmental requirements (including their peers' behaviors).

Closely related, \cite{ferrante2015evolution} also addressed task partition, with a task that evokes the environmental context of leafcutter ants who divide their foraging task into two sub-tasks: cutting and dropping leaf fragments into a storage area, on one hand, and collecting and bringing the fragments back to the nest on the other. Their experimental environment was composed of a slope separating a nest (below) and a source (above) area so that the robots could individually retrieve the food objects back to the nest, or deposit them on the slope and relying on other robots to fetch them. Task allocation was evolved on a probabilistic finite state machine, composed of simplistic pre-programmed behaviors, without specifying a preference for collective, rather than individual behavior. Their success demonstrates that even homogeneous controllers (i.e. a single phenotype) can handle task partition through individual experience, stigmergy and stochastic switching alone, given enough knowledge on the task to design viable behaviors. 

We aim to make collective specialized behavior emerge, without any explicit reward on the specific sub-tasks. Differently from the aforementioned works, our approach require minimal insight about the solution as we do not pre-define various subtasks/behaviors, such as bucket brigading. Instead, we define a reward on the overall group level performance of the whole task, and let specialization evolve as an optimal solution. This simplifies the design of a good performance metric, as it alleviate the necessity to define apriori what `good' specialized behavior constitutes. 
Furthermore, our method is more flexible than a modular design with pre-defined specialized behaviors that presuppose sub-group interactions affecting the overall swarm performance. Additionally, we automatically obtain task allocation through a heuristic approach that designs our online regulatory mechanism. This straightforward method minimizes the time and effort required, while demonstrably enhancing overall task performance. Altogether, this work shows that optimal task partition can be designed automatically using evolutionary computing, without any specific sub-task knowledge. To the best of our knowledge, no previous work exists that address task specialization and allocation in such a context.

\section{Methodology}\label{sec:meth}
Optimizing a heterogeneous swarm controller without specific knowledge on any sub-task requires a flexible approach that can be applied on any type of task. We test our method in an emergent perception task for gradient sensing where robots have to find the brightest spot in the center (inspired by fish behavior described in \cite{puckett2018collective}). For this, we utilize black box optimization in the form of an evolutionary algorithm on Reservoir Neural Networks (RNN), a method that has been employed on homogeneous swarms in a similar capacity \cite{van2022environment}. 
Full code base can be found in the following git: 
\href{https://github.com/fudavd/EC\_swarm/tree/PPSN\_2024}{https://github.com/fudavd/EC\_swarm/tree/PPSN\_2024} 

\subsection{Robot design}
Each robot in the swarm has an identical differential drive hardware design based on the Thymio II without any communication capabilities (Bluetooth, radio, WiFi). Our robot consists of a cart with two actuated wheels (max speed is $\pm \unit[14]{cm/s}$) in the back and a single omni-directional wheel in the front. We equip our robots with range and bearing sensing in 4 directions (specifics are detailed in \autoref{sec:Control}) and a local value sensor to measure local light intensity. These sensors are sampled at 10Hz to obtain control inputs only based on current information, i.e. no memory of previous state. 

It is important to be explicit on the capabilities of our robot design: 1) Robots do not communicate information to each other (e.g. local values at their position, future motor inputs, or any form of message passing); 2) The controller is memoryless, only current local sensor readings are known; 3) Their is no notion of specialization inside the controller, meaning robots are `unaware' of specialization inside the swarm. All in all, this grounds the idea of `limited sensing', as a single robot is incapable of estimating the gradient of the light.

\subsection{Controller design}\label{sec:Control}
For general applicability, we require our controller to be as flexible as possible while capable to learn quickly. For this we opt to use neural networks (which are expressive function approximators) with random functionalities in the form of a reservoir to speed up learning (i.e. RNN, \cite{lukovsevivcius2009reservoir}). This reservoir is created by freezing the network weights up to the last layer after random initialization, resulting in a fixed set of functions, of which we learn an optimal combination in the final network layer. 

To allow specialization, we divide our swarm into two sub-groups, with each sub-group containing a different RNN controller (all sub-group members have the same RNN). The two RNNs are randomly initialized with different reservoirs that we save at the start. We describe the swarm genotype as a single vector of weights from which the first half refers to last layer of the first RNN, and the second half to the last layer of the second RNN. Phenotypic plasticity of our single genome is expressed through our sub-group division.

\begin{figure}[hpt!]
  \includegraphics[width=0.8\linewidth]{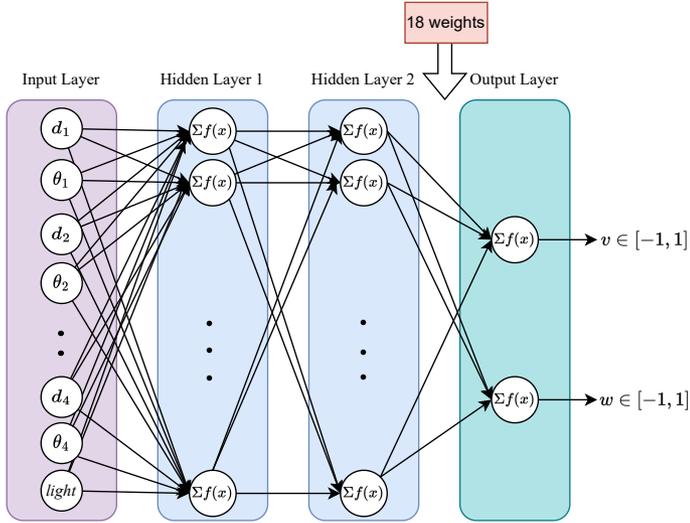}
  \centering
  \caption{\small Reservoir Neuron Network controller design.}
  \label{fig:NN_Special_s}
\end{figure}

The phenotype of the controller is illustrated in \autoref{fig:NN_Special_s}. The RNN has an input layer of 9 neurons that are re-scaled to [-1, 1], namely 4 directional sensors (each providing two values: a distance and a heading) and 1 local value sensor. The 4 directional sensors cover a combined 360 degrees view of the robot's surrounding (Front-, Back-, Left-, and Right- quadrant of 90 degrees each). Within each quadrant ($i$) the sensor obtains the distance ($d_i$) and relative heading ($\theta_i$) of the nearest neighbor up to a maximum range of 2 meter (outside of this range the sensor defaults $d_i$=2.01 and $\theta_i$=0). The RNN outputs target speed ($v\in[-1,1]$) and angular velocity ($w\in[-1,1]$), which are transformed to direct velocity commands for the two wheels.

The RNN architecture is a fully connected neural network with an input layer with 9 neurons ($\mathbf{s}_{in} \in [-1,1]^9$), 2 hidden ReLU layers of the same size (${h}_{1}, {h}_{2} \in \mathbb{R}^9$), and a final output layer with two tanh$^{-1}$ neurons, $RNN \in [-1,1]^2$. All hidden reservoir weights are randomly initialized with uniform distribution ($U[-1,1]$). We set all biases to 0 and only optimize the weights of the output layer during evolution (18 weights per RNN). The final RNN controller can be formalized as follows:

\begin{align*}
    RNN = \mathrm{tanh}^{-1}\left(\mathbf{W}_{out}\mathrm{ReLU}\left(\mathbf{W}_{h2}\mathrm{ReLU}\left(\mathbf{W}_{h1}\mathbf{s}_{in}\right)\right)\right) \\
    \text{$\textrm{with,} \quad  \mathbf{W}_{h1, h2} \in \mathbb{R}^{9\times9} \quad \textrm{and} \quad \mathbf{W}_{out} \in \mathbb{R}^{2\times9}$ }
\end{align*}

\subsection{Emergent perception task}
We aim to enhance our robots' sensing capability such that, when operating collectively as a swarm, they can perceive the gradient. Our setup consist of 20 robots that are rewarded for navigating to the brightest spot in the center of a 30x30m arena. Since an individual robot lacks the capacity to sense the direction of the gradient field, we anticipate a collective behavior to emerge. This `task' is not too dissimilar to a behavior found in schools of fish that tend to aggregate in the shadow as it decreases their visibility from predators \cite{puckett2018collective}.

\begin{figure}[ht]
\vspace{-0.5em}
  \centering
\begin{minipage}[c]{0.85\linewidth}
\centering
        \begin{minipage}[t]{.4\linewidth}
            \centering
            \subfloat[(a) Scalar field map]{\includegraphics[width=\textwidth]{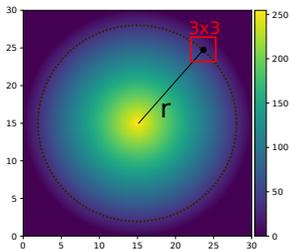}}
        \end{minipage}
\hspace{2em}
        \begin{minipage}[t]{.4\linewidth}
            \centering
            \subfloat[(b) Isaac gym]{\includegraphics[width=\textwidth]{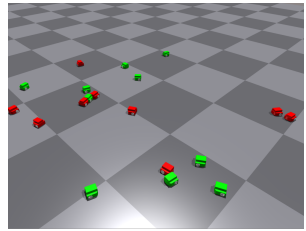}}
    \end{minipage}
\end{minipage}
\caption{\small The task environment. (a) The scalar map indicating a random instance of our swarm task. (b) Our swarm with different sub-groups (colored red and green) in simulation.}
\label{fig:swarm_env_Special_s}
\vspace{-1em}
\end{figure}

We use Isaac Gym from \cite{makoviychuk2021isaac} to simulate our swarm(s) ($\mathrm{dt=0.05s}$). The task environment is a scalar field map with its maximum value (255) in the center (see \autoref{fig:swarm_env_Special_s}a). We randomly place the swarm on a circle at a fixed distance (r=12m) away from the center. At this position, we randomly place each swarm member within a 3x3m bounding box (shown in red). The swarm is split into two sub-groups (red and green) of 10 members each as shown in \autoref{fig:swarm_env_Special_s}b. 

\begin{figure*}[ht] 
  \centering
  \includegraphics[width=\textwidth, trim={0 35px 0 105px},clip]{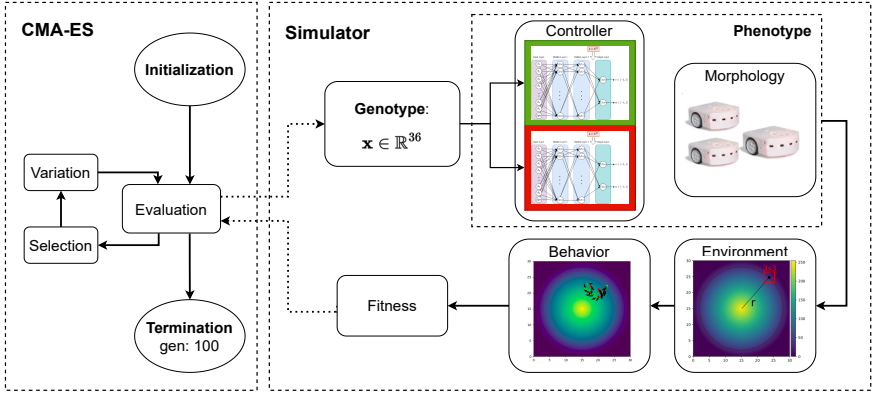}
  \vspace{0em}
  \caption{\small Experimental setup for optimizing swarm controllers, where an evolutionary algorithm evaluate different genotypes in our swarm simulator (big dashed box). Please note, our genotype being split up in 2 different controllers colored as green and white boxes.}
  \label{fig:swhole_process_Special_s}\vspace{-1em}
\end{figure*}

\subsection{Evolving swarm experiment}
We optimize our swarm using Covariance Matrix Adaptation Evolution Strategy (CMA-ES, \cite{hansen2001cmaes}). We run 10 repetitions of our experiments to obtain the overall best controller.

Let us draw a clear distinction between the components within \textit{CMA-ES}, where individuals in a population are being optimized, and our \textit{swarm} which refers to an instance of an individual, consisting of robots/members that are assigned to different a sub-group (see \autoref{fig:swhole_process_Special_s}).
Within our evolutionary algorithm we thus have individuals that we want to evaluate.
All individuals inside this evolving population have two RNNs with the same two reservoirs.
The differences between individuals is defined by their genotype (a vector of 36 weights, $\textbf{x} = \left[\mathbf{W}_{out_{1:}}, \mathbf{W}_{out_{2:}}\right]$ with, $\textbf{x} \in \mathbb{R}^{36}$) that encode the last layer weights of the two RNNs. 
We evaluate our individual by assigning the RNNs to two sub-groups in a single swarm of robots, where each sub-group member (i.e. a robot belonging to a specific sub-group) has the same RNN as the other constituents. After a trial we calculate a fitness value based on task performance of the swarm, and assign it to the corresponding individual. 

CMA-ES is a sampling based evolutionary strategy that aim to find a distribution in search space to sample high performing individuals with high probability. Pseudo-code is shown below \autoref{alg:cmaes}, actual implementation is in python using \cite{nikolaus_hansen_2023_7573532}. 
Here, CMA-ES samples new candidates $\textbf{x}$ according to a multivariate normal distribution. 
The covariance matrix of the sampling distribution is updated at each generation ($\textbf{C}_{gen}$) to increase the likelihood of sampling an individual with higher fitness. 

\begin{algorithm}[ht]
\SetAlgoNlRelativeSize{0}
\caption{\small CMA-ES Pseudocode\label{alg:cmaes}}
\footnotesize
\KwData{Initial mean $\mathbf{m}_0 = 0$, initial covariance $\sigma_0$, population size $\lambda = 30$ solutions $\{\mathbf{x}_1, \mathbf{x}_2, \ldots, \mathbf{x}_\lambda\}$ with genotype $\mathbf{x}_i \in \mathbb{R}^{36}$}
\KwResult{Optimal solution $\mathbf{x}^*$}

$N_{gen}\leftarrow 100$

$N_{repeats}\leftarrow 3$

\For{$gen$ in $N_{gen}$}{
  \For{$\mathbf{x}_i$ in  $\lambda$}{    
    \For{$rep$ in $N_{repeats}$}{
        $f_{rep} \leftarrow \textrm{SIMULATE}(\mathbf{x}_i)$ \hfill {\textit{simulate for }10min}
    }

    $F_i\leftarrow \textrm{MEDIAN}(f_1, \dots, f_{N_{repeats}})$ \hfill \textit{Final fitness}
  }
  Sort solutions by fitness
  
  Select the top-performing solutions to form a weighted mean
  
  $\mathbf{m}_{t+1} \leftarrow \sum_{i=1}^{w} w_i \mathbf{x}_i$, where $w_i$ are weights
  
  Update covariance matrix $\mathbf{C}_{gen+1}$
  
  Generate a new population covariance matrix $\mathbf{C}_{\text{new}}$ based on the selected solutions
  
  $\mathbf{C}_{gen+1} \leftarrow \mathbf{C}_{\text{new}}$
  
  Sample $\lambda$ solutions $\{\mathbf{x}_1, \mathbf{x}_2, \ldots, \mathbf{x}_\lambda\}$ using $\mathbf{C}_{gen+1}$
}
$\mathbf{x}^* \leftarrow$ Best solution found
\Return{$\mathbf{x}^*$}
\end{algorithm}

We set our population size to 30 individuals (i.e. 30 different swarms), and evolve for 100 generation. Every individual is tested thrice with the median performing run representing the final fitness. This reduces the sensitivity to lucky runs that are non-repeatable and therefore non-viable. 

\begin{table}[htp!]
\footnotesize
\caption{\small Evolving swarm experiment parameters}
\centering
\vspace{-0.5em}
\begin{tabular}[b]{{p{0.19\linewidth} p{0.12\linewidth} p{0.53\linewidth}}}
                 & Value           & Description\\
\toprule
Runs      &  10             & Number of runs of our experiment \\ 
\midrule
\multicolumn{3}{l}{\textbf{Learning task}: \textit{collective gradient sensing}} \\
\midrule
Swarm size   & 20    & Number of robots in a swarm \\
Ratio        & 1:1   & sub-group division          \\ 
r        & 12   & Spawn distance from center (meters)          \\ 
Arena type  & center     & Environment: center	    \\
Eval. time   & 10    & Test duration in minutes    \\ 
\midrule
\multicolumn{3}{l}{\textbf{Optimizer}: \textit{CMA-ES}}     \\
\midrule
genotype      & $U[-5,5]$ & Initial sampling $\mathbf{x}_{init} \in \mathbb{R}^{36}$   \\
$\lambda$   & 30        & Population size          \\
$N_{gen}$ & 100       & Termination condition            \\ 
$\sigma_0$  & 1.0       & Initial covariance value              \\ 
$N_{repeats}$     & 3         & Number of repeats per individual \\
\bottomrule 
\end{tabular}
\vspace{-1.5em}
\label{tab:parameters_Special_s}
\end{table}

\subsubsection{Fitness function}
We define the fitness of a swarm as generally as possible, solely based on the ability to follow the increasing gradient of the scalar field defined by the environment (shown in \autoref{fig:swarm_env_Special_s}-a). This is done by aggregating the average light intensity values of all members over time (see \autoref{eq:r_i_sum_Special_s}). 

\begin{equation}
{f} = \frac{\sum_{t=0}^{T}{l_t}}{G_{max}\cdot{T}}\quad\text{and}\quad{l}_t = \frac {\sum_{n=1}^{N}{G_n}}{N}
\label{eq:r_i_sum_Special_s}
\end{equation}

Where $G_n$ is the scalar value of the grid cell in which agent $n$ (of all agents, $N$) is located at a time $t$. Thus the fitness at a specific time ($l_t$) is calculated as the mean scalar light value of all swarm members. Trial fitness ($f$) is calculated by averaging all $l_t$ over total simulation time $T$. At last, we normalize using the maximum scalar value $G_{max}$, always equal to $255$ for all experiments. A theoretical maximum fitness of 1 can only be achieved if all swarm member would stack up at the center for the full duration of the run. Fitness is only evaluated on swarm level with no distinction between sub-groups or any other task-specific principles that promotes specialization. Additionally, we would like to emphasize that our fitness function does not distinguish between robots that are sensing and following the increasing gradient collectively or as solitaries, or sub-groups that cooperate or not.

\subsection{Validation experiments}
After our evolutionary experiment we obtain the overall best controller over 10 runs. We re-test this controller to test the emergent perception capability of the swarm to sense the gradient. Our validation is split in two: 1) We are particularly interested in the possible collective (sub-group) behavior(s) and the importance of sub-group interactions, which we elucidate by re-testing the best evolved controller and analysing the swarms behavior; 2) Using our two RNNs, we implement a straightforward online regulatory mechanism to mimic phenotypic plasticity induced by pheromones \cite{holman2018queen}. We investigate the viability of our (adaptive) heterogeneous swarm through scalability and robustness experiments.

\subsubsection{Collective behavior \& sub-group interactions}
First, we assess whether collective motion has emerged in a single re-test. We look at two different aspects of collective behavior over time: (1) \textit{performance} as mean scalar light value of the swarm (\autoref{eq:r_i_sum_Special_s}, $l_t$); (2) \textit{alignment} in terms of order (\autoref{eq:align_Special_s}, $\Phi$), which is defined as follows: 

\begin{equation}
\begin{aligned}
{\Phi} = \frac{\sum_{n=1}^{N}{\varphi_n}}{N}\quad 
\text{and}
\quad{\varphi_n}=\frac{\Big|\Big|\left(\sum_{p=1}^{P}{\angle{e^{j \theta_p}}}\right)+\angle{e^{j \theta_n}} \Big|\Big|}{P+1}
\end{aligned}
\label{eq:align_Special_s}
\end{equation}

Here, $\varphi_n$ defines the order value calculated for agent $n$. Which is the average current heading direction of agent $n$ (noted as $\angle{e^{j \theta_n}}$) and all its perceived neighbors $P$ (noted as $\angle{e^{j \theta_p}}$). The total swarm order $\Phi$ is then defined as the average $\varphi_n$ over all agents in the swarm. Order measure gives a powerful insight about the alignment of agent's direction of motion. If all agents move towards the same direction, then the order measure approaches to 1; and if they move in different directions, order approaches to 0.

Additionally, we investigate the benefits of sub-group interactions on two factors, namely performance and robustness. We retest our best controller in the same environment where we change the sub-group ratios (ratio $\in$ \{4:0, 3:1, 2:2, 1:3, 0:4\}). Different sub-group ratios can tell us if one sub-group is mainly responsible for the swarm performance or if sub-group interaction are important. These different ratios are tested by initializing the swarm at different distances away from the center (r$_{ratio}$ $\in \left\{ 0,0.25,0.5,0.75,1\right\}$ as a ratio of the original training distance, 12m).

\subsubsection{Online regulatory mechanism}
Based on the results of the sub-group interaction experiments we can heuristically find the best performing sub-group ratios at certain light intensities. Subsequently, we create a probabilistic finite state machine where the choice of sub-group controller within a robot is defined such that, on a holistic group level, the optimal sub-group ratios should emerge. This probability is only dependent on the local light value and thus no communication is implemented to adapt to the best ratio. For example, if we heuristically find a 1:3 ratio to be the best sub-group division at the current local light intensity, the probability to sample an action from the second reservoir is 75\%. We update the probabilistic reservoir state every 5 seconds for stable behavior (this update frequency is found to be optimal using grid search $\{1, 5, 10, 30, 60, 100\}$ seconds). 

\begin{figure}[ht]
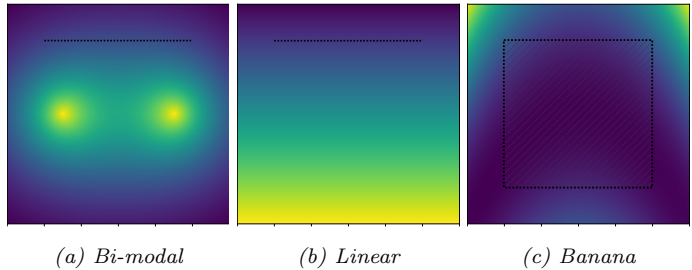

\vspace{-0.5em}
  \centering
\begin{minipage}[c]{0.85\linewidth}
\centering
        \begin{minipage}[t]{.3\linewidth}
            \centering
            \subfloat[(a) Bi-modal]{\includegraphics[trim={8em 2em 8em 1em},clip, width=\textwidth]{images/bimodal.pdf}}
        \end{minipage}
        \begin{minipage}[t]{.3\linewidth}
            \centering
            \subfloat[(b) Linear]{\includegraphics[trim={8em 2em 8em 1em},clip,width=\textwidth]{images/linear.pdf}}
    \end{minipage}
        \begin{minipage}[t]{.3\linewidth}
            \centering
            \subfloat[(c) Banana]{\includegraphics[trim={8em 2em 8em 1em},clip, ,width=\textwidth]{images/banana.pdf}}
    \end{minipage}
\end{minipage}
\caption{\small Validation environments. Black striped line indicate the randomized initial location of the swarm (similar to \autoref{fig:swarm_env_Special_s}). Striped box in (c) indicate the area of random initialization}
\label{fig:val_env}
\vspace{-1em}
\end{figure}

\subsubsection{Scalability \& Robustness}
The swarm should be able to operate with a wide range of group sizes (i.e. Scalability) and across different types of environment (i.e. Robustness), using the same best controller. In our Scalability experiment, we initiate the swarm in the same environment but with the following swarm sizes [10, 20, 50] to see the impact on performance. In our robustness experiments, we initiate the swarm (of size 20) in different gradient maps ['Bi-modal', 'Linear', 'Banana'] as shown below. Each new arena poses different challenges: \textit{Bi-modal}, requires collective decision on where to go (\autoref{fig:val_env}a); \textit{Linear} poses a less salient gradient stretched out over the full arena (\autoref{fig:val_env}b); \textit{Banana}, the banana function is a classic a non-linear minimization problem \cite{clerc1999swarm} with a curved shallow bottom. For our collective gradient ascent task this function is interesting as it has both shallow gradient and local maxima (\autoref{fig:val_env}c). The experimental parameters we used in all the validation experiments are described in \autoref{tab:parameters_Special_s_val}.


\renewcommand{\arraystretch}{0.6} 
\begin{table}[ht]
\footnotesize
\caption{\small Validation experiment parameters}
\centering
\begin{tabular}[b]{{p{0.18\linewidth} p{0.20\linewidth} p{0.43\linewidth}}}
 & Value & Description\\
\toprule     
Repetitions      &  60    & Number of runs per experiment \\ 
Statistics       & \textit{t}-test & Statistical test \\ \bottomrule 
\toprule
\multicolumn{3}{l}{\textbf{Ratio}: \textit{sub-group division environment}}  \\ \midrule
Swarm size 			 & 20     & Number of robots in a swarm 	\\
Ratio    & 0:4/1:3/2:2 & sub-group division (+inverse) \\
r$_{ratio}$    & \tiny{$0/\frac{1}{4}/\frac{1}{2}/\frac{3}{4}/1$} & Spawn distance from center (r) \\
Arena type & center     & Environment	\\
Eval. time  & 10    & Test duration in minutes \\ 
\bottomrule 
\toprule
\multicolumn{3}{l}{\textbf{Scalability}: \textit{Different swarm size}} \\ \midrule
Swarm size 			 & 10/20/50     & Number of robots in a swarm 	\\
Ratio      & 2:2/adaptive & sub-group division \\
Arena type 			 & center    & Environment 	\\
Eval. time  & 10    & Test duration in minutes \\ 
\bottomrule 
\toprule
\multicolumn{3}{l}{\textbf{Robustness}: \textit{Different environments}} \\ \midrule
Swarm size 			 & 20     & Number of robots in a swarm 	\\
Ratio      &2:2/adaptive & sub-group division \\
Arena type 			 & bi\nobreakdash-mod./lin./ban.     & Environment 	\\
Eval. time  & 10    & Test duration in minutes \\ 
\bottomrule 
\end{tabular}
\vspace{-1.5em}
\label{tab:parameters_Special_s_val}
\end{table}

\subsection{Real world experiment}
Finally, we test our three controllers in our real robot swarm: 1) Baseline; 2) Heterogeneous swarm; 3) Adaptive Heterogeneous swarm. Each controller is tested in a 7x10 meter environment with a light source at the centre producing a gradient (calibrated values are shown in \autoref{fig:thymio}). We start the Thymios at the edge (\texttildelow 8.5m) and log the sensed light value, position, and yaw angle using a serial radio communication channel. We always initialize the Thymios in a random position and orientations before the start of a run and test for 10 minutes. Each controller is tested 3 times to get a robust estimate of its behavior. In the end, we compare the fitness over time and the collective behavior of the swarm. Code base for the hardware implementation is available at: \url{https://github.com/fudavd/CrazyThymio}

\begin{figure}
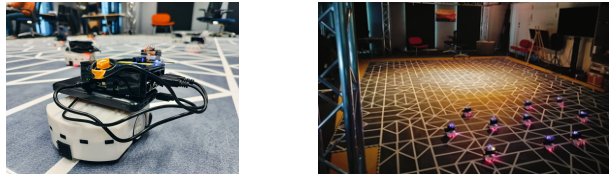

  \centering
\begin{minipage}[c]{0.85\linewidth}
\centering
        \begin{minipage}[t]{.4\linewidth}
            \centering
            \subfloat[(a) Robot]{\includegraphics[height=6.5em]{images/real_exp/thymio.png}}
        \end{minipage}
        \begin{minipage}[t]{.5\linewidth}
            \centering
            \subfloat[(b) Task environment]{\includegraphics[height=6.5em]{images/real_exp/arena.png}}
    \end{minipage}
\end{minipage}\vspace{-0.5em}
\caption{\small Experimental setup. (a) Robot design: Thymio with Raspberry Pi 5, and custom on-board position and light sensor. (b) Our swarm with different sub-groups (colored red and green) in simulation.}
\label{fig:thymio}
\end{figure}

\section{Results}\label{sec:res}
We provide full access to all the experimental data: \newline\href{https://doi.org/10.34894/0VSN8Z}{https://doi.org/10.34894/0VSN8Z}

\subsection{Evolutionary experiment}
We measure the efficacy by the mean and maximum fitness averaged over the 10 independent evolutionary runs at every generation. 

\begin{figure}[ht]
\vspace{-0.5em}
  \includegraphics[width=0.85\linewidth]{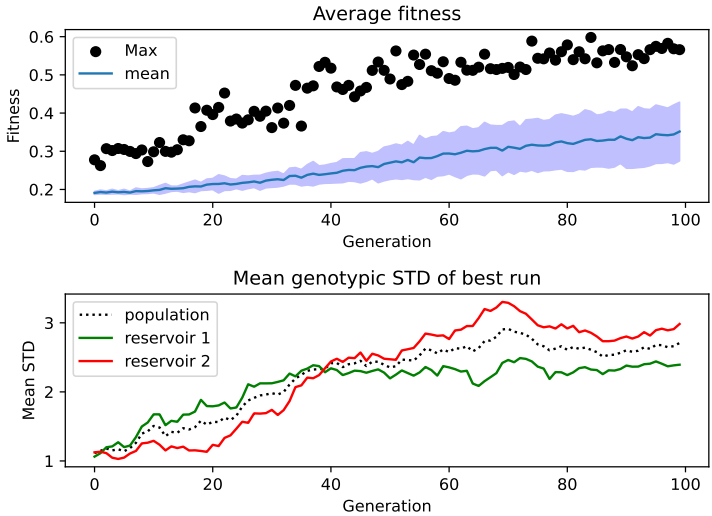}
  \caption{\small Top: The mean$\pm STD$ (line) and average max (dot) fitness over 100 generations (averaged over 10 runs). Bottom: Mean genotypic standard deviation (STD) within the population during the best run (per generation). The overall population mean as a black dotted line and each separate reservoir in solid red and green.}
  \label{fig:fitness_Special_s}
\end{figure}

The results of our evolutionary experiment are presented in \autoref{fig:fitness_Special_s}. We see a steady growth in mean fitness which indicate that the CMA-ES the collective gradient sensing task is learnable with our heterogeneous swarm. The maximum overall fitness is indicated by the black dots. We see a rapid increase around generation 20 which plateaus more or less around generation 50. 

At the bottom of \autoref{fig:fitness_Special_s} we show the genotypical variation of our population in the best evolutionary run (calculated as mean Standard Deviation, STD). The rapid increase at generation 20 shows as an increase in variation, while the plateauing of fitness at generation 50 coincides with the steepest decline in genotypical variation. It is interesting to see that the both reservoir seem to interchangeably adapt their genotypical variation. Where the first reservoir (green) increases first but stabilizes relatively soon, while the second reservoir (red) remains relatively fixed at first and increases when the first reservoir starts to flatten out. This may indicate concurrent adaptation due to learned task distribution.

\subsection{Validation experiments}

\begin{figure*}[ht!]
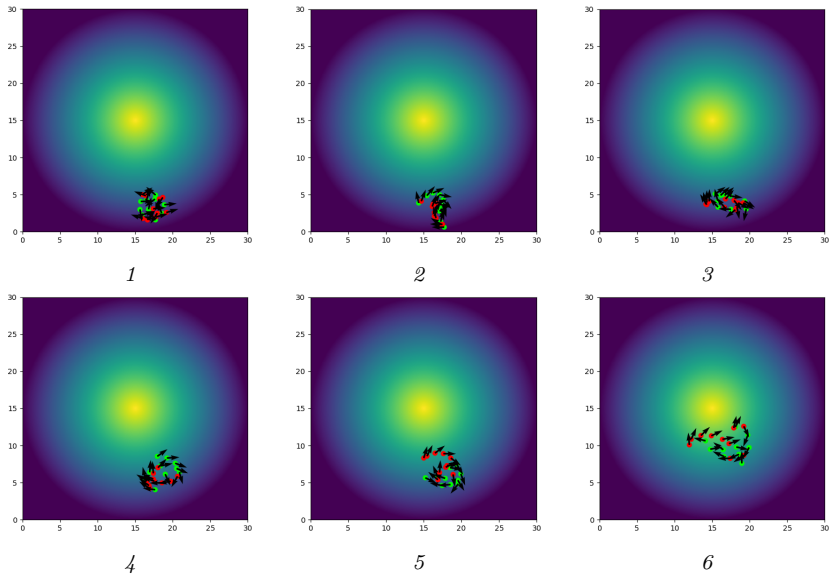

    \centering
    \vspace{-1em}
    \begin{minipage}[c]{0.95\textwidth}
        \begin{minipage}[t]{.3\textwidth}
            \centering
            \subfloat[1]{\includegraphics[width=\textwidth]{images/align/1.png}}
        \end{minipage}
        \hfill
        \begin{minipage}[t]{.3\textwidth}
            \centering
            \subfloat[2]{\includegraphics[width=\textwidth]{images/align/2.png}}
        \end{minipage}
        \hfill
        \begin{minipage}[t]{.3\textwidth}
            \centering
            \subfloat[3]{\includegraphics[width=\textwidth]{images/align/3.png}}
        \end{minipage}
        \hfill
        \begin{minipage}[t]{.3\textwidth}
            \centering
            \subfloat[4]{\includegraphics[width=\textwidth]{images/align/4.png}}
        \end{minipage}
        \hfill
        \begin{minipage}[t]{.3\textwidth}
            \centering
            \subfloat[5]{\includegraphics[width=\textwidth]{images/align/5.png}}
        \end{minipage}
        \hfill
        \begin{minipage}[t]{.3\textwidth}
            \centering
            \subfloat[6]{\includegraphics[width=\textwidth]{images/align/6.png}}
        \end{minipage}
\end{minipage}
    \caption{\small Snapshots of the interesting timeframes during the re-test experiment. The numbers correspond to the order of the vertical bars in Figure are corresponding to each line plot in \autoref{fig:retest}. (1) In the beginning, the swarm spreads out to search for the gradient. (2) The green sub-group senses a gradient and 'aligns' the swarm to the left. (3) The gradient is lost which results in the swarm dispersing in different directions (i.e. decreasing alignment). (4) Red sub-group senses the gradient and directs the swarm towards the light source. (5) Red slowly pulls in more green swarm members. (6) the swarm starts to spread around the light source. }
    \label{fig:order_Special_s}\vspace{-1em}
\end{figure*}

\begin{figure}[ht]
\vspace{0.5em}
  \includegraphics[width=0.85\linewidth]{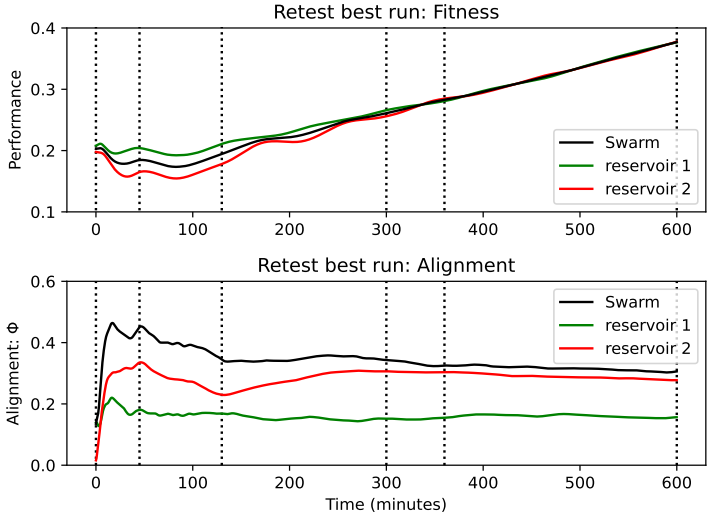}
  \caption{\small Flocking behavior of the final best controller re-tested in the same environment. The three line plots on the left show the fitness (i.e. mean scalar light value) over time on the top, and the order (alignment) at the bottom. The red and green indicate the order of each sub-group, while the black line indicates the swarm as a whole. Vertical lines that correspond with interesting time frames in \autoref{fig:order_Special_s}.}
  \label{fig:retest}
\vspace{-1.em}
  \end{figure}

\subsubsection{Collective behavior \& sub-group interactions}
\autoref{fig:order_Special_s} and \ref{fig:retest} show the results of our best controller retested in the same environment. A video of this run is provided in the supplementary material (videos can be found here\footnote{\url{https://tinyurl.com/mr38uxcu}}). During this trial we can measure the fitness over time, with its final value around $0.38$. More interestingly, we measure the overall alignment of the swarm and the alignment for each sub-group. We provide snapshots of the swarm in the arena in \autoref{fig:order_Special_s}, with the corresponding the time-frames represented as dotted vertical lines in \autoref{fig:retest} (snapshot order corresponds with time). 

In \autoref{fig:retest}, at the start we see a low initial swarm alignment (black line) that quickly increases (corresponding to \autoref{fig:order_Special_s}: 1-2). This rapid increase is mainly caused by sub-group 2 (in red) who tends to align more amongst themselves during the whole run. Sub-group 1 (in green) does not align that much but finds the gradient faster at the start (see snapshot: 3), indicated by a higher fitness. Red follows shortly as a group given their alignment increase. In a later stage, red dominates the swarm's alignment and behavior, thereby concurrently pulling and pushing green towards the center (corresponding to \autoref{fig:order_Special_s}: 4-5). This is also visible by the oscillating sub-group performance.

In \autoref{tab:ratio_robustness} we evaluated the same best swarm in different setups; we repeat the evaluation at different ratios of sub-groups and at different starting distances from the gradient center. Every configuration was repeated $60$ times.
We observe how starting closer to the target results in higher fitness.
We find that evenly mixed sub-groups at distances far away from the center ($r_{ratio}\ge0.5$) perform statistically significantly higher than any of the extremes (i.e. fully green/red, $p\le0.05, df=118$).

When visually inspecting the behavior of the swarms at $r_{ratio}=0$, we observe how both extreme ratios of sub-groups are capable of staying at the center of the gradient, but with different strategies. While the swarm made of only red robots seem to circle around the center with some distance, the swarm made of only green robots seem to fully occupy the space around the center, coinciding with a higher fitness. When evaluating the sub-group ratios at their extremes, we see that both independent sub-groups consistently outperform the best controllers of the first generation in our evolutionary experiments ($\sim 0.3$ at $r_{ratio}=1.0$). Nevertheless, a mix sub-group ratio out performs for all $r_{ratio}\ge0$, indicating an advantage of sub-group interaction. Furthermore, we see a tendency of the second red sub-group to correlate with a higher performance at $r_{ratio}\ge1.0$.

\newcommand*{\MinNumber}{0.05}
\newcommand*{\MidNumber}{0.7}
\newcommand*{\MaxNumber}{0.8}
\newcommand{\mytextcolor}[2]{\ifdim#1pt>\MidNumber pt\textcolor{gray!30}{#1}\else #1\fi}
\newcommand{\circletextcolor}[2]{\ifdim#1pt>\MidNumber pt\textcolor{gray!30}{\circlered{#2}{#1}}\else \circlered{#2}{#1}\fi}

\newcommand{\ApplyGradient}[2]{
    \FPeval{\result}{100*(#1-\MinNumber)/(\MaxNumber-\MinNumber)}
    \centering
    \edef\x{\noexpand\cellcolor{black!\result}}
    \x\mytextcolor{#1}
}
\newcommand{\ApplyGradientCircle}[2]{
    \FPeval{\result}{100*(#1-\MinNumber)/(\MaxNumber-\MinNumber)}
    \centering
    \edef\x{\noexpand\cellcolor{black!\result}}
    \x\circletextcolor{#1}{#2}
}
\newcommand{\ApplyGradientCircleA}[2]{\ApplyGradientCircle{#1}{a}}
\newcommand{\ApplyGradientCircleB}[2]{\ApplyGradientCircle{#1}{b}}
\newcommand{\ApplyGradientCircleC}[2]{\ApplyGradientCircle{#1}{c}}
\newcommand{\ApplyGradientCircleD}[2]{\ApplyGradientCircle{#1}{d}}
\newcommand{\ApplyGradientCircleE}[2]{\ApplyGradientCircle{#1}{e}}
\newcommand{\ApplyGradientCircleF}[2]{\ApplyGradientCircle{#1}{f}}
\newcolumntype{A}{>{\collectcell\ApplyGradientCircleA}p{0.6cm}<{\endcollectcell}}
\newcolumntype{B}{>{\collectcell\ApplyGradientCircleB}p{0.6cm}<{\endcollectcell}}
\newcolumntype{C}{>{\collectcell\ApplyGradientCircleC}p{0.6cm}<{\endcollectcell}}
\newcolumntype{D}{>{\collectcell\ApplyGradientCircleD}p{0.6cm}<{\endcollectcell}}
\newcolumntype{E}{>{\collectcell\ApplyGradientCircleE}p{0.6cm}<{\endcollectcell}}
\newcolumntype{F}{>{\collectcell\ApplyGradientCircleF}p{0.6cm}<{\endcollectcell}}
\newcolumntype{R}{>{\collectcell\ApplyGradient}p{0.6cm}<{\endcollectcell}}
\newcolumntype{Q}{>{\collectcell\ApplyGradient}p{0.6cm}<{\endcollectcell}}
\begin{table}[ht]
  \centering
  \renewcommand{\arraystretch}{1.6}
  \small
  \begin{tabular}{
    c  *{5}{R}
  }
  \toprule
      \multicolumn{1}{c}{ratio~$\rightarrow$}& 
      \multicolumn{1}{c}{\cellcolor{green!30}$4:0$} &
      \multicolumn{1}{c}{$3:1$} & 
      \multicolumn{1}{c}{$2:2$} & 
      \multicolumn{1}{c}{$1:3$} &
      \multicolumn{1}{c}{\cellcolor{red!30}$0:4$} \\
    \midrule


    
    $r_{ratio}=0.00$ & \multicolumn{1}{A}{0.79} & 0.78 & 0.77 & 0.73 & 0.69 \\
    $r_{ratio}=0.25$ & \multicolumn{1}{B}{0.76} & 0.76 & 0.75 & 0.72 & 0.68 \\
    $r_{ratio}=0.50$ & 0.70 & \multicolumn{1}{C}{0.71} & 0.70 & 0.66 & 0.64 \\
    $r_{ratio}=0.75$ & 0.56 & 0.60 & \multicolumn{1}{D}{0.60} & 0.57 & 0.52 \\
    $r_{ratio}=1.00$ & 0.33 & 0.41 & \multicolumn{1}{E}{0.43} & 0.43 & 0.38 \\
    $r_{ratio}=1.25$ & 0.06 & 0.08 & \multicolumn{1}{F}{0.13} & 0.09 & 0.12 \\
    \bottomrule            
  \end{tabular}
  \begin{tikzpicture}[remember picture,overlay]
    \draw[red,thick] (a) ++(-4.85mm,-2.7mm) rectangle ++(9.85mm,5.8mm);
    \draw[red,thick] (b) ++(-4.85mm,-2.7mm) rectangle ++(9.85mm,5.8mm);
    \draw[red!70, dashed] (b) ++(5.3mm,-2.7mm) rectangle ++(9.85mm,5.8mm);
    \draw[red,thick] (c) ++(-4.85mm,-2.7mm) rectangle ++(9.85mm,5.8mm);
    \draw[red!70, dashed] (c) ++(5.3mm,-2.7mm) rectangle ++(9.85mm,5.8mm);
    \draw[red,thick] (d) ++(-4.85mm,-2.7mm) rectangle ++(9.85mm,5.8mm);
    \draw[red!70, dashed] (d) ++(-15.2mm,-2.7mm) rectangle ++(9.85mm,5.8mm);
    \draw[red,thick] (e) ++(-4.85mm,-2.7mm) rectangle ++(9.85mm,5.8mm);
    \draw[red,thick] (f) ++(-4.85mm,-2.7mm) rectangle ++(9.85mm,5.8mm);
    \draw[red!70, dashed] (f) ++(15.55mm,-2.7mm) rectangle ++(9.85mm,5.8mm);
  \end{tikzpicture}
  \hspace{1mm}
  \vspace{1em}
  \caption{\small Average fitness values ($N=60$) of different sub-group ratios (green:red) at different starting positions ($r_{ratio}$ = distance to the optimum). Sub-group ratios vary from solely sub-group 1 (green) to solely sub-group 2 (red). Deviation from the mean varies from $0.01$ to $0.14$, depending on the configuration. Solid red boxes indicate maximum fitness at a given $r_{ratio}$, while the dashed boxed indicate no statistically significant differences with respect to the maximum.
    \label{tab:ratio_robustness}
  }
  \vspace{-2em}
\end{table}

\subsubsection{Online regulatory mechanism}
Based on the results of \autoref{tab:ratio_robustness} we link light intensity threshold values with the best performing sup-group ratio. Heuristically, we design a probabilistic state machine such that members in the swarm adapt their behavior automatically, to reflect the optimal ratio on a swarm level using only local information. Thresholds are based on the light intensities at $[0.125, 0.375, 0.625, 0.875]\times r_{ratio}$. This results in the following function for the probability of expressing the first green sub-group behavior ($P_{green}$).
Here, $light$ refers to the current light intensity measured by the robot ([0, 255]) and $P_{red} = 1 - P_{green}$. 

\[   
P_{green}(light) = 
     \begin{cases}
       \text{1.0} &\quad\text{if $light$} >229\\
       \text{0.75} &\quad\text{if $light$} \in (76,229] \\
       \text{0.50} &\quad\text{if $light$} \le76\\
     \end{cases}
\]

\subsubsection{Scalability \& Robustness}
We test the best swarm controller with and without our online regulatory mechanism in 6 different environments (3 scalability experiments, and 3 robustness). The results of these experiments are presented below (see \autoref{tab:res_val}). For \textit{Scalability}, the performance of the best controller seems to be positively correlated by the size of the swarm which indicate sensitivity to swarm size. This is positive correlation is less apparent in the adaptive controller whose performance is statistically significantly higher for each swarm size comparison (10: $p\le0.05$, 20: $p\le0.01$, 50: $p\le0.001$). \textit{Robustness} results show the same tendency of the adaptive controller outperform the controller without regulatory mechanism, although these differences were not statistically significant. On an aggregate our online regulatory mechanism outperforms the best controller ($N=360$): Best $0.40\pm0.18$ \textit{vs.} Adaptive $0.45\pm0.22$. The aggregated means are statistically significantly different with Bonferroni correction ($p\le0.01/\alpha$ where $\alpha=6$, $df=718$).

\begin{table}[ht]
    \centering
    \begin{tabular}{c ccc}
       \toprule
       \textbf{Scalability} & \multicolumn{3}{c}{\textit{Swarm size}}\\
       N=60  &  10 & 20 & 50\\
         \midrule
       Best  &  $0.35\pm0.10$ & $0.39 \pm 0.12$ & $0.47 \pm 0.03$\\
       
       Adaptive  &  $\hphantom{^*}\bm{0.40\pm0.12}^*$ & $\hphantom{^*}\bm{0.43\pm0.077}^*$ & $\hphantom{^*}\bm{0.49\pm0.03}^*$ \\
       \bottomrule
       \toprule
       \textbf{Robustness} & \multicolumn{3}{c}{\textit{Arena type}}\\
       N=60 &  \small{Bi-modal} & \small{Linear} &  \small{Banana} \\
       \midrule
       Best  &  $0.43 \pm 0.17$ & $0.51 \pm 0.19$ & $0.25 \pm 0.29$\\
       
       Adaptive  &  $\bm{0.47 \pm 0.19}$ & $\bm{0.59 \pm 0.25}$ & $\bm{0.31 \pm 0.35}$ \\
       \bottomrule
       
    \end{tabular}
    \caption{\small Validation experiments}
    \label{tab:res_val}
\end{table}

\subsection{Real world experiment}
\begin{figure*}[ht!]
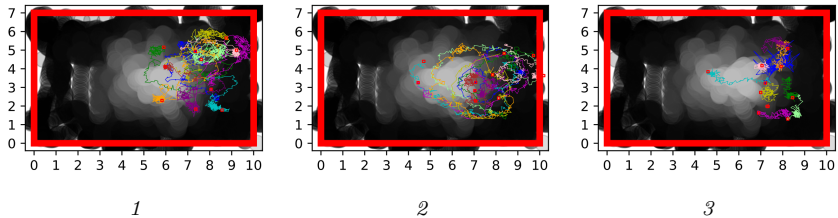

    \centering
    \begin{minipage}[c]{\textwidth}
    \centering
        \begin{minipage}[t]{.32\textwidth}
            \centering
            \subfloat[1]{\includegraphics[height=0.675\textwidth]{images/real_exp/baseline_1.png}}
        \end{minipage}
        \begin{minipage}[t]{.32\textwidth}
            \centering
            \subfloat[2]{\includegraphics[height=0.675\textwidth]{images/real_exp/hetero_2.png}}
        \end{minipage}
        \begin{minipage}[t]{.32\textwidth}
            \centering
            \subfloat[3]{\includegraphics[height=0.675\textwidth]{images/real_exp/hetero_adaptif_3.png}}
        \end{minipage}
\end{minipage}\vspace{-0.5em}
\caption{\label{fig:res_traj} Trajectories of the best run on real robot swarm.}
\vspace{-1em}
\end{figure*}

In the end, we tested our controllers on a real robot swarm consisting of 10 members. We show the trajectories of the best runs per controller in \autoref{fig:res_traj}. In all runs, the swarm spread out from the initial positions and moved towards the centre. Similar to the simulated behavior, we can see the robot trajectories making a clockwise circular pattern around their centroid. Furthermore, in the Adaptive Heterogeneous control (3rd panel) there is some behavioral switching visible in robots that moved closer to the centre (Blue, Pink, and Mint).  

\section{Discussion}\label{sec:disc}\normalfont
We successfully evolved self-organized sub-group specialization in a swarm of robots, a difficult task due to the complexity of swarm dynamics. This becomes even more evident when factoring in the added complexity of optimizing between-group interactions in conjunction with specific specialization itself. The strength of our framework lies in its simple approach to learning these behaviors, lending itself to broad applicability.
We demonstrated our methods' effectiveness in the context of an emergent perception task for gradient sensing, which is similar to most source localization tasks common in the swarm robotics literature.
Transferring our method to more complex tasks is also trivial, as 1) the use of task-agnostic nature of RNNs can express a wide array of behaviors, thus require no controller design adaptation; and 2) the simplicity of our group-level fitness function can be easily adapted, as its design requires minimal insight into the optimal solution (i.e. no presupposition on certain behaviors or specific task divisions are required).
Our general evolutionary approach to learn `sub-group specialization' is thus extremely useful for any type of task. 

The results of our validation experiments show the emergence of specialization in our sub-groups. Sub-group 2 (red) seems to follow the gradient better at low intensities than sub-group 1 (see start of \autoref{fig:retest}) and tends to move more in coordination (i.e. higher alignment).
In contrast, sub-group 1 performs better when initialized near the center, and shows consistently lower alignment. This indicates that sub-group 1 shows less sensitivity to the swarm's overall behavior and tend to be more greedy as a sub-group. In contrast, sub-group 2 shows more exploratory behavior which provides more coordinated movement of the swarm when the gradient is found. 

Exploration and exploitation are fundamental principles in optimization in general. You could interpret our task as such, where we learn our swarm to act as an optimizer that maximizes its local light value. The above mentioned behavioral findings show that the solution converged on a similar exploitation-exploration task division within their sub-group specializations. 
This task division is interesting as we did not encourage such collective behavior or specialization in our fitness. Arriving at these fundamental optimization principles in the context of sub-group swarm interactions without prior knowledge shows the power of automated design for finding collective behaviors suitable for a user-defined task.

From \autoref{tab:ratio_robustness} we can see that only employing one of the two sub-group at lower light intensities leads to lower performances than using mixed groups. This means that the effectiveness of specialization is enhanced by sub-group interactions. 
The collaboration becomes unnecessary when the robots are placed too close to the center of the gradient ($r_{ratio}\le 0.25$). 
The online regulatory mechanism furthermore shows the successful division of specialized tasks within our swarm as it significantly improves performance of the best controller. The idea to utilize the evolved specializations by biasing an individual's behavior towards a certain reservoir is similar to nature, where phenotypic plasticity emphasizes the expression of specific parts of the genome (i.e. our reservoirs) to obtain higher task competency on a swarm level. In our experiments we successfully show that this straightforward implementation of phenotypic plasticity results in better scalable control with significantly higher performance overall.

\section{Conclusion}\label{sec:conc}
Incorporating specialized behaviors within robot swarms holds the potential to significantly enhance overall swarm efficacy and robustness. However, this endeavor poses formidable challenges. Designing controllers for homogeneous swarms is inherently complex, and extending this to sub-groups within the swarm compounds the difficulty due to the added intricacy of sub-group interactions. In this paper, we show a viable approach to solve this challenge by co-evolving controllers for heterogeneous swarm behavior with only specifying group level task performance in the fitness function. We demonstrate that our evolved controllers show clear specialized sub-group behavior and that sub-group interactions lead to enhanced collective behavior. Furthermore, we show that learned sub-group behaviors can be used dynamically for an adaptive swarm using an online regulatory mechanism to further enhance performance and scalability. 

In the future, we propose to extend our work to encompass a broader spectrum of tasks, which could reveal other emergent specializations (communication, line-following, mapping). Additionally, we foresee further improvements in the creation of more sophisticated controller designs where we can evolve the number of task-specializations and possibly the online adaptation rules to regulate phenotypic plasticity. Finally, we would like to test our controllers in more complex and dynamic real world applications. For this we would require  continuously adapting heterogeneous control.




